\UseRawInputEncoding
\documentclass[runningheads]{llncs}
\usepackage{geometry}
\usepackage{graphicx}
\usepackage[ruled,vlined]{algorithm2e}
\usepackage{setspace}
\usepackage{algorithmicx}
\algnewcommand{\LeftComment}[1]{\Statex \(\triangleright\) #1}
\SetAlCapFnt{\footnotesize}
\SetAlCapNameFnt{\footnotesize}
\usepackage{adjustbox}
\usepackage{booktabs}
\usepackage{makecell}

\usepackage{subfigure}
\usepackage{color}
\usepackage{tabularx}
\usepackage{amssymb}
\usepackage{amsmath}
\usepackage{footmisc}
\usepackage{nameref}
\usepackage{balance}

\usepackage{float}
\usepackage{placeins}
\usepackage{afterpage}

\begin{document}
%                       TITLE
%==========================================================================
\title{SplitFed: When Federated Learning Meets Split Learning}
\author{Chandra Thapa\inst{1} \and M.A.P. Chamikara\inst{1} \and Seyit Camtepe\inst{1} \and Lichao Sun\inst{2}}
\institute{CSIRO Data61, Australia \and
     Lehigh University, USA\\
	\email{\{chandra.thapa, chamikara.arachchige, seyit.camtepe\}@data61.csiro.au}\\
	\email{lis221@lehigh.edu}
}

\authorrunning{Thapa et al.}
\maketitle

%==========================================================================
%                       ABSTRACT
%==========================================================================

\begin{abstract}
%\textcolor{red}{Double-blind review, 7 pages technical contains, 2 pages for references, deadline: September 1}

Federated learning (FL) and split learning (SL) are two  popular distributed machine learning approaches. Both follow a model-to-data scenario; clients train and test machine learning models without sharing raw data. 
SL provides better model privacy than FL due to the machine learning model architecture split between clients and the server. Moreover, the split model makes SL a better option for resource-constrained environments.
However, SL performs slower than FL due to the relay-based training across multiple clients. In this regard, this paper presents a novel approach, named \emph{splitfed learning (SFL)}, that amalgamates the two approaches eliminating their inherent drawbacks, along with a refined architectural configuration incorporating differential privacy and PixelDP to enhance data privacy and model robustness.
%
%Moreover, SFL splits the network architecture between the clients and server as in SL and incorporates the parallel ML model update paradigm as in FL to achieve efficiency and model privacy. 
%
Our analysis and empirical results demonstrate that (pure) SFL provides similar test accuracy and communication efficiency as SL while significantly decreasing its computation time per global epoch than in SL for multiple clients. 
Furthermore, as in SL, its communication efficiency over FL improves with the number of clients. 
Besides, the performance of SFL with privacy and robustness measures is further evaluated under extended experimental settings.
%using differential privacy and PixelDP. 
%
%Our test results demonstrate that splitfed performance on AlexNet with the MNIST dataset under various privacy levels. As a trade-off to privacy, the model convergence is slower with an increase in noise addition in the mechanism.

\end{abstract}

%==========================================================================
%                       INTRODUCTION
%==========================================================================
\section{Introduction}

Distributed Collaborative Machine Learning (DCML) is popular due to its default data privacy benefits~\cite{survey_p1}. Unlike the conventional approach, where the data is centrally pooled and accessed, DCML enables machine learning without having to transfer data from data custodians to any untrusted party. Moreover, analysts have no access to raw data; instead, the machine learning (ML) model is transferred to the data curator for processing. Besides, it enables computation on multiple systems or servers and distributed devices. 

The most popular DCML approaches are federated learning~\cite{fedkonecny2,fed1} and split learning~\cite{SplitNN}. Federated learning (FL) trains a full (complete) ML model on the distributed clients with their local data and later aggregates the locally trained full ML models to form a global model in a server. The main advantage of FL is that it allows parallel, hence efficient, ML model training across many clients.

\textbf{Computational requirement at the client-side and model privacy during ML training in FL.}  The main disadvantage of FL is that each client needs to run the full ML model, and resource-constrained clients, such as available in the Internet of Things, could not afford to run the full model. This case is prevalent if the ML models are deep learning models. Besides, there is a privacy concern from the model's privacy perspective during training because the server and clients have full access to the local and global models. 

To address these concerns, split learning (SL) was introduced. SL splits the full ML model into multiple smaller network portions and train them separately on a server, and distributed clients with their local data. Assigning only a part of the network to train at the client-side reduces processing load (compared to that of running a complete network as in FL), which is significant in ML computation on resource-constrained devices~\cite{splitlearning}. Besides, a client has no access to the server-side model and vice-versa.

%Although DCML optimization has been a frequently addressed topic, in this work, our primary focus is on the intersection of privacy preservation and efficiency ({\em e.g.,} training time for an ML model) in DCML. %(without DP, HE and MPC). %

%We broadly divide these approaches into two types; the approaches (1) without network splitting, {\em e.g.,} FL and distributed synchronous SGD, and (2) with network splitting, {\em e.g.,} SL. Distributed synchronous SGD, {\em e.g.,} downpour SGD~\cite{distrubedml21}, works in a similar essence to FL. The difference is that the (local and global) updates are based on one batch of training data in the distributed synchronous SGD, whereas, in FL, a client trains the network over all its local data for some local epochs\footnote{In one local epoch of client model training, one forward and its respective back-propagation completes for all available local data in that client.}, before updating the model to the server. Considering the similarity and high latency in distributed synchronous SGD, we choose only FL from the first type in this work. 

%	SL limits the client-side network portion down to a few layers. Thus, it enables the reduction in client-side computation, and model privacy due to the network split.
	
\textbf{Training time overhead in SL.}  Despite the advantages of SL, there is a primary issue. The relay-based training in SL makes the clients' resources idle because only one client engages with the server at one instance; causing a significant increase in the training overhead with many clients.

%Now by considering the pros and cons of FL and SL, we have a natural question: \emph{Can we combine the SL and FL strategically to exploit their main advantages?}

To address these issues in FL and SL, this paper proposes a novel architecture called \emph{splitfed learning (SFL)}. SFL considers the advantages of FL and SL, while emphasizing on data privacy, and robustness of the model. Refer to Table~\ref{table:compare} for its abstract comparison with FL and SL. Our contributions are mainly two-fold: Firstly, we are the first to propose SFL. Data privacy and model's robustness are enhanced at the architectural level in SFL by the differential privacy-based measures~\cite{abadi2016deep} and PixelDP~\cite{lecuyer2019certified}. Secondly, to demonstrate the feasibility of SFL, we present comparative performance measurements of FL, SL, and SFL by considering four standard datasets and four popular models. Based on our analyses and empirical results, SFL provides an excellent solution that offers better model privacy than FL, and it is faster than SL with a similar performance to SL in model accuracy and communication efficiency. 

%Moreover, SFL enables distributed processing across clients with low computing resources as SL, but with an improvement in its training speed.  
	    
Overall, SFL is beneficial for resource-constrained environments where full model training and deployment are not feasible, and fast model training time is required to periodically update the global model based on a continually updating dataset over time ({\em e.g.}, data stream). These environments characterize various domains, including health, {\em e.g.}, real-time anomaly detection in a network with multiple Internet of Medical Things\footnote{The examples of the Internet of  Medical Things include glucose monitoring devices, open artificial pancreas systems, wearable electrocardiogram (ECG) monitoring devices, and smart lenses.} connected via gateways and finance, {\em e.g.}, privacy-preserving credit card fraud detection.       

\begin{table}[t]
	  \centering
	    \setlength{\tabcolsep}{0.5pt}
	    \renewcommand{\arraystretch}{1.2}
		\caption{An abstract comparison of split learning (SL), federated learning (FL), and splitfed learning (SFL).}
		\label{table:compare}
		
		\begin{adjustbox}{max width=0.6\textwidth}
		%\begin{tabular}{P{1.5cm}P{1.5cm}P{2.8cm}P{1.5cm}P{1.5cm}P{1.5cm}}
		\begin{tabular}{c|c|c|c|c|c}
			\toprule
			\makecell{Learning\\ approach} 		& \makecell{Model\\ aggregation}			& \makecell{Model privacy advantage\\ by splitted model} &  \makecell{Client-side\\ training} 	& \makecell{Distributed\\ computing} &  \makecell{Access to\\ raw data}\\
			\midrule
			SL					&  No 	& Yes & Sequential & Yes & No \\
			\hline
			FL			& Yes 		& No & Parallel & Yes & No\\
			\hline
			SFL  	&  Yes 		& Yes & Parallel & Yes & No\\
			\bottomrule
		\end{tabular}
	\end{adjustbox}
	\vspace{-0.3cm}	
	\end{table}
	
%==========================================================================
%                       BACKGROUND
%==========================================================================
\section{Background and Related Works} \label{sec:1}
% In this section, we provide some background on federated learning, split learning and differential privacy before introducing our proposed architecture. 
	%\subsection{Federated learning}
	
	%Federated learning~\cite{fedkonecny2,fed1,fedkonecny3} is a collaborative machine learning technique.
	%developed by google to train machine learning models on distributed devices ({\em e.g.,} mobile phones). 
	%
	%It pushes the computations to the edge devices and removes the necessity to pool the raw data from the data curator to train an ML algorithm. This way, it enables the privacy of the raw data. 
	Federated learning~\cite{fedkonecny2,fed1,fedkonecny3} trains a complete ML network/algorithm at each client on its local data in parallel for a certain number of local epochs, and then the local updates are sent to the server for aggregation~\cite{fed1}. This way, the server forms a global model and completes one global epoch\footnote{When forward propagation and back-propagation are completed for all available datasets across all participating clients for one cycle, it is called one global epoch.}. The learned parameters of the global model are then sent back to all clients to train for the next round. This process continues until the algorithm converges. %Besides, in FL, the local epochs contribute to stabilizing the locally trained model by learning iteratively over the local data for some time before updating to the server. This can contribute to the fast convergence of the global model in terms of the number of communications with the server.  
	In this paper, we consider the federated averaging (FedAvg) algorithm~\cite{fed1} for model aggregations in FL. FedAvg considers a weighted average of the gradients for the model updates. 
	Split learning~\cite{splitlearning,SplitNN} splits a deep learning network $\mathbf{W}$ into multiple portions, and these portions are processed and computed on different devices. In a simple setting, $\mathbf{W}$ is split into two portions $\mathbf{W}^{\textup{C}} $ and $\mathbf{W}^{\textup{S}}$, called client-side network and server-side network, respectively. 
	%$\mathbf{W}$ includes weights and bias. 
	The clients, where the data reside, commit only to the client-side portion of the network, and the server commits only to the server-side portion of the network. The communication involves sending activations, called \emph{smashed data}, of the split layer, called \emph{cut layer}, of the client-side network to the server, and receiving the gradients of the smashed data from the server-side operations.   
	%
	%The client-side and server-side portions collectively form the full network $\mathbf{W}$. 
	%
	%The training of the network is done by a sequence of distributed training processes. In the simple setup, the forward propagation and the back-propagation take place in the following way; With the raw data, a client trains the network up to a certain layer of the network, called the \emph{cut layer}, and sends the activations of the cut layer, also called \emph{smashed data}, to the server. Then, the server carries out the training of the remaining layers with the smashed data that it received from the client. 
	%This completes a single forward propagation. 
	%Next, the server carries out the back-propagation up to the cut layer and sends the gradients of the smashed data to the client. With the gradients, the client carries out its back-propagation on the remaining network (i.e., up to the first layer of the network). 
	%
	%This completes a single pass of the back-propagation between a client and the server. 
	%
	%This process continues till model reaches its convergence.
	%This forward propagation and back-propagation process continues until the network gets trained with all the available clients and reaches its convergence. 
	%
	%In SL, the architectural configurations are assumed to be conducted by a trusted party ({\em e.g.,} researchers) with direct access to the main server. This authorized party selects the ML model (based on the application) and network splitting (finding the cut layer) at the beginning of the learning. 
	The synchronization of the learning process with multiple clients is done either in a centralized mode or peer-to-peer mode in SL~\cite{SplitNN}. 
Differential privacy (DP) is a privacy model that defines privacy in terms of stochastic frameworks \cite{diffprivacybook,differentialprivacy}. DP can be formally defined as follows: 
%DP comes with two important parameters, namely $ \epsilon $ and $\delta $, where $ \epsilon $ and $ \delta $ refer to \emph{privacy level}, and \emph{probability of failure}, respectively. These parameters indicate the level of difficulty in predicting individual data points with high confidence. 
%Let us consider two datasets $ x $ and $ y $ such that $ y $ is formed from $ x $ by removing only one data point of $ x $. Hence, $x$ and $y$ are called adjacent datasets. The requirement of differential privacy for any mechanism $ \mathcal{M} $ on these two datasets is that $ \mathcal{M}(x) $ and $ \mathcal{M} (y) $ should be almost indistinguishable for a given range of results, $R$. We can provide a formal definition of differential privacy as given below:

	\begin{definition}
		A mechanism $\mathcal{M}$ is considered to be $ (\epsilon,\delta) $-differential private if, for all adjacent datasets, $ x $ and $ y $, and for all possible subsets of results, $ R $ of the mechanism, the following holds: 
		%$\mathbb{P}[\mathcal{M}(x) \in R] \leq e^\epsilon*\mathbb{P}[\mathcal{M}(y)\in R] + \delta$.
        \begin{align*}
			\mathbb{P}[\mathcal{M}(x) \in R] \leq e^\epsilon * \mathbb{P}[\mathcal{M}(y)\in R] + \delta. 
		\end{align*}
 	\end{definition}
	
	%If $ \delta = 0 $ in the above equation, then it is called $ \epsilon$-differential privacy, and for $ \epsilon = 0 $, $ \mathcal{M} (x) $ and $ \mathcal{M}(y) $ are identically distributed. This implies no information leakage on the data point they differ; thus absolute privacy is guaranteed. However, for $ \epsilon > 0 $, some information about the data point they differ can be revealed. 
	Practically, the values of $ \epsilon $ (privacy budget) and $ \delta $ (probability of failure) should be kept as small as possible to maintain a high level of privacy.
	%, and it can not be $0$, because of the utility issue. 
	However, the smaller the values of $ \epsilon $ and $ \delta $, the higher the noise applied to the input data by the DP algorithm. 
	%A higher level of noise will result in a lower level of accuracy. Hence, there is always a trade-off between utility and privacy. 

%	
%	
%	The differentially-private weighted averaging limits the client-to-fed information leakage and also addresses the possible client-to-client leakage. In this process, each client updates their differentially private gradients to the fed. Thus the server can not accurately predict the actual gradients (due to the differential privacy guarantee). All the clients get the global client-side model, which is updated after weighted averaging of the differentially private local gradients. Thus, it is extremely difficult for a client to predict the inputs of the other clients by observing the global client-side model.

	\begin{figure}[t]
	    \centering
		\includegraphics[width=0.5\linewidth]{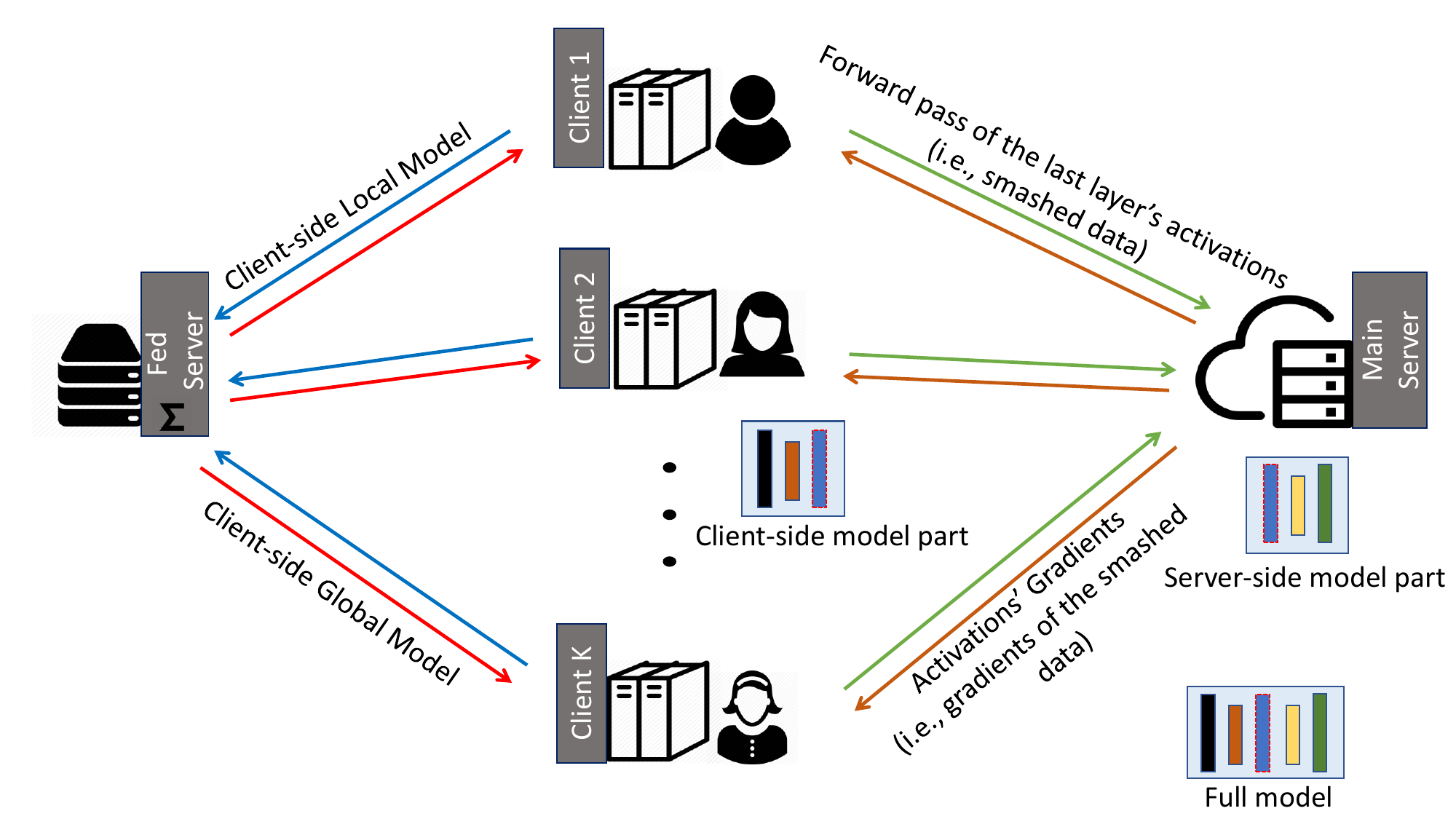}
		\caption{Overview of splitfed learning (SFL) system model.}
		\label{fig:outsystem}
		\vskip-2pt
	\end{figure}

	%================================================================================
% 	\begin{figure}[t]
% 	    \centering
% 		\includegraphics[width=0.9\linewidth]{fig/splitfed_new.pdf}
% 		\caption{Overview of splitfed learning system model with label sharing.}
% 		\label{fig:oursystem}
% 		\vspace{-0.3cm}
% 	\end{figure}

% \begin{figure} [!tbh]
% 	    \centering
% 		\includegraphics[width=0.9\linewidth]{fig/Usplitfed_new.pdf}
% 		\caption{Overview of splitfed learning system model without label sharing. In this example, each client has two different portions; the part1 (the initial layers) and the part2 (the final layers) of the full model.}
% 		\label{fig:oursystem2}
% 		\vspace{-0.3cm}
% 	\end{figure}
%============================================================================
%                               OUR APPROACH
%============================================================================
	\section{The Proposed Framework}\label{sec:splitfedlearning}
	
	The framework SFL is presented in this section. We first give the overview of SFL. Then we detail three key modules: (1) the differentially private knowledge perturbation, (2) the PixelDP for robust learning, and (3) total cost analysis of SFL.

	%In this section, we present splitfed learning that utilizes the strengths of SL and FL. 

	\subsection{Overall Structure}
	
	SFL combines the primary strength of FL, which is parallel processing among distributed clients, and the primary strength of SL, which is network splitting into client-side and server-side sub-networks during training. Refer to Fig.~\ref{fig:outsystem} for a representation of the SFL architecture. Unlike SL, all clients carry out their computations in parallel and engage with the main server and fed server. A client can be a hospital or an Internet of Medical Things with low computing resources, and the main server can be a cloud server or a researcher with high-performance computing resources.  
	The fed server is introduced to conduct FedAvg on the client-side local updates. Moreover, the fed server synchronizes the client-side global model in each round of network training. The fed server's computations, which is mainly computing FedAvg, are not costly. Hence, the fed server can be hosted within the local edge boundaries. Alternatively, if we implement all operations at the fed server over encrypted information, i.e., homomorphic encryption-based client-side model aggregation, then the main server can perform the operations of the fed server.
	%the clients and the main server perform the network training. All clients carry out their computations in parallel and engages with the main server. Then the main server, which is assumed to have sufficient computing resources ({\em e.g.,} cloud server and researchers with high-performance computing resources), process the forward propagation and back-propagation on its server-side model with each client's smashed data separately in (somewhat) parallel.  

	\textbf{SFL workflow.}  All clients perform forward propagation on their client-side model in parallel, including its noise layer, and pass their smashed data to the main server. Then the main server processes the forward propagation and back-propagation on its server-side model with each client's smashed data separately in (somewhat) parallel. It then sends the gradients of the smashed data to the respective clients for their back-propagation. Afterward, the server updates its model by FedAvg, i.e., weighted averaging of gradients that it computes during the back-propagation on each client's smashed data.  At the client's side, after receiving the gradients of its smashed data, each client performs the back-propagation on their client-side local model and computes its gradients. A DP mechanism is used to make these gradients private and send them to the fed server. The fed server conducts the FedAvg of the client-side local updates and sends them back to all participating clients. 

	\setlength{\textfloatsep}{9pt}% Remove \textfloatsep
	\begin{algorithm} [tb] 
	    \footnotesize
		\setstretch{0.52}
		%\small
		\SetNoFillComment
		\caption{Splitfed Learning (SFL)}
		
		%------------------------------------------------------------------------------------------------------
		\SetAlgoNoLine
		
    	   {\textbf{Notations:} (1) $S_t$ is a set of $K$ clients at $t$ time instance, (2) $ \mathbf{A}_{k,t} $ is the smashed data of client $k$ at $t$, (3) $\mathbf{Y}_{k}$ and $\hat{\mathbf{Y}}_{k}$ are the true and predicted labels, respectively, of the client $k$, (4) $\triangledown\ell_k$ is the gradient of the loss for the client $k$, (5) $n$ and $n_k$ are the total sample size and sample size at a client $k$, respectively.}

		\vspace{5pt}
		\SetKwProg{Fn}{EnsureMainServer executes:} {} {} 
		\tcc{	\scriptsize Runs on Main Server}
		\Fn{}{
		\eIf {\textup{time instance} t=0}{
		Initialize $\mathbf{W}^{\textup{S}}_{t}$ (global server-side model)}{
		\For{\textup{each client $ k\in S_t $ in parallel}} {
		    \While{local epoch $e \neq E$}{      
				$ (\mathbf{A}_{k,t}, \mathbf{Y}_{k}) \leftarrow$ ClientUpdate$(\mathbf{W}^{\textup{C}}_{k,t})$ \\
				%\begin{minipage}{6.5cm}
				Forward propagation with $ \mathbf{A}_{k,t} $ on $ \mathbf{W}^{\textup{S}}_t$, compute $ \hat{\mathbf{Y}}_{k} $  \\
				%\end{minipage}
				
				Loss calculation with $ \mathbf{Y}_{k}$ and  $\hat{\mathbf{Y}}_{k}$ \\
				
				Back-propagation calculate $\triangledown \ell_k (\mathbf{W}^{\textup{S}}_t; \mathbf{A}^{\textup{S}}_t)$\\
				Send $ d\mathbf{A}_{k,t} := \triangledown\ell_k (\mathbf{A}^{\textup{S}}_t; \mathbf{W}^{\textup{S}}_t) $ (i.e., gradient of the $ \mathbf{A}_{k,t} $) to client $ k $ for ClientBackprop($d\mathbf{A}_{k,t})$\\
			}}
			Server-side model update: $ \mathbf{W}^{\textup{S}}_{t+1
			}  \leftarrow \mathbf{W}^{\textup{S}}_t - \eta\frac{n_k}{n} \sum_{i=1}^{K} \triangledown \ell_i (\mathbf{W}^{\textup{S}}_t;\mathbf{A}^{\textup{S}}_t)$\\}}

		\vspace{5pt}
		\SetKwProg{Fn}{EnsureFedServer executes:}{}{}	 \tcc{	\scriptsize Runs on Fed Server}	
		\Fn{}{
		    \eIf {t=0}{
		Initialize $\mathbf{W}^{\textup{C}}_{t}$ (global client-side model)\\
		Send $\mathbf{W}^{\textup{C}}_{t}$  to all $K $ clients for ClientUpdate$  (\mathbf{W}^{\textup{C}}_{k,t})$
		}{
		    \For{\textup{each client $ k \in  S_t$ in parallel}}{
				$ \mathbf{W}^{\textup{C}}_{k,t} \leftarrow$ ClientBackprop($ d\mathbf{A}_{k,t} $) 
			}
			Client-side global model updates: $ \mathbf{W}^{\textup{C}}_{t+1}  \leftarrow \sum_{k=1}^{K} \frac{n_k}{n}  \mathbf{W}^{\textup{C}}_{k,t}$
			
			Send $\mathbf{W}^{\textup{C}}_{t+1} $  to all $K $ clients for ClientUpdate$  (\mathbf{W}^{\textup{C}}_{k,t}) $}}
		\label{algo3a}
	\end{algorithm}

   	\subsubsection{Variants of Splitfed Learning.} There can be several variants of SFL. We broadly divide them into two categories in the following:
   \vspace{-0.3cm}
	\paragraph{\textbf{Based on Server-side Aggregation.}} This paper proposes two variants of SFL. The first one is called \emph{splitfedv1 (SFLV1)}, which is depicted in Algorithm~\ref{algo3a} and~\ref{algo3b}. 
	%In SFLV1, all the client-side local models are aggregated at the fed server and the server-side models are aggregated at the main server at the end of each global epoch for the model synchronization.  
	%
	The next algorithm is called \emph{splitfedv2 (SFLV2)}, and it is motivated by the intuition of the possibility to increase the model accuracy by removing the model aggregation part in the server-side computation module in Algorithm~\ref{algo3a}. In Algorithm~\ref{algo3a}, the server-side models of all clients are executed separately in parallel and then aggregated to obtain the global server-side model at each global epoch. In contrast, SFLV2 processes the forward-backward propagations of the server-side model sequentially with respect to the client's smashed data (no FedAvg of the server-side models). The client order is chosen randomly in the server-side operations, and the model gets updated in every single forward-backward propagation. Besides, the server receives the smashed data from all participating clients synchronously. The client-side operation remains the same as in the SFLV1; the fed server conducts the FedAvg of the client-side local models and sends the aggregated model back to all participating clients. These operations are not affected by the client order as the local client-side models are aggregated by the weighted averaging method, i.e., FedAvg. Some other SFL versions are available in the literature, but they are developed after and influenced by our approach~\cite{splitfed_improved,our_latest}. 
	
	\paragraph{\textbf{Based on Data Label Sharing.}} Due to the split ML models in SFL, we can carry out ML in the two settings; (1) sharing the data labels to the server and (2) without sharing any data labels to the server. 
	Algorithm~\ref{algo3a} considers SFL with data label sharing. 
	In cases without sharing data labels, the ML model in SFL can be partitioned into three parts, assuming a simple setup. Each client will process two client-side model portions; one with the first few layers of $\mathbf{W}$, and another with the last few layers of $\mathbf{W}$ and loss calculations. The remaining middle layers of $\mathbf{W}$ will be computed at the server-side.   
	All possible configurations of SL, including vertically partitioned data, extended vanilla, and multi-task SL~ \cite{splitlearning}, can be carried out similarly in SFL as its variants.
    
    \subsection{Privacy Protection}
    \label{sflprivacy}
    The inherent privacy preservation capabilities of SFL are due to two reasons: firstly, it adopts the model-to-data approach, and secondly, SFL conducts ML over a split network.     
    A network split in ML learning enables the clients/fed server and the main server to maintain the full model privacy by not allowing the main server to get the client-side model updates and vice-versa. The main server has access only to the smashed data (i.e., activation vectors of the cut layer). The curious main server needs to invert all the client-side model parameters, i.e., weight vectors, to infer data and client-side model. The possibility of inferring the client-side model parameters and raw data is highly unlikely if we configure the client-side ML networks' fully connected layers with sufficiently large numbers of nodes~\cite{SplitNN}. 
    %The possibility of inferring the client-side model parameters to obtain raw data is highly unlikely if we configure the client-side ML networks' fully connected layers with sufficiently large numbers of nodes~\cite{SplitNN}.
    However, for a smaller client-side network, the possibility of this issue can be high. This issue can be controlled by modifying the loss function at the client-side~\cite{splitlearning3}. Due to the same reasons, the clients (having access only to the gradients of the smashed data from the main server) and the fed server (having access only to the client-side updates) cannot infer the server-side model parameters. 
    %\textcolor{red}{The possible client-side leakage (i) between the clients through the fed server, and (ii) fed server and the clients is maintained at low probability by applying differential privacy.} 
    %
    Since there is no network split and separate training on the client-side and server-side in FL, SFL provides superior architectural configurations for enhanced privacy for an ML model during training compared to FL.
  
%====================================================================
\subsubsection{Privacy Protection at the Client-side.}  
\label{sec:diffprivtheory}
%\textcolor{red}{noise is not added to the lables if transmitted to the server: privacy analysis}
%While providing strict privacy guarantees, DP can affect the utility of an ML model. Hence, 
%In this section, we investigate the impact of strict privacy configurations with differential privacy to evaluate the performance of splitfed further. In splitfed, clients communicate with two servers: (1) main server and (2) fed server. The clients share their smashed data (activations from their cut layer) to the main server, whereas the fed server aggregates the client-side model portion. During both of these communications, the clients do not share their raw data with these two servers or other clients. 

We discuss the inherent privacy of the proposed model in the previous section. However, there can be an advanced adversary exploiting the underlying information representations of the shared smashed data or parameters (weights) to violate data owners' privacy. This can happen if any server/client becomes curious though still honest. To avoid these possibilities, we apply two measures in our studies; (i) differential privacy to the client-side model training and (ii) PixelDP noise layer in the client-side model.
%
%To enable data privacy while client-side model aggregation and synchronization, we apply differential privacy to the client-side model training algorithm based on the differentially private deep learning approach.
%developed by Abadi et al.~\cite{abadi2016deep}. As discussed in Section \ref{sec:splitfedlearning}, the clients and the main server collaboratively train the client-side model portion and the server-side model portion separately while training one whole model that is split between the clients and the main server. Thus, the application of the differential privacy on the client-side model guarantees a differentially private client-side model training that is independent of the main server-side model training.  

	 \begin{algorithm} [tb] 
	    \footnotesize
		\setstretch{0.52}
		%\small
		\SetNoFillComment
		\caption{ClientUpdate}
		
		%------------------------------------------------------------------------------------------------------
		\SetAlgoNoLine
		\SetKwProg{Fn}{EnsureClientUpdate$  (\mathbf{W}^{\textup{C}}_{k,t}) $:}{}{} \tcc{	\scriptsize Runs on Client $ k $}	
		\Fn{}{
			Model updates $ \mathbf{W}^{\textup{C}}_{k,t}\leftarrow$ FedServer() \\
			Set $ \mathbf{A}_{k,t} $ = $ \phi $ \\
			\For{\textup{each local epoch $ e$ from $ 1 $ to $E$}}{
				Forward propagation with data $X_k$ up to a layer $L\geq 1$ in $ \mathbf{W}^{\textup{C}}_{k,t}$\\
				\emph{Noise layer:} Perturbs the outputs of the layer $L$ based on Equation~\eqref{noiseq1}\\ 
				
				With the output from the noise layer, continue forward propagation to the remaining layers of $\mathbf{W}^{\textup{C}}_{k,t}$, and get the activations of its final layer $ \mathbf{A}_{k,t} $ (smashed data)\\
				$ \mathbf{Y}_{k}$ is the true labels of $X_k$\\ 
				Send $ \mathbf{A}_{k,t} $ and $\mathbf{Y}_{k}$ to the main server\\
				Wait for the completion of ClientBackprop($d\mathbf{A}_{k,t}$)\\
			}
			}	
		
		\vspace{5pt}
		\SetKwProg{Fn}{EnsureClientBackprop$  (d\mathbf{A}_{k,t}) $:} {} {}
		\tcc{	\scriptsize Runs on Client $ k $}
		\Fn{}{
			%\tcp{Run on client $ k $}
			%Set $ \triangledown  \ell_k (\mathbf{W}^{\textup{C}}_{k,t}) = \phi$\\
			\While{local epoch $e \neq E$}{ 
			    $d\mathbf{A}_{k,t} \leftarrow$ MainServer() \\
				Back-propagation, calculate gradients $ \triangledown  \ell_k (\mathbf{W}^{\textup{C}}_{k,t})$ with $d\mathbf{A}_{k,t}$ \\
			$\ell_{2}$-norm of each gradient is clipped and a calibrated noise is added based on Equation~\eqref{dippeq2} and \eqref{dippeq3} to calculate $\tilde{\mathbf{g}}_{k,t}$\\
				
				Update $\mathbf{W}^{\textup{C}}_{k,t}\leftarrow \mathbf{W}^{\textup{C}}_{k,t}-\eta \tilde{\mathbf{g}}_{k,t}$	
			}
			Send $ \mathbf{W}^{\textup{C}}_{k,t} $ to the fed server}
		\label{algo3b}
	\end{algorithm}

\subsubsection{Privacy Protection on Fed Server.}
Considering Algorithm~\ref{algo3b}, we present the process for implementing differential privacy at a client $k$. We assume the following: $\sigma$ represents the noise scale, and $C'$ represents the gradient norm bound. Now, firstly, after $t$ time, the client $k$ receives the gradients $d\mathbf{A}_{k,t}$ from the server, and with this, it calculates client-side gradients $\triangledown  \ell_k (\mathbf{W}^{\textup{C}}_{k,i,t})$ for each of its local sample $x_i$, and %$\mathbf{g}_{k,t}\left(x_{i}\right) \leftarrow \triangledown  \ell_k (\mathbf{W}^{\textup{C}}_{k,i,t})$.
\begin{equation}
\mathbf{g}_{k,t}\left(x_{i}\right) \leftarrow \triangledown  \ell_k (\mathbf{W}^{\textup{C}}_{k,i,t}).
\label{dippeq1}
\end{equation}

Secondly, the $\ell_{2}$-norm of each gradient is clipped according to the following equation: 
\begin{equation}
\overline{\mathbf{g}}_{k,t}\left(x_{i}\right) \leftarrow  \mathbf{g}_{k,t}\left(x_{i}\right) / \max \left(1, \frac{\left\|\mathbf{g}_{k,t}\left(x_{i}\right)\right\|_{2}}{C'}\right).
\label{dippeq2}
\end{equation}

Thirdly, calibrated noise is added to the average gradient:
\begin{equation}
\tilde{\mathbf{g}}_{k,t} \leftarrow \frac{1}{n_k} \sum_{i}\left(\overline{\mathbf{g}}_{k,t}\left(x_{i}\right)+\mathcal{N}\left(0, \sigma^{2} C'^{2} \mathbf{I}\right)\right).
\label{dippeq3}
\end{equation}

Finally, the client-side model parameters of client $k$ are updated as follows; $\mathbf{W}^{\textup{C}}_{k,t+1} \leftarrow \mathbf{W}^{\textup{C}}_{k,t}-\eta_{t} \tilde{\mathbf{g}}_{k,t}$.
% \begin{equation}
% \mathbf{W}^{\textup{C}}_{k,t+1} \leftarrow \mathbf{W}^{\textup{C}}_{k,t}-\eta_{t} \tilde{\mathbf{g}}_{k,t}.
% \label{dippeq4}
% \end{equation}

%Assume that, $\theta$ represents model parameters, $\left\{x_{1}, \ldots, x_{N}\right\}$ represents the examples, $\mathcal{L}(\theta)=\frac{1}{N} \sum_{i} \mathcal{L}\left(\theta, x_{i}\right)$ represents the loss function, $\eta_{t}$ represents the learning rate, $\sigma$ represents the noise scale, and $C$ represents the gradient norm bound. Then, at each step of training, the gradients $(\nabla_{\theta_{t}} \mathcal{L}\left(\theta_{t}, x_{i}\right))$ are computed for a randomly chosen subset of examples, as shown in Eq. \ref{dippeq1}. Next, the $\ell_{2}$ norm of each gradient is clipped according to Eq. \ref{dippeq2}. After that, calibrated noise is added to the average gradient, as shown in Eq. \ref{dippeq3}, and take a step in the opposite direction, as shown in Eq. \ref{dippeq4}.

We apply calibrated noise iteratively until the model converges or reaches a specified number of iterations. As the iterations progress, the final convergence will exhibit a privacy level of $(\varepsilon, \delta)$- differential privacy, where $(\varepsilon, \delta)$ is the overall privacy cost of the client-side model. 

Differential privacy is used to enforce strict privacy to the client-side model training algorithm based on Abadi et al.'s approach ~\cite{abadi2016deep}. %Hence, our approach follows the privacy guarantees discussed in ~\cite{abadi2016deep}. 
Equation \ref{dippeq2} (norm clipping) guarantees that $\left\|\mathbf{g}_{k,t}\left(x_{i}\right)\right\|_{2}$ is preserved when  $\left\|\mathbf{g}_{k,t}\left(x_{i}\right)\right\|_{2} \leq C'$. This step also guarantees that $\left\|\mathbf{g}_{k,t}\left(x_{i}\right)\right\|_{2}$ scaled down to $C'$ when $\left\|\mathbf{g}_{k,t}\left(x_{i}\right)\right\|_{2} > C'$. This step also helps clipping out the effect of Equation \ref{noiseq1} on the gradients. Hence, norm clipping step allows bounding the influence of each individual example on $\mathbf{g}_{k,t}$ in the process of guaranteeing differential privacy. It was shown that, the corresponding noise addition (refer to Equation \ref{dippeq3}) provides $(\epsilon,\delta)$-{DP} for each step of $b$ ($b = n_k/batch\_size$), if we choose $\sigma$ (noise scale) to be $\sqrt{2 \log \frac{1.25}{\delta}} / \varepsilon$~\cite{dwork2014algorithmic}. Hence, at the end of $b$ steps, this will result in $(b\epsilon,b\delta)$-{DP}. As shown by Abadi et al., for any $\varepsilon<c_{1} b^{2} T$ and $\delta >0$, by choosing $\sigma \geq c_{2} \frac{b \sqrt{T \log (1 / \delta)}}{\varepsilon}$, the privacy can be maintained at $(\epsilon,\delta)$-{DP}~\cite{abadi2016deep}. Moments accountant (a privacy accountant) is used to track and maintain $(\epsilon,\delta)$. Hence, at the end of $b$, a client model guarantees $(\epsilon,\delta)$-{DP}. With the strict assumption that all clients work on IID data, we can confirm that all clients maintain and guarantee $(\epsilon, \delta)$-{DP} while client-side model training and synchronization.

\setlength{\textfloatsep}{15pt}% Remove \textfloatsep
\subsubsection{Privacy Protection on Main Server.}
The above DP measures do not stop potential leakage from the smashed data to the main server though it has some effect on the smashed data after the first global epoch. Thus, to avoid privacy leakage and further strengthen data privacy and model robustness against potential adversarial ML settings, we integrate a noise layer in the client-side model based on the concepts of PixelDP~\cite{lecuyer2019certified}.
This extended measure utilizes the noise application mechanism involved in differential privacy to add a calibrated noise to the output ({\em e.g.,} activation vectors) of a layer at the client-side model while maintaining utility. In this process, firstly, we calculate the sensitivity of the process. The sensitivity of a function $\mathbf{A}$ is defined as the maximum change in output that can be produced by a change in the input, given some distance metrics for the input and output  (p-norm and q-norm, respectively):
\begin{equation}
    \Delta I_{p,q} = \Delta I^{\mathbf{A}}_{p,q} = \text{max}_{i,j,i \neq j} 
    \frac{\|\mathbf{A}_{k,i}-\text{min}\ \mathbf{A}_{k,j}\|_q}{\|x_i - x_k\|_p}
\end{equation}
Secondly, Laplacian noise with scale $\frac{\Delta I^{\mathbf{A}}_{p,q}}{\varepsilon'} $ is applied to randomize any data as follows:
\begin{equation}
\mathbf{A}^\textup{P}_{k,i}= \mathbf{A}_{k,i}+\operatorname{Lap}\left(\frac{\Delta I^A_{p,q}}{\varepsilon'}\right),
\label{noiseq1}
\end{equation}
where, $\mathbf{A}^\textup{P}_{k,i}$ represents a private version of $\mathbf{A}_{k,i}$, and $\epsilon'$ is the privacy budget used for the Laplacian noise. This method enables forwarding private versions of the smashed data to the main server; hence, preserving the privacy of smashed data. The private version of the smashed data is due to the post-processing immunity of the DP mechanism applied at the noise layer in the client-side model.  
%Empirical results are presented in Section~\ref{sec:diffprivacy}. 
The noisy smashed data is more private than the original data due to the calibrated noise. Moreover, PixelDP not only can provide privacy for smashed data, but also can improve the robustness of the model against adversarial examples.  However, detailed analysis and mathematical guarantees are kept for future work to preserve the main focus of the proposed work.

\paragraph{Robustness via PixelDP.} The primary intuition behind using random DP mechanism to robust ML against adversarial examples is to create a DP scoring function.
For example, feeding any data sample through the DP scoring function, the outputs are DP with regards to the features of the input.
Then, stability bounds for the expected output of the DP function are given by the following Lemma \cite{lecuyer2019certified}:
\begin{lemma}
    Suppose a randomized function $\mathcal{M}$, with bounded output $\mathcal{M} \in [0, b], b \in \mathbb{R}^+$, satisfies ($\epsilon, \delta$)-DP. Then the expected value of its output meets the following property:
    \begin{equation}
        \forall \alpha \in B_p(1). \mathbb{E}(\mathcal{M}(x)) \leq e^{\epsilon}  \cdot \mathbb{E}(\mathcal{M}(x+\alpha)) + b\delta,
    \end{equation}
    where $B_p(r):= {\alpha \in \mathbb{R}^n:\|\alpha\|_p \leq r}$ is the $p$-norm ball, and the expectation is taken over the randomness in $\mathcal{M}$.
\end{lemma}
Combined with Equation, $\forall \alpha \in B_p(L), k=f(x). \; y_k(x + \alpha) > \max_{i:i\neq k} y_i(x + \alpha)$, the bounds provide a rigorous certification for robustness to adversarial examples.

%============================================================================
%                               COST ANALYSIS
%============================================================================	
\subsection{Total Cost Analysis}
\label{sec:costanalysis}
This section analyzes the total communication cost and model training time for FL, SL, and SFL under a uniform data distribution.
Assume $K$ be the number of clients, $p$ be the total data size, $q$ be the size of the smashed layer, $R$ be the communication rate, $T$ be the time taken for one forward and backward propagation on the full model with dataset of size $p$ (for any architecture), $T_{\textup{fedavg}}$ is the time required to perform the full model aggregation (let $\frac{T_{\textup{fedavg}}}{2}$ be the aggregation time for an individual server), $|\mathbf{W}|$ be the size of the full model, and $\beta$ be the fraction of the full model's size available in a client in SL/SFL, i.e., $|\mathbf{W}^{\textup{C}}| = \beta |\mathbf{W}|$. The term $ 2\beta |\mathbf{W}|$ in communication per client is due to the download and upload of the client-side model updates before and after training, respectively, by a client. The result is presented in Table~\ref{totcomp}.  
As shown in the table, SL can become inefficient when there is a large number of clients. Besides, we see that when $K$ increases, the total training time cost increases in the order of SFLV2$<$SFLV1$<$SL. Also, we observe this in our empirical results.
%In FL, the cost of the (federation) server per each global epoch equals to $K|W|$.

\begin{table}[!t]
    \centering
    \caption{Total cost analysis of the four DCML approaches for one global epoch.}
    \begin{adjustbox}{max width=0.5\linewidth}
    \begin{tabular}{cccc}
    \toprule
    Method                           & Comms. per client                   & Total comms.      & Total model training time                                                                      \\ \midrule
    FL         & $2 |\mathbf{W}|$                  & $2 K |\mathbf{W}|$      & {$T+ 2\frac{|\mathbf{W}|}{R}+ T_{\textup{fedavg}}$}                                           \\ \midrule
    
    SL         & $(\frac{2p}{K})q+2 \beta |\mathbf{W}|$     & $2pq+2\beta K |\mathbf{W}|$      & $T + 2\frac{pq}{R}+ 2\beta \frac{|\mathbf{W}|}{R}K $                                     \\ \midrule
    
   SFLV1 & $(\frac{2p}{K})q+2 \beta |\mathbf{W}|$ & $2pq + 2\beta K |\mathbf{W}|$ & $T + 2 \frac{pq}{KR} + 2 \frac{\beta |\mathbf{W}|}{R} + T_{\textup{fedavg}}$ \\ \midrule
    
    SFLV2                       & $(\frac{2p}{K})q+2 \beta |\mathbf{W}|$                      & $2pq + 2\beta K |\mathbf{W}|$                      & $T + 2 \frac{pq}{KR} + 2 \frac{\beta |\mathbf{W}|}{R} + \frac{T_{\textup{fedavg}}}{2}$                                                            \\ \bottomrule
    \end{tabular}
    \end{adjustbox}
    \label{totcomp}
\end{table}

%============================================================================
%                              EXPERIMENTAL SETUP
%============================================================================	
	\section{Experiments}
	Experiments are carried out on uniformly distributed and horizontally partitioned image datasets among clients. All programs are written in python 3.7.2 using the PyTorch library (PyTorch 1.2.0). For quicker experiments and developments, we use the High-Performance Computing (HPC) platform that is built on Dell EMC's PowerEdge platform with partner GPUs for computation and InfiniBand networking. We run clients and servers on different computing nodes of the cluster provided by HPC. We request the following resources for one slurm job on HPC: 10GB of RAM, one GPU (Tesla P100-SXM2-16GB), one computing node with at most one task per node. The architecture of the nodes is x86\_64. 
	%During the experiments, we investigate the performance by observing the training time (also called latency) with respect to global epochs, and the amount of data communication by each client. 
	In our setup, we consider that all participants update the model in each global epoch (i.e., $C =1 $ during training). We choose ML network architectures and datasets based on their performance and their need to include proportionate participation in our studies. The learning rate for LeNet is maintained at 0.004 and 0.0001 for the remainder of network architectures (AlexNet, ResNet, and VGG16). We choose the learning rate based on the models' performance during our initial observations. For example, for LeNet on FMNIST, we observed train and test accuracy of 94.8\% and 92.1\% with a learning rate of 0.004, whereas 87.8\% and 87.3\% with a learning rate of 0.0001 in 200 global epochs. We set up a similar computing environment for comparative analysis. 
	%We do not conduct hyperparameter tunning or any other model optimization scenarios, as our work is on the privacy-preserving DCML approaches rather than the optimization of the individual ML network.   
    
 %=======================================================================	  
	%\subsection{Datasets and Model Architectures}	
	
	We use four public image datasets in our experiments, and these are summarized in Table~\ref{table:dataset}. HAM10000 dataset is a medical dataset, i.e., the Human Against Machine with 10000 training images~\cite{hamdata}. It consists of colored images of pigmented skin lesions, and has dermatoscopic images from different populations, acquired and stored by different modalities. 
	%Each sample has $ 600 \times 450  $ pixel images. HAM10000 has a total of $10,015$ samples with 
	It has seven cases of important diagnostic categories of lesions: Akiec, bcc, bkl, df, mel, nv, and vasc. MNIST, Fashion MNIST, and CIFAR10 are standard datasets, all with 10 classes. 
	%
	%MNIST dataset consists of handwritten digit images of $ 0 $ to $ 9 $ (i.e., $ 10 $ classes). Fashion MNIST consists of ten clothing images, including T-shirts, trousers, pullover, dress, and coat. CIFAR10 consists of color images of ten objects (classes), including airplane, cat, dog, bird, automobile, horse, and ship. 
	%
	%For our experiments, we consider the training and testing sample sizes described in Table~\ref{table:dataset}. 
	%
	%In our DCML setup, the dataset is randomly, disjointly, and uniformly distributed among clients.
%MNIST dataset~\cite{MNIST}
%Fashion MNIST~\cite{FMNIST}
%CIFAR10~\cite{cifar10}
	%-----------------------------------------------------------
    \begin{table}[t]
    \centering
		\scriptsize
		\caption{Datasets}
		\label{table:dataset}
		\begin{adjustbox}{max width=\linewidth}
		\begin{tabular}{cccc}
			\toprule
			Dataset		& Training samples			& Testing samples & Image size \\
			\midrule
			
			HAM10000~\cite{hamdata} 	&  $ 9013 $ 		&$ 1002 $  & $ 600\times 450 $   \\
			\hline	
			MNIST~\cite{MNIST}  					& $ 60,000 $ 	& $ 10,000 $ & $ 28\times 28 $  \\
			\hline
			FMNIST~\cite{FMNIST}				&  $ 60,000 $ 		& $ 10,000 $ & $ 28\times 28  $ \\
			\hline
			CIFAR10~\cite{cifar10}					&  $50,000$ 		& $ 10,000 $  & $ 32\times 32  $ \\
			\bottomrule
		\end{tabular}
		\end{adjustbox}
	\end{table}
%=======================================================================	
	%\subsection{Model Architecture}
	%
	In regard to ML models, we consider four popular architectures in our experiments. These four architectures fall under Convolutional Neural Network (CNN) architectures and are summarized in Table~\ref{table:Model}. We restrict our experiments to CNN architectures to maintain the cohesiveness of our work proposed in this paper. We will conduct further experimental evaluations on other architectures such as recurrent neural networks in future work. 
	%LeNet~\cite{lenet} is a five-layer CNN consisting of convolution, average pooling, Sigmoid or Tanh, and fully connected layers. It uses the $ 5 \times 5 $ and $ 2\times 2 $ sized kernels in its layers. The input image dimension is $ 32\times 32\times 1 $. 
	%AlexNet~\cite{alexnet} consists of eight network layers, including convolution (five layers), pooling (three layers), and fully connected (two) layers. It uses $ 11\times 11 $, $5\times 5$, and $ 3\times 3$ sized kernels in its layers. The input image dimension is $ 227\times 227\times 3 $.
	%VGG16~\cite{vgg16} consists of sixteen network layers, including convolution (sixteen layers), max pooling, and fully connected layers. It uses $ 3\times 3 $ sized kernels in its layers. The input image dimension is $ 224\times 224\times 3 $.
	%ResNet18~\cite{resnet} consists of eighteen network layers, including convolution, max pooling, ReLU, and fully connected layers. It uses $ 7\times 7 $ and $ 3\times 3 $ sized kernels in its layers. 

	For all experiments in SL, SFLV1, and SFLV2, the network layers are split at the following layer: second layer of LeNet (after 2D MaxPool layer), second layer of AlexNet (after 2D MaxPool layer), fourth layer of VGG16 (after 2D MaxPool layer), and third layer (after 2D BatchNormalization layer) of ResNet18. For a fair comparison, while performing the comparative evaluations of SFLV1 and SFLV2 with FL and SL, we do not consider the addition of differential privacy-based measures and PixelDP in SFLV1 and SFLV2.

%============================================================================
%                               RESULTS AND DISCUSSION
%============================================================================	
% 	\subsection{Results and Discussion}
% 	\label{results}
	
	%\THA{Add standard deviation of the results among the clients not only mean for all experiments --  maybe in one table.}
	%\THA{make the best result in bold in the tables}
	
	\subsection{Performance of FL, SL, SFLV1 and SFLV2}
	
	We consider the results under normal learning (centralized learning) as our benchmark. 
	%In standard learning, all data are available centrally to a server that performs ML training and testing. 
	Table~\ref{table:testaccuracy} summarizes our first result, where the observation window is 200 global epochs with one local epoch, batch size of 1024, and five clients for DCML. 
	The table shows the best accuracy observed in 200 global epochs. 
	%We picked the highest accuracy because the training/testing curves were almost smooth after some epoch (for example see Fig.~\ref{fig:resnetham}) in our experiments.
	%
	Moreover, the test accuracy is averaged over all clients in the DCML setup at each global epoch. 
	%The process of calculating the accuracy is the following: the model is trained with each client, then based on the model's predictions and true labels, the training accuracy is calculated. Afterward, the test accuracy is calculated on the test data. The performance observations ({\em e.g.,} calculating the average accuracy) are carried out in each global epoch, both in centralized and DCML setups.

    %-----------------------------------------------------------
	\begin{table}[t]
	\centering
		\scriptsize
		\caption{Model Architecture}
			\label{table:Model}
		\begin{adjustbox}{max width=0.6\textwidth}
		\begin{tabular}{c@{\hspace{0.5cm}}cccc}
			\toprule
			Architecture 	& No. of parameters & Layers & Kernel size \\
			\midrule
			LeNet~\cite{lenet} 					& $ 60 $ thousands  & $ 5 $ 		&  $ (5 \times 5) $, $ (2\times 2) $  \\
			\hline
			AlexNet~\cite{alexnet}				& $ 60 $ million 	& $ 8	 $ 	& $(11\times 11) $, $(5\times 5)$, $ (3\times 3)$ \\
			\hline
			VGG16~\cite{vgg16}					& $ 138 $ million	  &	$ 16 $		&  $ (3\times 3) $\\
			\hline		
			ResNet18~\cite{resnet}	 		& 11.7 million		& $ 18 $		 & $ (7\times 7) $, $( 3\times 3) $  \\
			\bottomrule
		\end{tabular}
		\end{adjustbox}
	\end{table}
    %-----------------------------------------------------------
	%====================================================================
% 	\begin{table}[t]
% 	\centering
% 		\scriptsize
% 		\caption{Training Results}
% 		\label{table:trainaccuracy}
% 		\begin{adjustbox}{max width=0.489\textwidth}
% 		\begin{tabular}{ccccccc}
% 			\toprule
% 			Dataset 		& Architecture			& Normal & Federated & Split & Splitfedv1 & Splitfedv2 \\
% 			\midrule
% 			HAM10000					&  ResNet18 	&99.8\%  &92.2\%  &99.5\% &90.2\% &99.3\% \\
% 			\hline
% 			HAM10000					&AlexNet  	&98.8\%  &74.2 \%  &70.3\% &69\% &72.1\% \\
% 			\hlineeffe
% 			FMNIST				&  LeNet 		& 95.1\%  &92.5 \% &90.6\% &89\% &89.6\% \\
% 			\hline
% 			FMNIST				&  AlexNet 		& 97.2\% &90.6 \% &83.9\% &84.9\% &79\% \\
% 			\hline
% 			CIFAR10				&  	LeNet	&  78.7\%  &70.5 \% &62.6\% &61.3\% &62\% \\
% 			%\hline	
% 			%CIFAR10				&  	VGG16	& 99.3\%   &65.2 \% &X &X &X \\
% 			\hline		
% 			MNIST	& AlexNet 		& 99.7\%   &98.8\% &93.7\% &95.7\% &89.8\% \\
% 			\hline		
% 			MNIST	&  ResNet18		& 99.9\%  &99.9\% &99.9\% &99.8\% &99.9\% \\
% 			\bottomrule
% 		\end{tabular}
% 		\end{adjustbox}
% 	\end{table}
%-------------------	

	\begin{table}[t]
	\centering
		\scriptsize
		\caption{Test Results (five clients for DCML)}
		\label{table:testaccuracy}
		\begin{adjustbox}{max width=0.9\textwidth}
		\begin{tabular}{ccccccc}
			\toprule
			Dataset 		& Architecture			& Normal &  FL & SL & SFLV1 & SFLV2 \\
			\midrule
			HAM10000					&  ResNet18 	&79.3\%  &77.5\%  &79.1\% &79\% &79.2\% \\
			\hline
			HAM10000					&AlexNet  	&80.1\%  &75 \%  &73.8\% &70.5\% &74.9\% \\
			\hline
			FMNIST				&  LeNet 		& 92.7\%  &91.9 \% &90.4\% &89.6\% &90.4\% \\
			\hline
			FMNIST				&  AlexNet 		& 90.5\% &89.7\% &84.7\% &86\% &81\% \\
			\hline
			CIFAR10				&  	LeNet	&  72.1\%  &69.4 \% &62.7\% &62.6\% &63.8\% \\
			%\hline	
			%CIFAR10				&  	VGG16	& 77.8\%   &66.6\% &X &X &X \\
			\hline		
			MNIST	& AlexNet 		& 98.8\%   &98.7 \% &95.1\% &96.9\% &92\% \\
			\hline		
			MNIST	&  ResNet18		& 99.3\%  &99.2 \% &99.2\% &99\% &99.2\% \\
			\bottomrule
		\end{tabular}
		\end{adjustbox}
	\end{table}

	% this figure has issue with compilation error maybe it is corrupted
% 	\begin{figure}[t!]
% 	\centering
% % 		\subfigure[]{
% % 			\includegraphics[width=0.6\columnwidth]{fig/resnetham_train}
% % 		}
% % 		\hskip1pt
% 		\subfigure[]{
% 			\includegraphics[width=0.6\columnwidth]{fig/resnetham_test}
% 		}
% 		\caption{Testing convergence of ResNet18 on HAM10000 under various learning with five clients.}
% 		\label{fig:resnetham}
% 		\vspace{0.5cm}
% 	\end{figure}

	 %\textbf{Analyzing Table~\ref{table:testaccuracy}}: 
	 As presented in Table~\ref{table:testaccuracy}, SL and SFL (both versions) performed well under the proposed experimental setup. However, we also observed that among DCML, FL shows better learning performance in most cases, possibly due to the FedAvg of the full models at each global epoch. 
	 %In all cases except ResNet18 on MNIST, the four approaches failed to achieve the benchmark results. 
	 Based on the results, we can observe that SFLV1 and SFLV2 have inherited the characteristics of SL. In a separate experiment, we noticed that VGG16 on CIFAR10 did not converge in SL, which was the same for both versions of splitfed learning, although there were around 66\% and 67\% of training and testing accuracies, respectively, for FL. We assume that this was because of the unavailability of certain other factors such as hyper-parameters tuning or change in data distribution or additional regularization terms in the loss function, which are beyond the scope of this paper. 
	 %Besides, in some cases, the training accuracy was less than the test accuracy. This may be due to the use of dropouts, the less diversity in test data, and the consideration of the highest accuracy within the observation window (i.e., 200 global epochs).
	
\begin{figure}[t]
	\centering
		\includegraphics[width=0.6\linewidth]{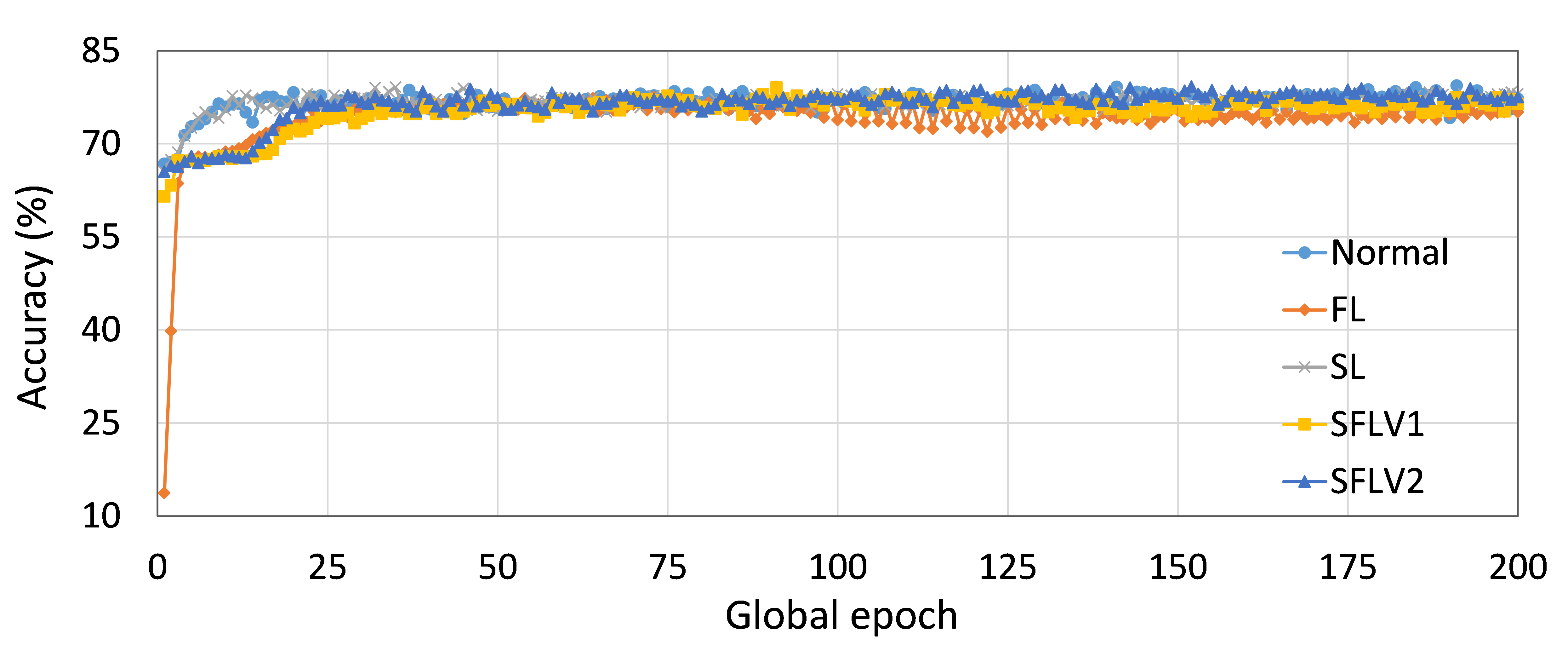}
		\caption{Testing convergence of ResNet18 on HAM10000 under various learning with five clients.}
		\label{fig:resnetham}
	\end{figure}

	Further diving into individual cases, as an example, we present the performance of ResNet18 on the HAM10000 dataset for normal (centralized learning), FL, SL, SFLV1, and SFLV2, under similar settings. %The training and testing performances of the remaining architectures and datasets are available in our long paper~\cite{samepaper}.  
	%%--------------------------------------------------
    % ham100000------------------------------------------
    %\textbf{Analyzing Fig.~\ref{fig:resnetham}}: 
    For ResNet18 on HAM10000, the test accuracy convergence was almost the same for FL, SL, SFLV1, and SFLV2, and they reached around 76\% in the observation window of 200 global epochs (refer to Fig.~\ref{fig:resnetham}). However, SFLV1 and SFLV2 struggled to converge if SL failed to converge. This was observed for the case of VGG16 on CIFAR10 in our separate experiments.
    %(refer to~\cite{samepaper}). 
    %During training, the training accuracy convergence of SFLV1, and FL training were close to each other but slower than SFLV2 and SL. The training accuracy reached around 90\% for them, whereas others got 99\% in the observation window of 200 global epochs. 
    
\begin{figure}[!t]
	\centering
		\subfigure[Train]{
			\includegraphics[width=0.27\linewidth]{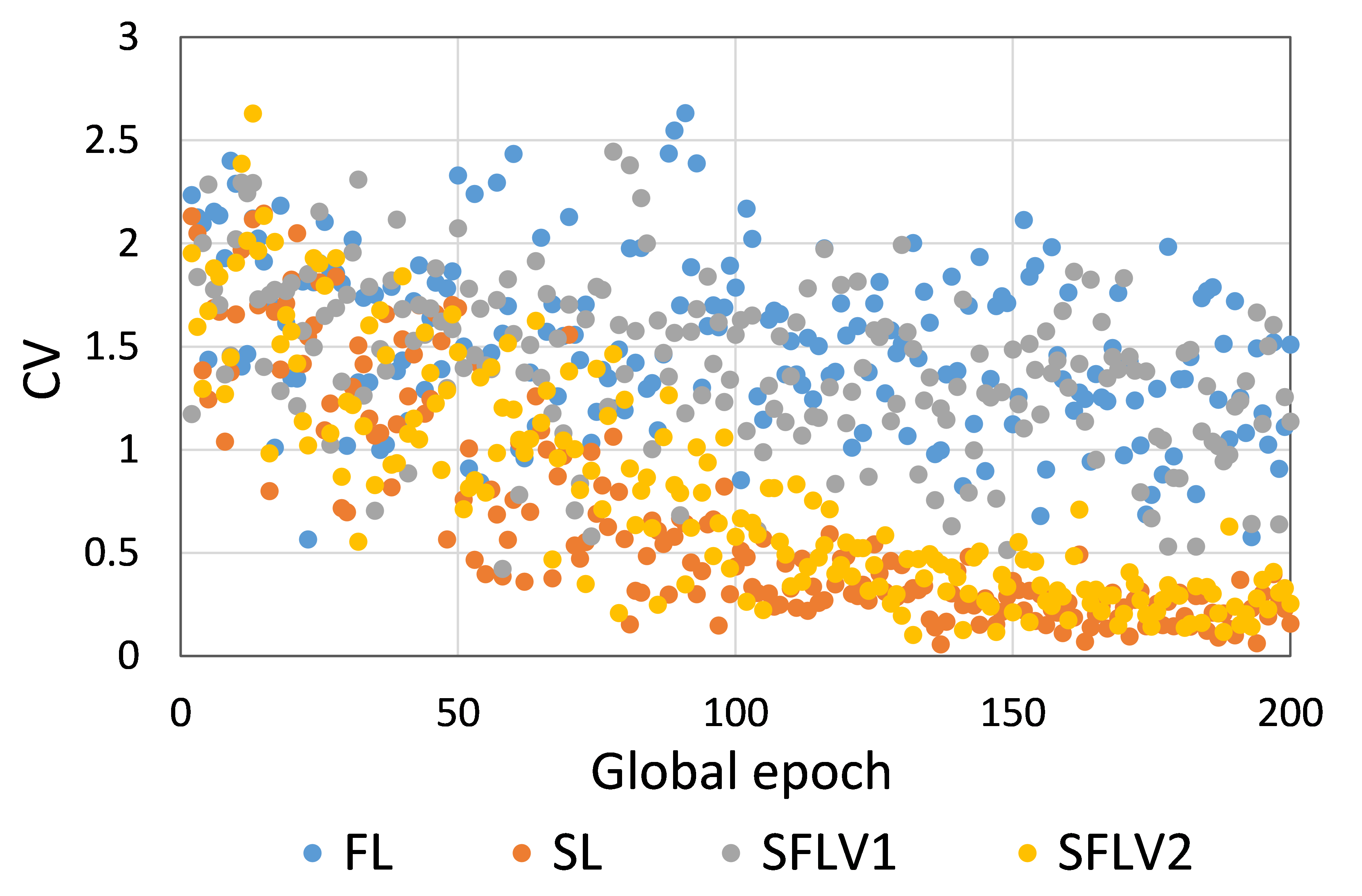}
		}
		\hskip-5pt
		\subfigure[Test]{
			\includegraphics[width=0.27\linewidth]{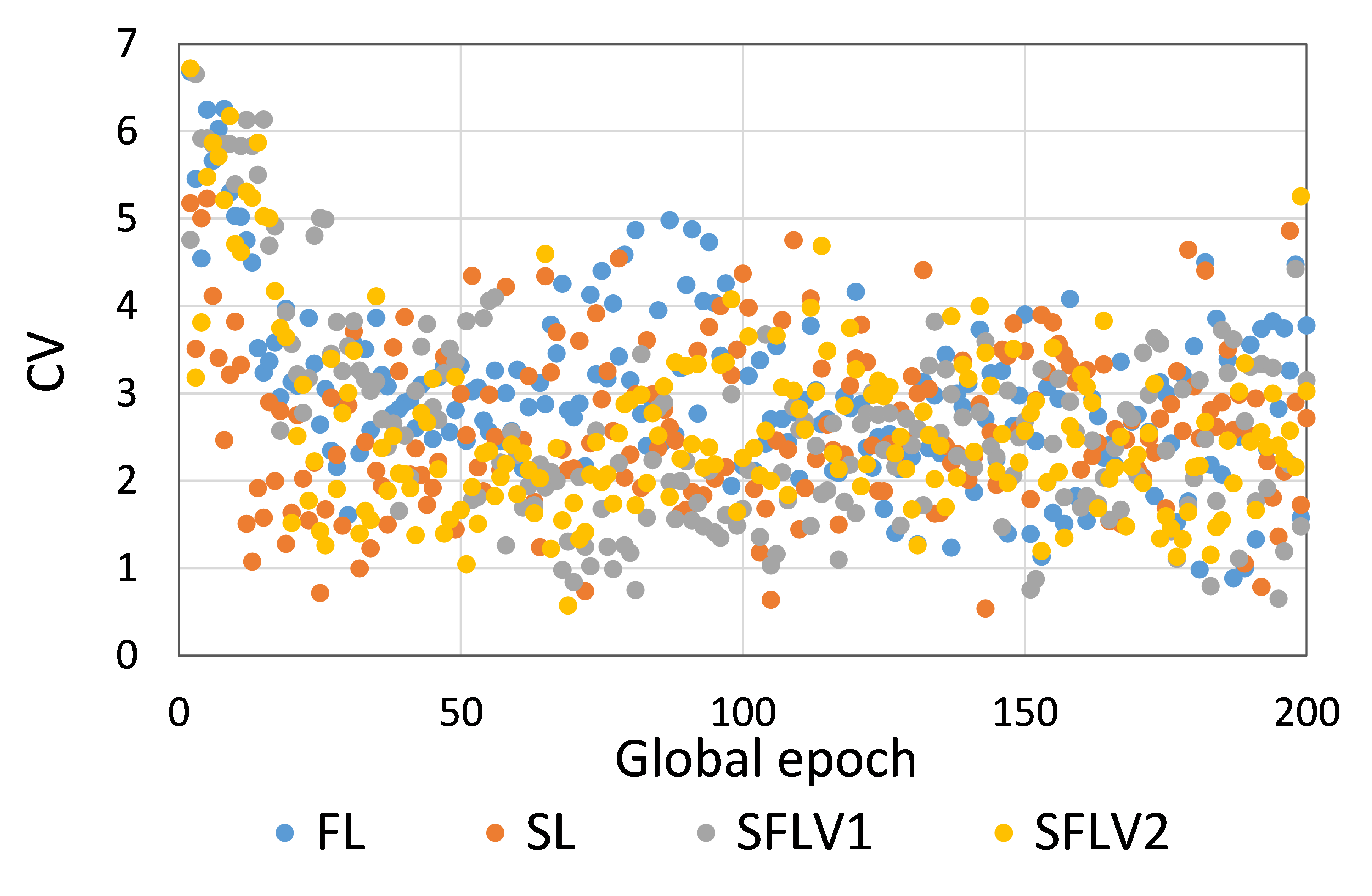}
		}
		\caption{Coefficient of variation (CV) of ResNet18 on HAM10000 under various learning settings with five clients.}
		\label{fig:resnetham_std}

	\end{figure}

    %\textbf{Analyzing Fig.~\ref{fig:resnetham_std}}: 
    So far, we considered the testing mean accuracy in our results. Fig.~\ref{fig:resnetham_std} illustrates the variations of the performance (i.e., accuracy) over five clients at each global epoch. In this regard, we compute the coefficient of variation (CV), which is a ratio of the standard deviation to the mean, and it measures the dispersion. Moreover, we calculate the CV over the five accuracies generated by the five clients at each global epoch. Based on our results for ResNet18 on HAM10000, the CVs for SL, FL, SFLV1, and SFLV2 are bounded between 0.06 and 2.63 while training, and 0.54 and 6.72 while testing after epoch 2; at epoch 1, the CV is slightly higher. The results indicate uniform individual client-level performance across the clients, as the CV coefficient values below 10 are considered a good range in literature.

    %--------------------------------------------------
	%-------------------------------------------------

    In some datasets and architectures, the training/testing accuracy of the model was still improving and showing better performance at higher global epochs than 200. For example, going from 200 epochs to 400 epochs, we noticed training and testing accuracy increment from around 83\% to around 86\% for FL with LeNet on FMNIST with 100 users. However, we limited our observation window to 100 or 200 global epochs as some network architecture such as AlexNet on HAM10000 in FL was taking an extensive amount of training time on the HPC (a shared resource).

    %As HPC is a shared resource and usually works on the batch jobs, it was hard to get sufficient resources ({\em e.g.,} computing nodes) for training on time if resources were requested for a more extended period.  
    
    %\subsection{Does the position of the cut layer affects the final result?}
    %Theoretically, the position of the cut layer in SL does not change the model performance as the 
    
    %Theoretically it is not in SL. But there can be effect due to the averaging of the model at the client-side as the size of the client-side network changes.
    
    	%===============================================================================
	\begin{figure*}[t]
	\centering
% 		\subfigure[]{
% 			\includegraphics[width=0.48\columnwidth]{fig/train-resnet-ham-fed-allusers.png}
% 		}
% 		\hskip-4pt
		\subfigure[FL]{
			\includegraphics[width=0.23\linewidth]{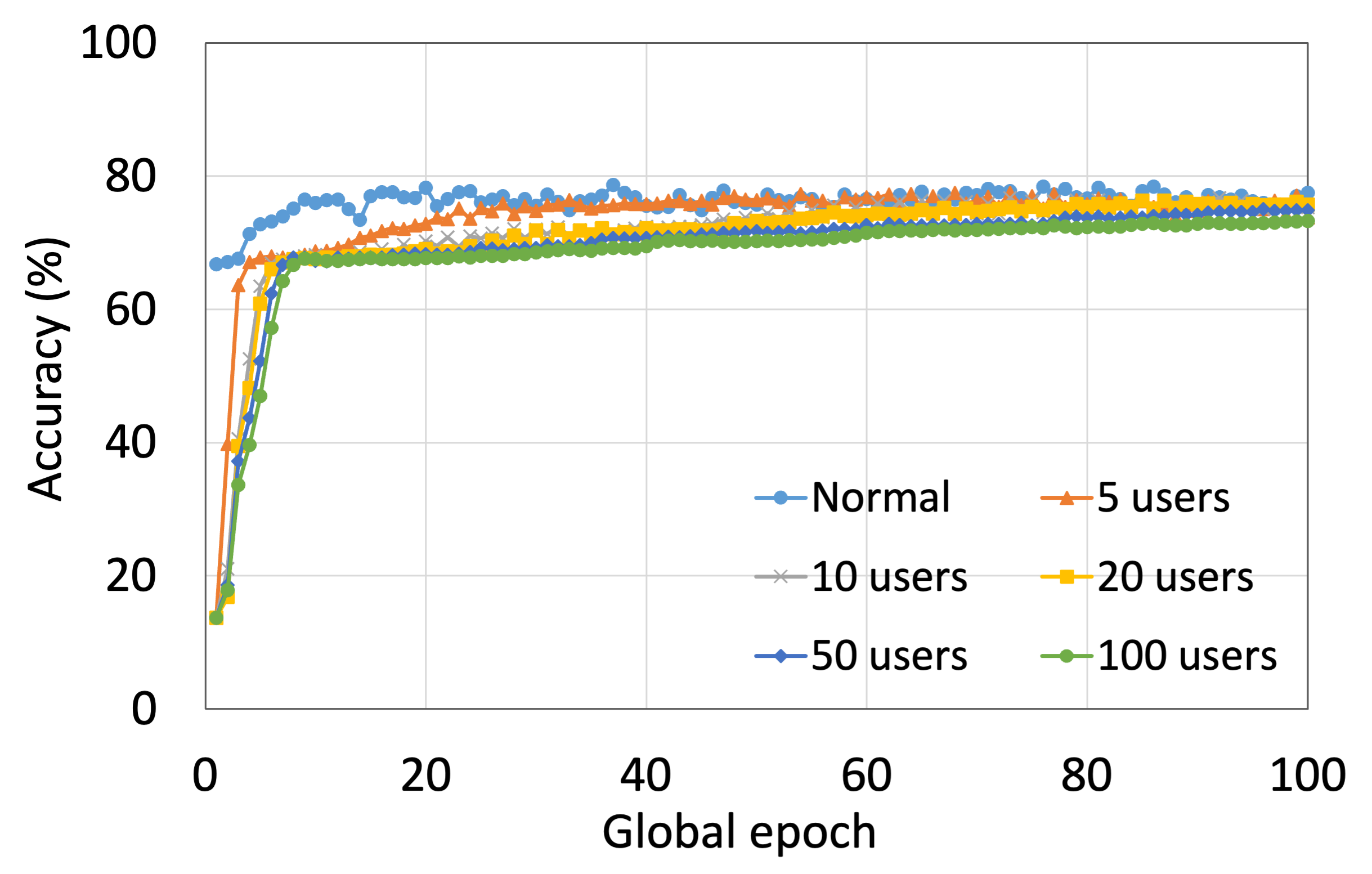}
		}
		\hskip-8pt
		%----------------------------
% 		\subfigure[]{
% 			\includegraphics[width=0.48\columnwidth]{fig/train-resnet-ham-split-allusers.png}
% 		}
		\subfigure[SL]{
			\includegraphics[width=0.21\linewidth]{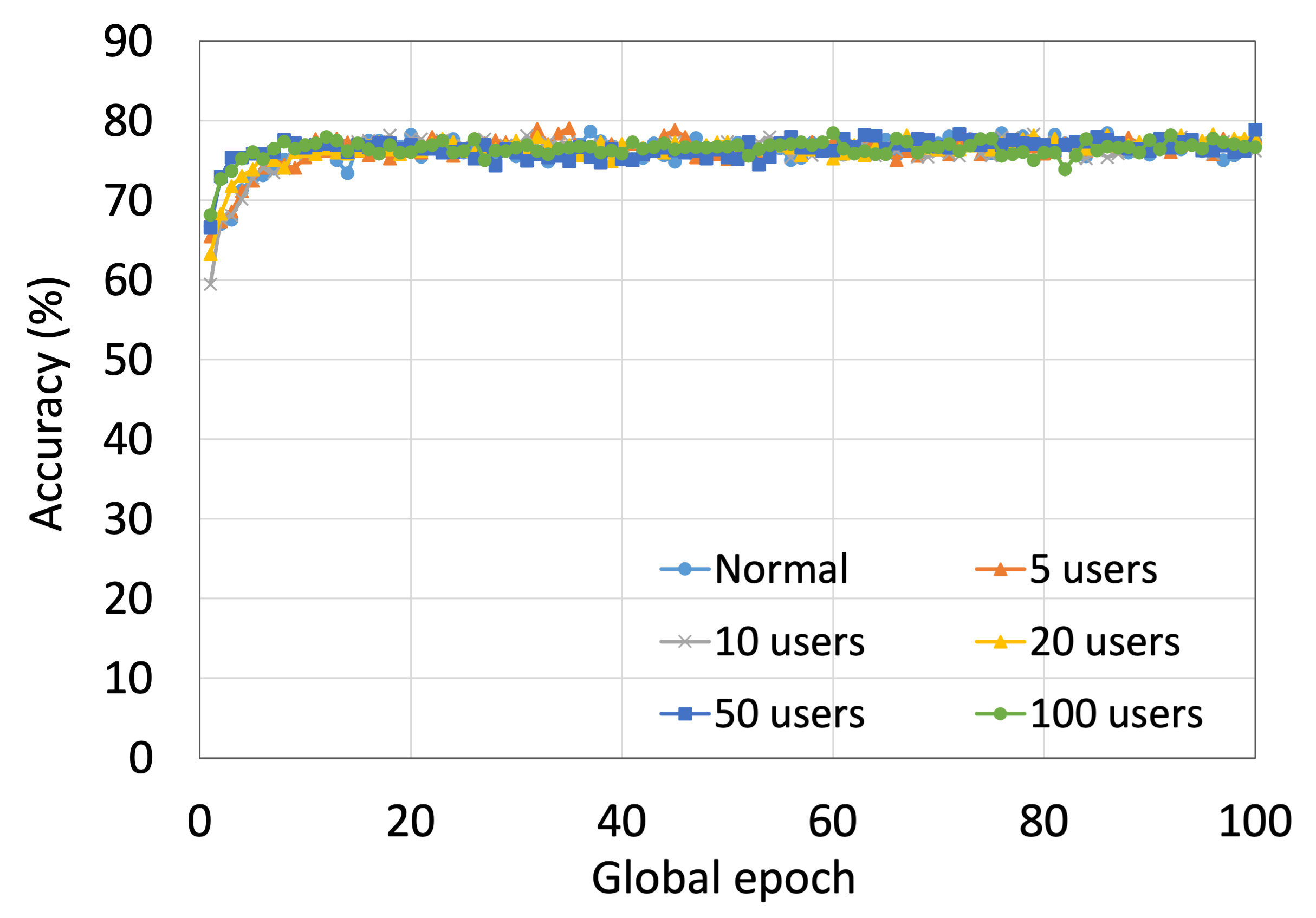}
		}
			\hskip-8pt
		%----------------------------
% 		\subfigure[]{
% 			\includegraphics[width=0.47\columnwidth]{fig/train-resnet-ham-splitfedv1-allusers.png}
% 		}
		\subfigure[SFLV1]{
			\includegraphics[width=0.24\linewidth]{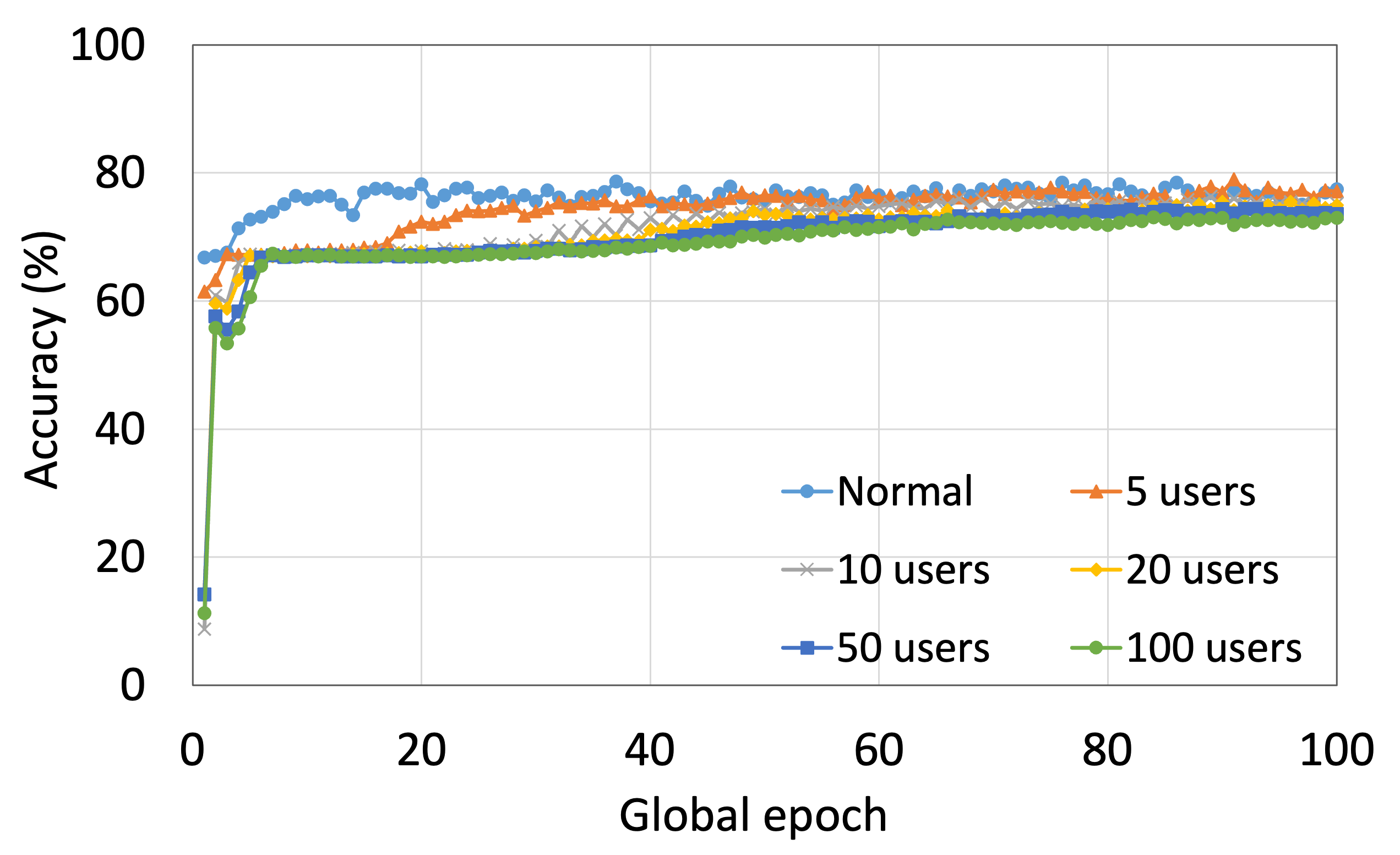}
		}
	    %----------------------------
% 	    \subfigure[]{
% 			\includegraphics[width=0.5\columnwidth]{fig/train-resnet-ham-splitfedv2-allusers.png}
% 		}
		\hskip-8pt
		\subfigure[SFLV2]{
			\includegraphics[width=0.25\linewidth]{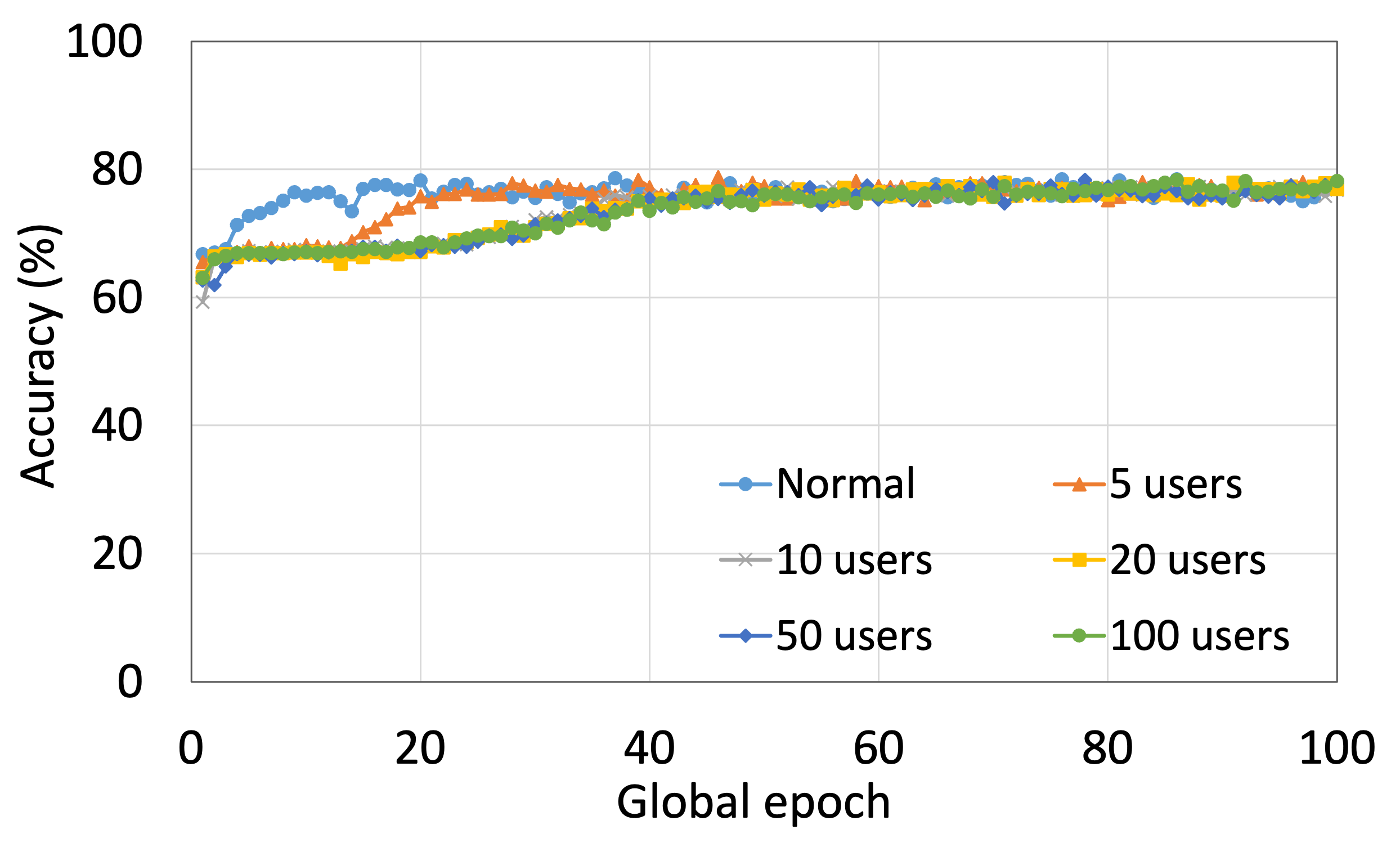}
		}	
		\caption{Effect of the number of client/users on testing accuracy for ResNet18 on HAM10000.}
		\label{fig:testingconvergence_resnet}
	
		%----------------------------
	 \end{figure*}
	%=====================================================================
	
	%-------------------------------------------------
    
	%============================subsection section=============================
	\subsection{Effect of Number of Users on the Performance}
		%--------------------------Resnet-------------------------------------------

	%	\caption{(a) Learning curve and (b) performance of split learning with ResNet18 on HAM10000 with various number of clients.}
	%	\label{fig:splitfmnistresnet}
	%	\caption{(a) Learning curve and (b) performance of splitfedv1 learning with ResNet18 on HAM10000 with various number of clients.}
	%	\label{fig:splitfedv1fmnistresnet}
%		\caption{(a) Learning curve and (b) performance of splitfedv2 learning with ResNet18 on HAM10000 with various number of clients.}
%		\label{fig:splitfedv2fmnistresnet}
	%
	This section presents the analysis of the effect of the number of users for ResNet18 on HAM10000. 
	%Refer to~\cite{samepaper} for the results of LeNet on FMNIST and AlexNet on HAM10000. 
	We observed that up to 100 clients (clients ranging from 5 to 100), the training and testing curves for all numbers of clients followed a similar pattern in each plot. Moreover, they achieved a similar level of accuracy within each of our DCMLs. We got comparative average test accuracies of around 74\% (FL), 77\% (SL), 75\% (SFLV1), and 77\% (SFLV2) at 100 global epochs. While training, only SL and SFLV2 achieved the centralized training (normal learning) accuracy at around 100 global epochs. In contrast, FL and SFLV1 could not achieve this result even at 200 global epochs.
	%(figures are presented only up to 100 global epochs to avoid any cluttering effects in the plots). 
	The experimental results for clients ranging from five to hundred showed a negligible effect on the performance due to the increase in the number of clients in FL, SL, SFLV1, and SFLV2 (for example, refer to Fig.~\ref{fig:testingconvergence_resnet}).  
	%For an individual DCML approach in our studies, results suggested a nominal effect on the learning and performance due to the increase in the number of clients from 5 up to 100. 
	However, this observation was not the case in general. For LeNet on FMNIST with fewer clients, the testing performances of FL and SL were close to the normal learning. Moreover, for SL with AlexNet on HAM10000, the performance degraded and even failed to converge with the increase in the number of clients, and we saw a similar effect on the SFLV2. 
	%(refer to~\cite{samepaper}). 
	Overall, the convergence of the learning and performance slowed down (sometimes failed to progress) with the increase in the number of clients (in our DCML techniques) due to the resource limitations and other constraints, such as the change in data distribution among the clients with the increase in its number, and a regular global model aggregation to synchronize the model across the multiple clients. %However, the model convergence will not slow down in a non-resource-constrained environment.

    %=====================================================
    % \begin{figure*}[tb]
    % \centering
    % \subfigure[FL]{\label{fig:cifar10-tsne_avg_master}
    % \includegraphics[width=1.8in]{fig/test-resnet-ham-fed-allusers.png}}
    % \subfigure[SL]{\label{fig:cifar10-tsne_prox_master}
    % \includegraphics[width=1.8in]{fig/test-resnet-ham-split-allusers.png}}
    % \subfigure[SFLv1]{\label{fig:cifar10-tsne_gkd_master}
    % \includegraphics[width=1.8in]{fig/test-resnet-ham-splitfedv1-allusers.png}
    % \subfigure[SFLv2]{\label{fig:cifar10-tsne_gkd_master}
    % \includegraphics[width=1.8in]{fig/test-resnet-ham-splitfedv2-allusers.png}
    % \caption{Testing convergence of ResNet18 on HAM10000 with various number of clients in federated, split and splitfed learning.}\label{fig:fmnistresnet}
    % \end{figure*}
%======================================================

%============================= subsection Discussion ==========	    
    \subsection{SFL with Differential Privacy at the Client-side Model with a PixelDP Noise Layer}
    \label{sec:diffprivacy}
   \begin{figure}[tb!]
	    \centering
	    \includegraphics[width=0.5\linewidth]{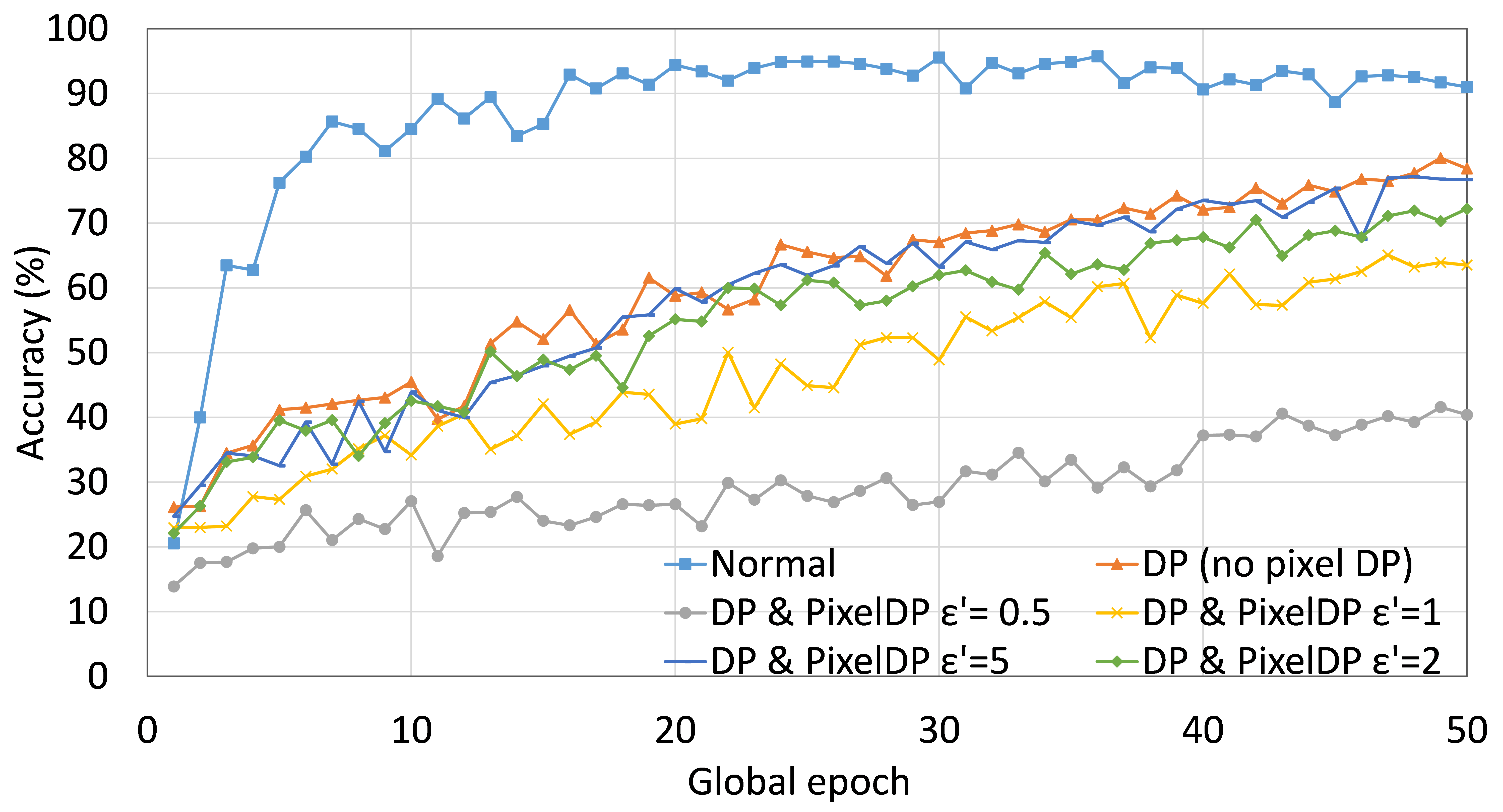}
    	\caption{SFLV1 testing convergence of AlexNet on MNIST with five clients, sensitivity $\delta = 1e^{-5}$, $\varepsilon = 0.5$, $\sigma = 1.3$ (DP), and under various choices of $\varepsilon'$ (PixelDP).} 
	    \label{fig:diffprivacy}
   \end{figure}

    %\paragraph{Privacy on Fed Server.}
    We implemented the differential privacy measures as described in Section ``Privacy Protection." For illustration, experiments were performed for SFLV1 with AlexNet on MNIST data distributed over five clients. 
    %With the application of differential privacy measures, the training time increases for our programs, which is evident due to the introduction of additional privacy-related computations.
    %We limit our observation window to 50 global epochs for this case. 
    For 50 global epochs with 5 local epochs at each client per global epoch, the testing accuracy curves converged as shown in Fig.~\ref{fig:diffprivacy}. Besides, as for illustration, we change the values of $\varepsilon'$, which is the privacy budget used by the PixelDP noise layer placed after the first convolution layer of AlexNet, to see the effect on the overall performance. Moreover, we maintain $\varepsilon$ at 0.5 (privacy budget of client-side model training) during all experiments to examine the behavior of SFLV1 under strict client-side model privacy. As expected, the convergence of accuracy curves with DP measures is gradual and slow compared to non-differentially private training. Besides, testing accuracy of around 40\%, 64\%, 73\%, 77\%, and 78\% are observed at global epoch 50 for $\varepsilon'$ equal to $0.5, 1, 2, 5$, and no PixelDP, respectively. Clearly, the accuracy increases with the increase in the privacy budget, i.e., $\varepsilon+\varepsilon'$. Overall, the utility is decreased with a decrease in the privacy budget. As the client-side architecture in SFLV2 is the same as in SFLV1, the application of differential privacy in SFLV2 can be done in the same way as in SFLV1.

	\section{Conclusion}
	\vskip2pt
	By bringing federated learning (FL) and split learning (SL) together, we proposed a novel distributed machine learning approach, named splitfed learning (SFL). SFL offered model privacy by network splitting and differential private client-side model updates. It is faster than SL by performing parallel processing across clients. Our results demonstrate that SFL provides similar performance in terms of model accuracy compared to SL. Thus, being a hybrid approach, it supports machine learning with resource-constrained devices (enabled by network splitting as in SL) and fast training (enabled by handling clients in parallel as in FL).
	%and privacy-preserving robust data analytics over sensitive data (enabled by differential privacy measures, including PixelDP). 
	%Our empirical results for ResNet18 on HAM10000 and AlexNet on MNIST showed that the two versions of splitfed learning (presented in this paper) were significantly faster than SL for multiple clients. Moreover, their speed was similar or even better than FL in some instances, and they had the same communication efficiency as SL and improved its efficiency than FL with the increase in the number of clients. 
	The performance of SFL with privacy and robustness measures based on differential privacy and PixelDP was further analyzed to investigate its feasibility towards data privacy and model robustness. 
	%
	%To further analyze and strengthen the privacy and robustness of splitfed learning, differential privacy measures were implemented on the gradients' update process, and noisy smashed data were transmitted from the clients to the main server. These steps provide enhanced flexibility of splitfed towards eliminating potential privacy leakage among the client-client, client-main server, and client-fed server during collaborative learning. 
	%
	Studies related to the detailed trade-off analysis of privacy and utility, and integration of homomorphic encryption~\cite{gentry} for guaranteed data privacy are left for future works.   

	%\section{Limitations  and future works}
	%In this study, we have analyzed the performance of FL, SL, SFLV1, and SFLV2 based on their privacy-preserving features, model accuracy, communication, and computation costs. In addition, we have implemented and analyzed the SFLV1 with differential privacy measures on the client side. 
	%The comparative performance analysis of these \sloppy privacy-preserving distributed ML approaches with the integration of HE~\cite{gentry} for guaranteed privacy has remained as future work. Also, in our experiments on SFLV1 with differential privacy at the client-side, a detailed trade-off analysis of privacy and utility is not done. Thus, the studies related to privacy-utility trade-off analysis are left as future work. 
	
%\afterpage{\FloatBarrier}

%==========================================================================
%                       END
%==========================================================================

\bibliographystyle{splncs04}
\bibliography{bibliography}

\appendix
\section*{Appendix}

\section{Threat Model}
	\label{tmsection}
	
	In this work, for federated learning (FL), split learning (SL), and splitfed learning (SFL), we consider the fed server and the main server of the system model are honest-but-curious adversaries. They perform the tasks as specified but can be curious about the local private data of the clients and the full model. Moreover, the attack scenario for the servers are considered here is passive, where they only observe the updates and possibly do some calculations to get the information, but they do not maliciously modify their own inputs or parameters for the attack purpose. Besides, the servers and clients are non-colluding with each other. Regarding clients, they are considered as curious adversaries; behave honestly while training, but can input adversarial inputs only during inference (testing).  
	
	We assume that our approaches splitfedv1 (SFLV1) and splitfedv2 (SFLV2) adhere to the standard client-server security model where the clients and servers establish a certain level of trust before starting the network model training. For example, in the health domain, the hospitals (clients) only allow platform (server) and researchers (model owners) that have a level of trust. Hospitals opt out if the corresponding platform has malicious clients or servers. Besides, we assume that all communications between the participating entities (e.g., exchange of smashed data and gradients between clients and the main server) are performed in an encrypted form. 
	%Privacy in SFL (both versions) is enabled by the MTD approach, and network split, as discussed in Section~\ref{sflprivacy}. 
	Compared with the threat model of FL, SFL has an extra honest-but-curious and non-colluding fed server at the client-side (similar to the case of SL in centralized mode). The fed server and clients in SFL have access only to the gradients/weights of the client-side model portions only. In FL, the main server has access to the entire model. In contrast, the main server has access only to the server-side portion of the model and the smashed data (i.e., activation vectors of the cut layer) from the client in SL and SFL.

\section{Total Cost Measurement}

In addition to our theoretical analyses of the communication cost and the model training time cost of FL, SL, SFLV1, and SFLV2 learning, we perform the empirical test on communication and training time. These results complement our analysis in Section ``Total Cost Analysis."

\subsection{Communication Measurement}
	
\begin{figure}[t]
	\centering
		\subfigure[]{
			\includegraphics[width=0.24\linewidth]{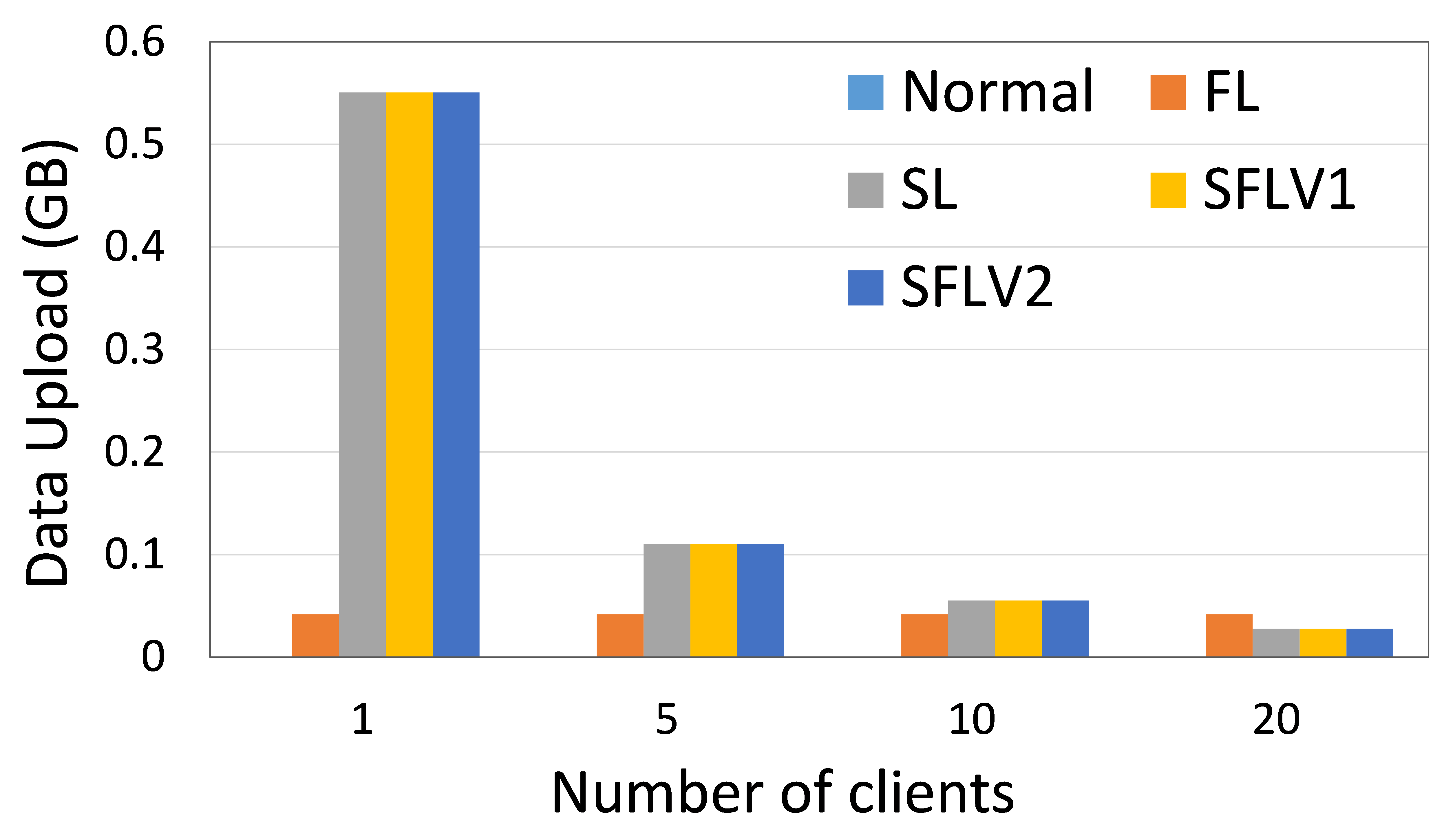}
		}
		\hskip-4pt
    	\subfigure[]{
			\includegraphics[width=0.24\linewidth]{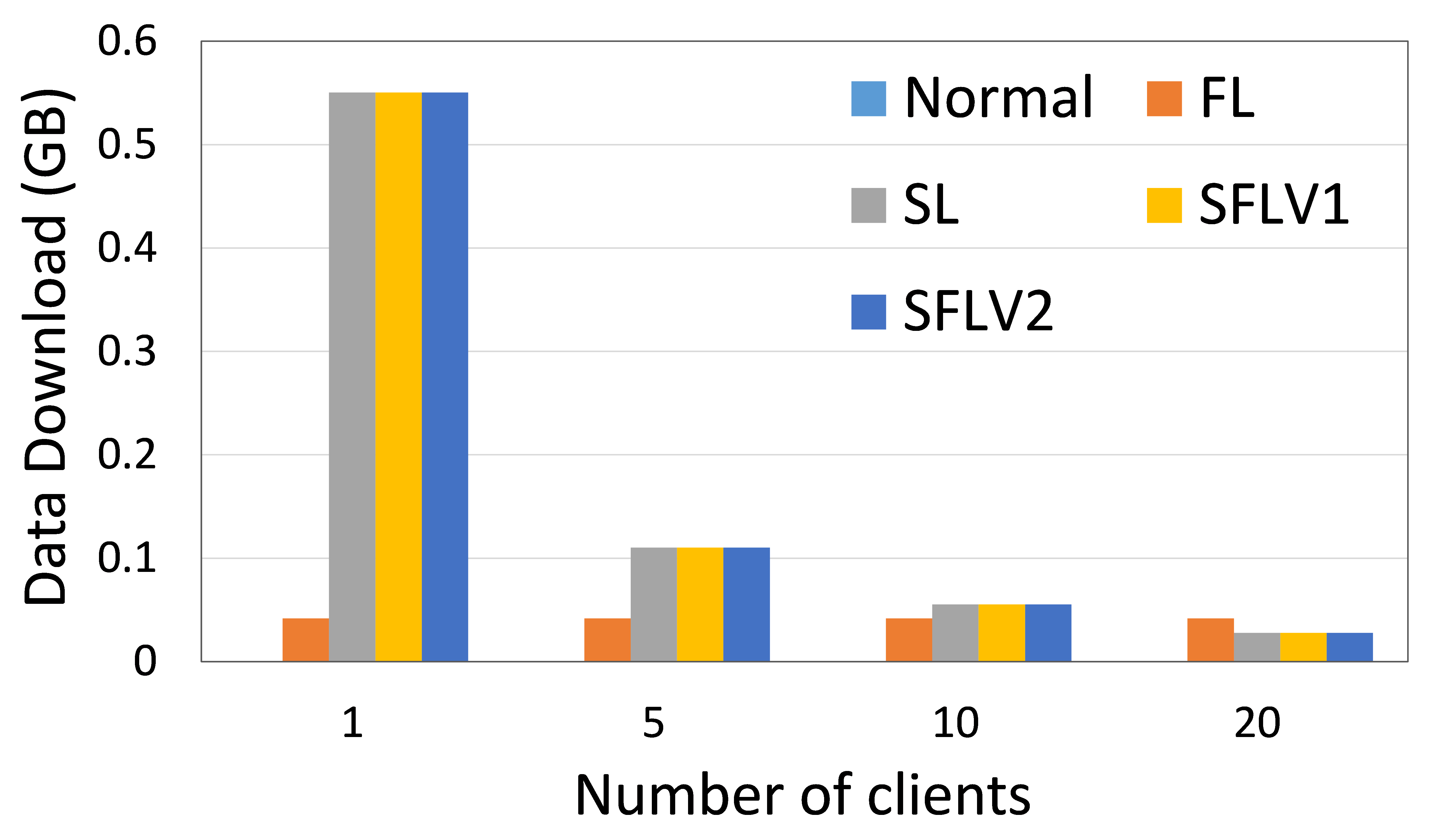}
		}
	%\caption{(a) Data upload, and (b) data download for ResNet18 on HAM10000 with a various number of clients.}
% 	\end{figure}
	\hskip-4pt
% 	\begin{figure}[htb]
		\subfigure[]{
			\includegraphics[width=0.23\linewidth]{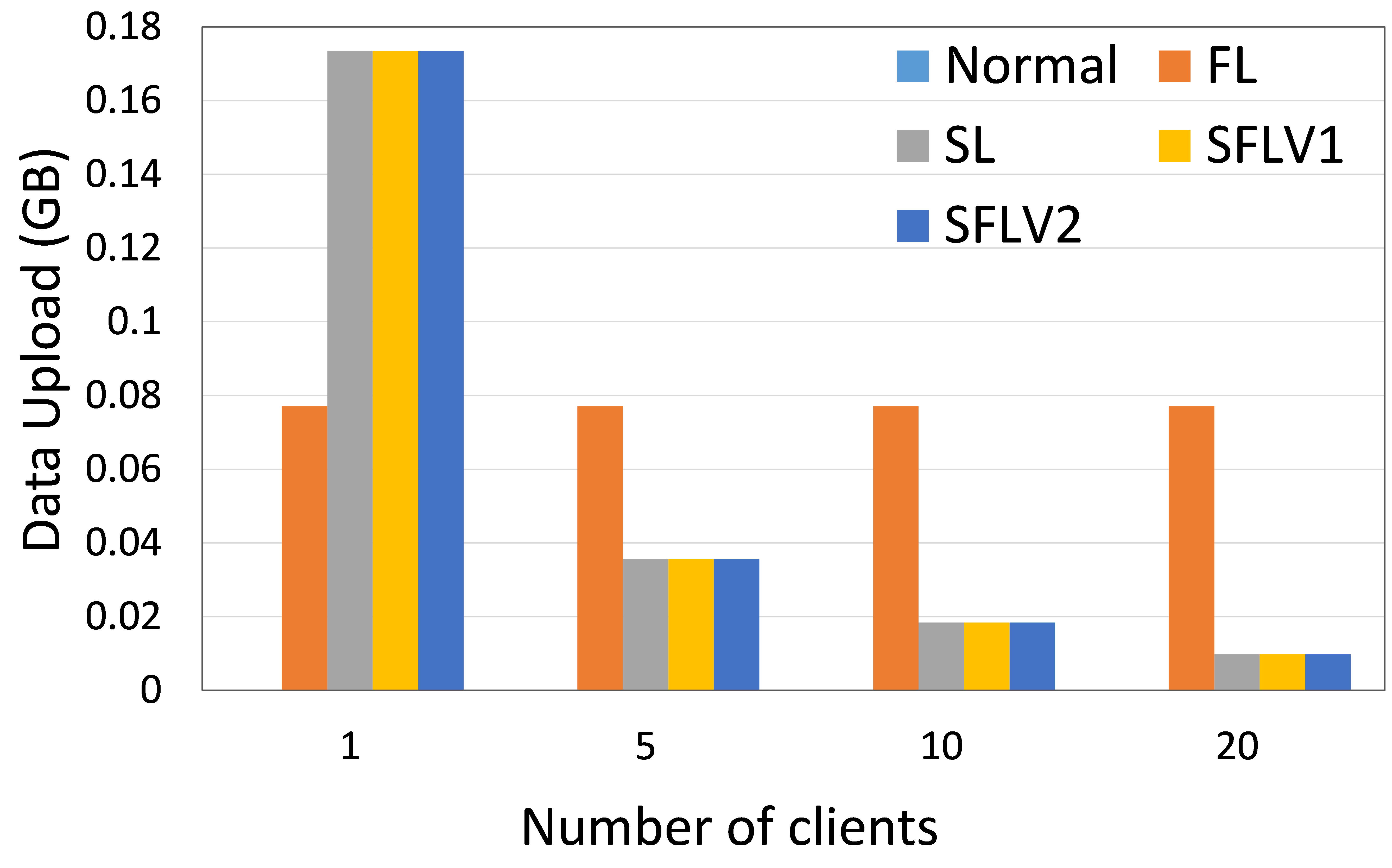}
		}
		\hskip-4pt
% 		\end{figure}
% 		\begin{figure}[H] 
% 		\ContinuedFloat
% 		\setcounter{subfigure}{3}
% 		\centering
		\subfigure[]{
			\includegraphics[width=0.23\linewidth]{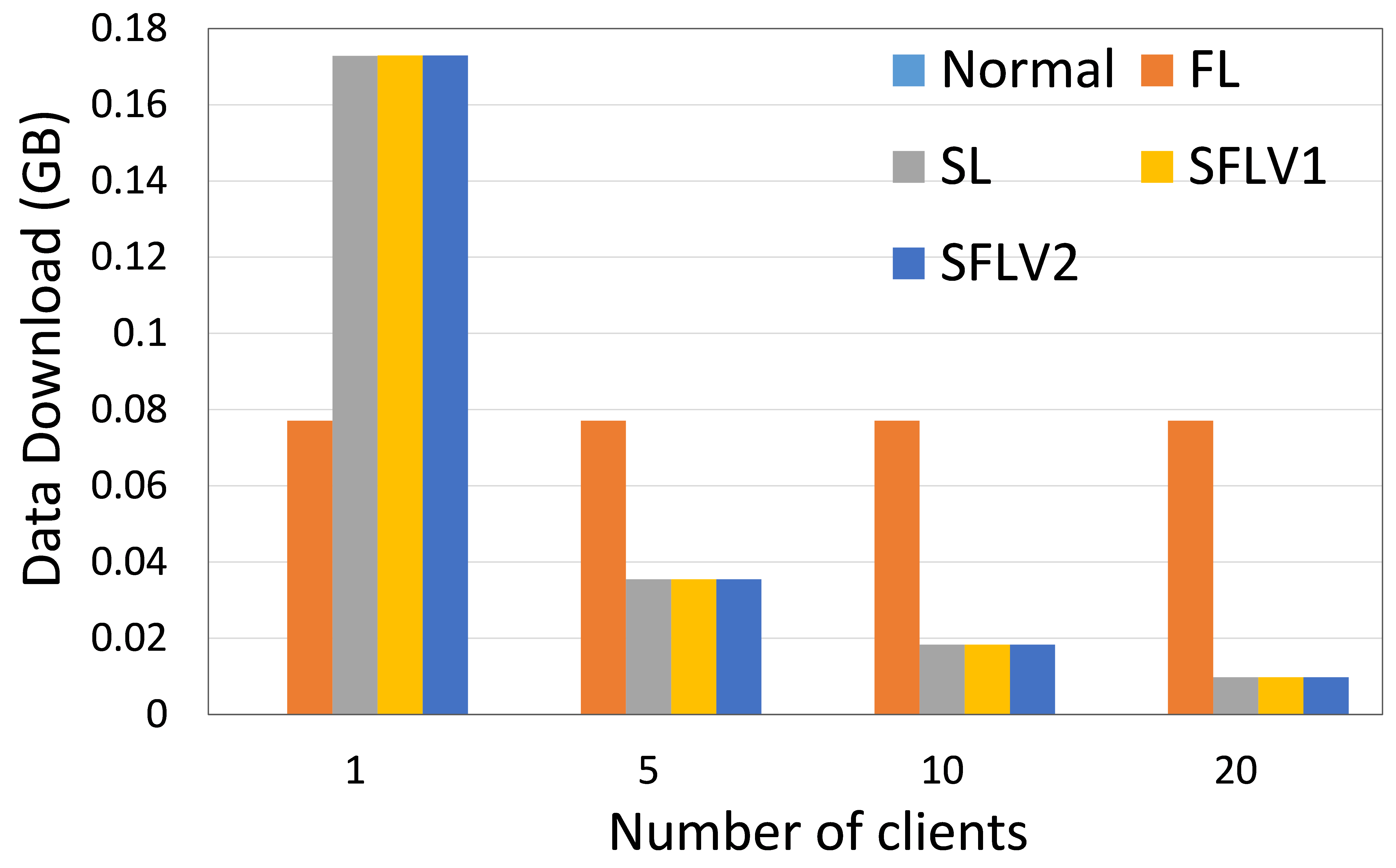}
		}
		\caption{Communication measurement: average data per client per global epoch (a) upload for ResNet18 on HAM10000, (b) download for ResNet18 on HAM10000, (c) upload for AlexNet on MNIST, and (d) download for AlexNet on MNIST.}
		\label{fig:datafmnistalexnet}
	\end{figure}

	The amount of data uploaded and downloaded by a client indicates the operability of a Distributed Collaborative Machine Learning (DCML) approach in a resource-constrained environment. High data communication slows down the machine learning (ML) training and testing process, and the clients need to have sufficient resources to handle the high communication cost. In this regard, we measure the amount of data communication in our experiments and present the relative performance of the four DCML techniques. To make the experimental setup normalized for different numbers of clients, we run our program under the following configurations: The main server's and fed server's programs run in two different HPC nodes, and clients' programs run in five separate nodes (except for experiment with one client). Each client HPC node runs one, two, and four clients (client-side programs) for five, ten, and twenty client cases. We record the total communication for the observation window of 11 global epochs with one local epoch and a batch size of 256. The results were then averaged over all global epochs and clients to obtain the communication cost per client per global epoch. Refer to Fig.~\ref{fig:datafmnistalexnet} for the communication load of ResNet18 on HAM10000 and AlexNet on MNIST. 
    
    %During the experiments with various models and dataset, we observed the same amount of data communication for SL, SFLV1, and SFLV2. For example, around 0.27 GB of average data upload and download per global epoch per client for 20 clients scenario. This complements our analysis in section~\ref{sec:costanalysis}, and shows that SFLV1 and SFLV2 provided the same communication efficiency as in SL. The communication load in SL, SFLV1 and SFLV2 depend on the size of activations, which depends on the dataset size and size of the cut layer, whereas the communication load in FL depends only on the size of model. 
    %As the number of clients increased, the amount of data communication (which was averaged over clients) per global epoch decreased for SL, SFLV1, and SFLV2. This was because of the following two reasons: (1) SL and both versions of SFL were communicating smashed data rather than (model) weights. Thus, the amount of communication was the function of the data samples. (2) The overall dataset was uniformly distributed among clients, so the dataset size per client decreased with the increase in the number of clients. On the other hand, this effect was not applicable to FL, which had the same level of data communication in all cases. This was because, in FL, for given network architecture, each client sends the locally trained network to the server and receives the global network independent of the sample size. 

    Recent works  have shown that SL is more communication efficient than FL while increasing the number of clients or model size~\cite{splitcommefficiency}. In contrast, FL is preferred if the amount of data samples is increased by keeping the number of clients or model size low~\cite{splitcommefficiency,ourpaper1}.

    %---------------------------------------------------

	%============================= subsection time measurement==========

	\subsection{Time Measurement}
	\label{sec:timemeasurement}
	
	\begin{figure}[t]
	\centering
	\hskip-2pt
		\subfigure[ResNet18 on HAM10000]{
			\includegraphics[width=0.25\linewidth]{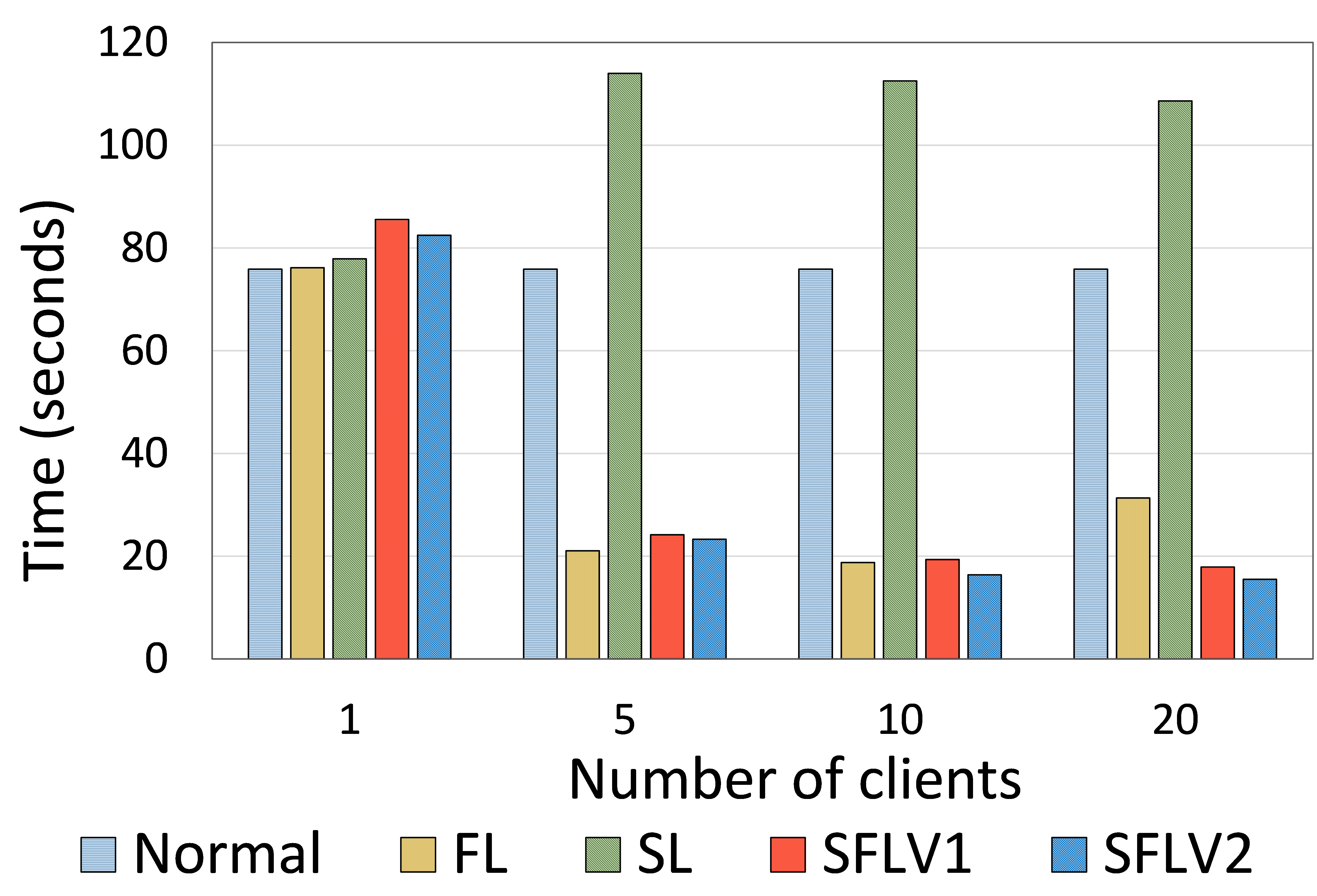}
		}
		\subfigure[AlexNet on MNIST]{
			\includegraphics[width=0.25\linewidth]{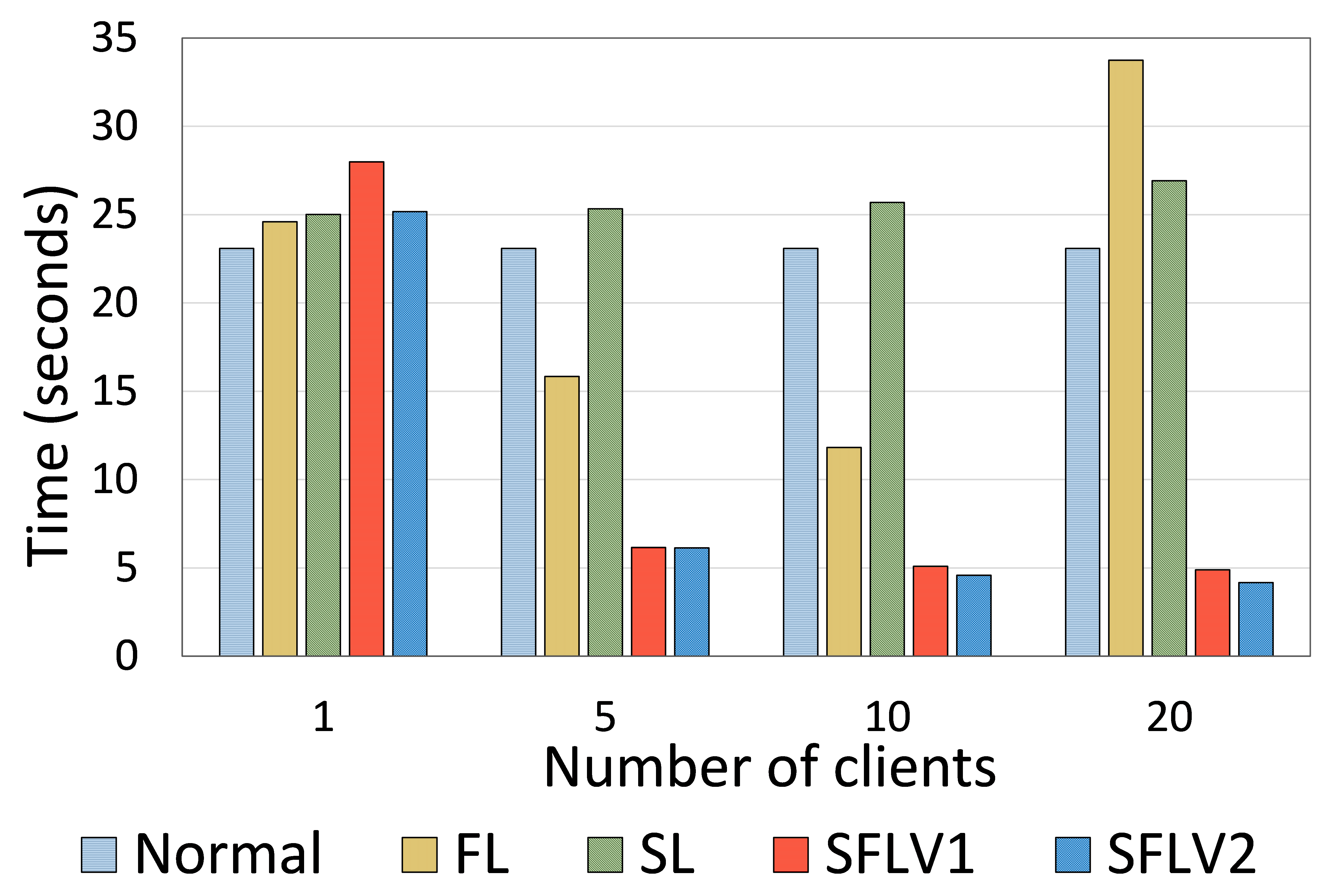}
		}
		\caption{Time measurement: training time per global epoch.}
		\label{fig:timemeasurement}
	\end{figure}

% 	\begin{figure}[!t]
% 	    \centering
% 	    \includegraphics[width=0.7\linewidth]{fig/time_measurement_resnet}
%     	\caption{Time measurement for ResNet18 on HAM10000 under various number of clients.}
% 	    \label{fig:timemeasurementresnet}
%     \end{figure}

	To show the time efficiency of SFLV1 and SFLV2 compared to SL, we analyze the training time taken for one global epoch.
	Considering Section ``Total Cost Analysis," Algorithm 1 and 2, it is not difficult to see the following: For SL, the main overhead is due to the client-side model upload and download by each client, and for a global epoch, it is a multiple of the total number of clients ($K$). In contrast, there are no such product terms while calculating the time in SFLV1 and SFLV2 because the server can be a supercomputing resource and processes all clients in parallel. Consequently, SFL (both versions) is faster than SL for multiple clients. Besides, this also shows that the training time measurement depends not only on the implementation but also on the algorithm. For our experiments in SFL, we implemented a \emph{multithreaded python program} for the main server. Using the same experimental setup (which was used to measure the communication cost), we ran each experiment for eleven global epochs and recorded the time for each global epoch. Unlike in the communication measurement setup, the training time was averaged by considering the time from the second global epoch onward for all clients after running each experiment ten times. The time for the first global epoch was excluded because it included the time taken by clients to connect to the server, i.e., the initial connection overhead (in our setup, all clients got connected to the server at first and kept the connection during the experiment). We ran each experiment ten times - different HPC slurm jobs in each instance - to exclude the effects of the change in the computing environment in each run. 
    
    Based on our observations on ResNet18 on HAM10000 and AlexNet on MNIST, the time statistics for the cases with multiple clients indicated that SFLV1 and SFLV2 were significantly faster - \emph{by four to six times} - than SL. It had a similar or even better speed than FL (refer to Fig.~\ref{fig:timemeasurement}). For the single client case, SL and FL approaches spent similar time; SFLV1 and SFLV2 spent slightly more time than the other.

%========================================================    
 
 \section{Additional Empirical Results}
	\label{learningnperformance}

	\subsection{Performance of FL, SL, SFLV1, and SFLV2}
	
	This section presents the training and testing convergence plots of normal (centralized) learning, FL, SL, SFLV1, and SFLV2. The experiments are performed considering five clients and using AlexNet on HAM10000, ResNet18 on MNIST, AlexNet on MNIST, LeNet5 on FMNIST, AlexNet on FMNIST, LeNet5 on CIFAR10, and VGG16 on CIFAR10. We present the training and testing results to demonstrate both the pictures of training and testing instances.

\begin{figure*}[t]
	\vspace{0.5cm}
	\centering
	% AlexNet on Ham10000 ----------------------------------------->
		\subfigure[{\scriptsize Train: AlexNet on HAM10000}]{
			\includegraphics[width=0.21\linewidth]{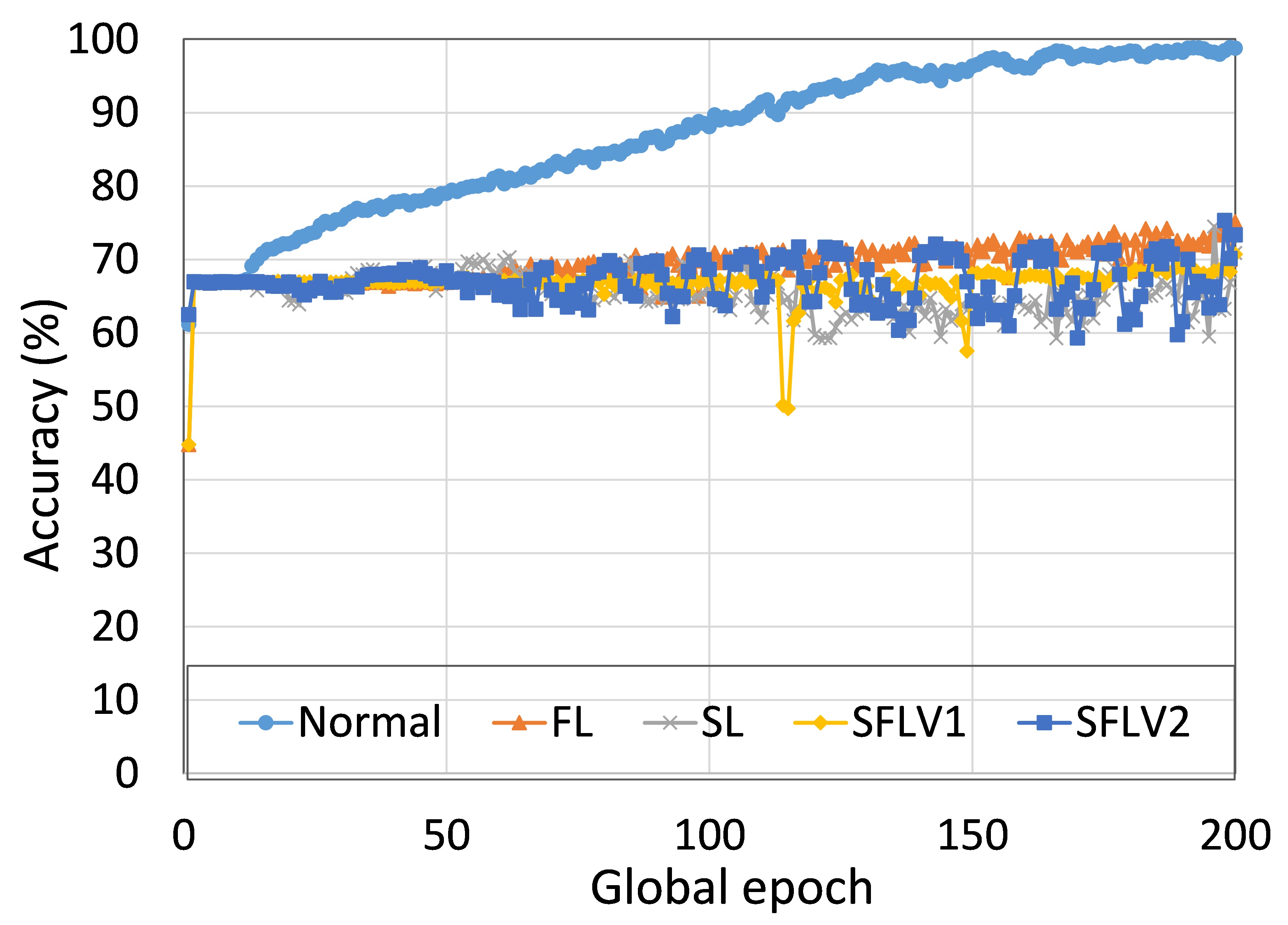}
		}
		\hskip-2pt
		\subfigure[{\scriptsize Test: AlexNet on HAM10000}]{
			\includegraphics[width=0.21\linewidth]{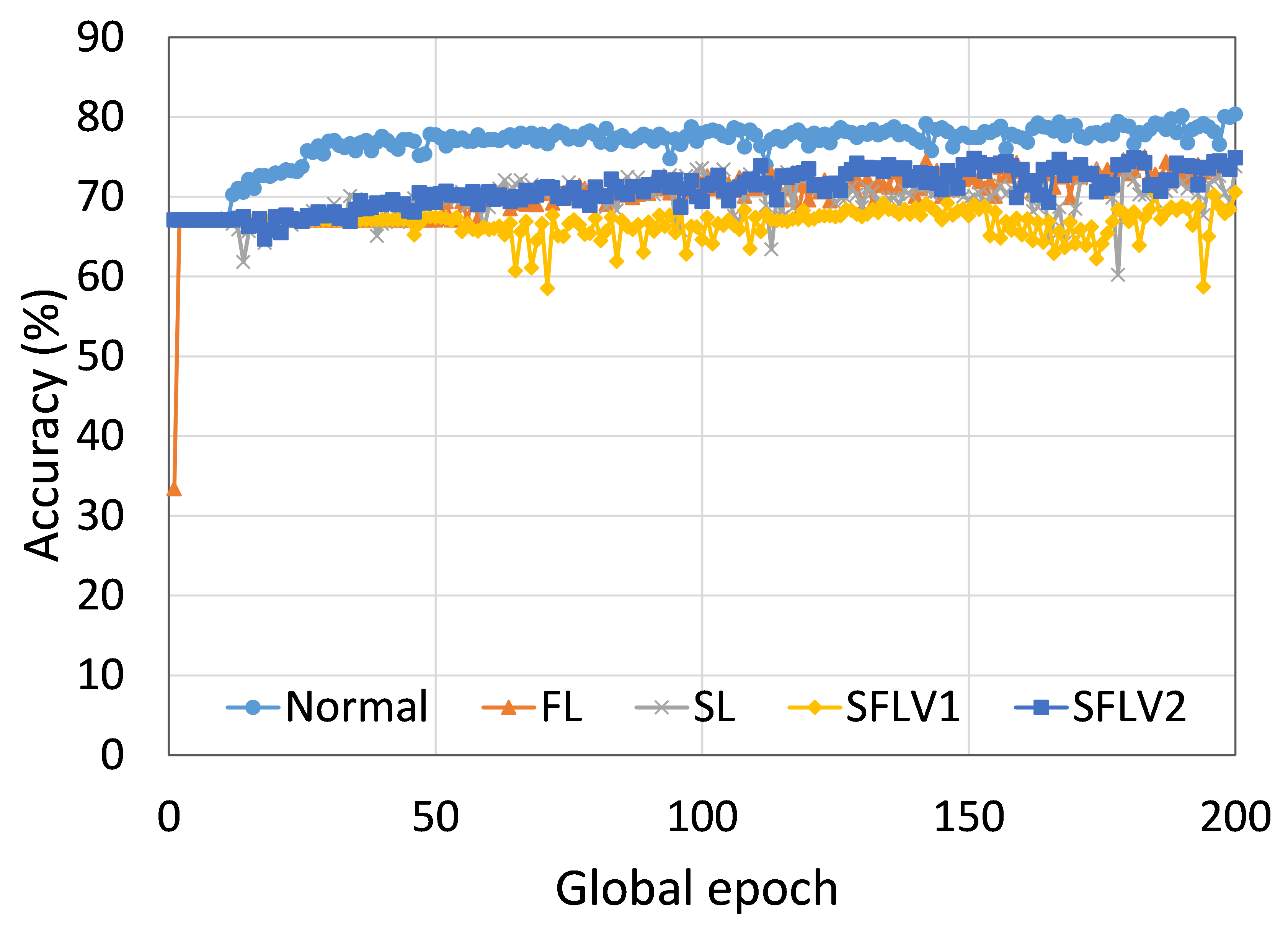}
			
		}
			\hskip-2pt
	%-------------ResNet18 on mnist-------------------------------->
	    \subfigure[{\scriptsize Train: ResNet on MNIST}]{
			\includegraphics[width=0.25\linewidth]{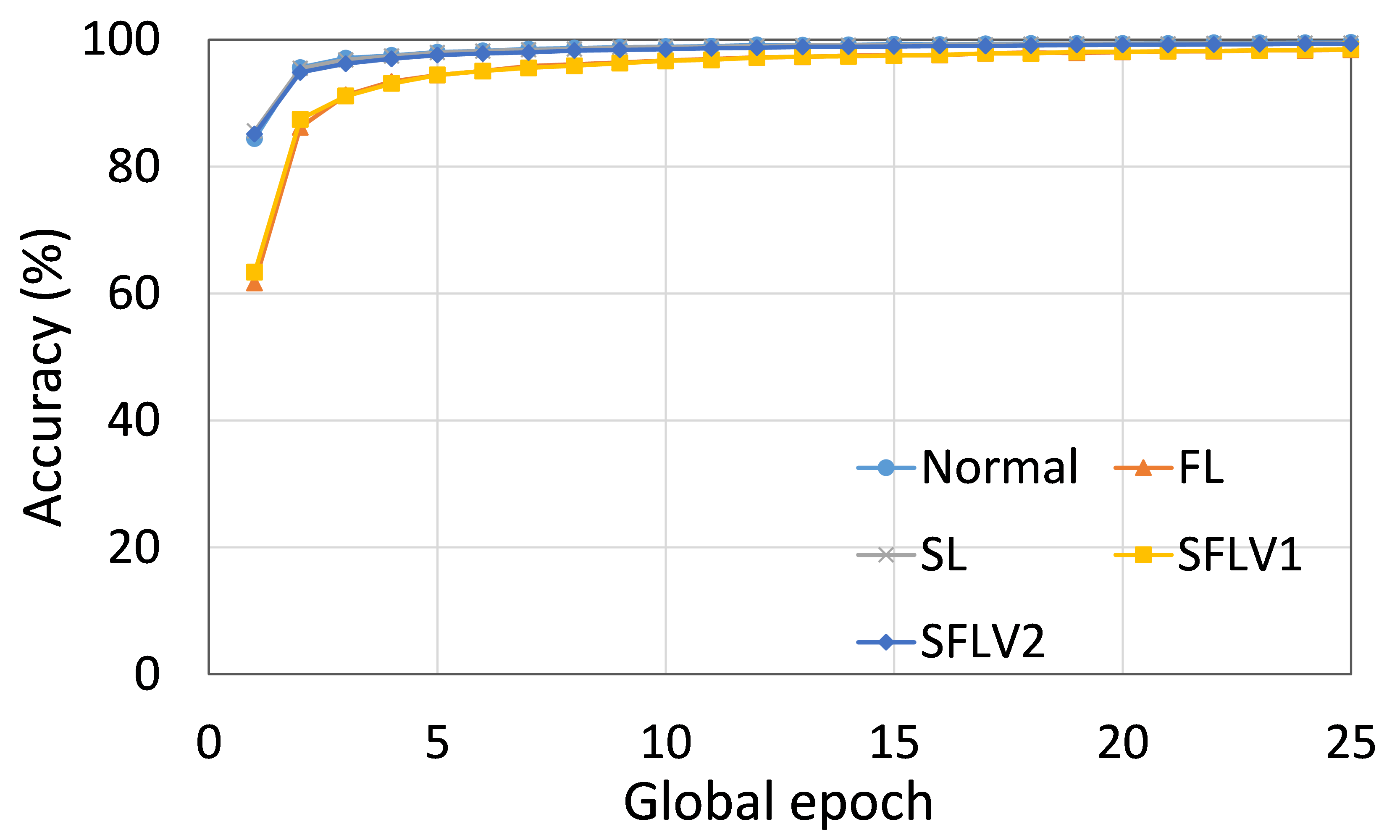}
		}
			\hskip-2pt
		\subfigure[{\scriptsize Test: ResNet on MNIST}]{
			\includegraphics[width=0.25\linewidth]{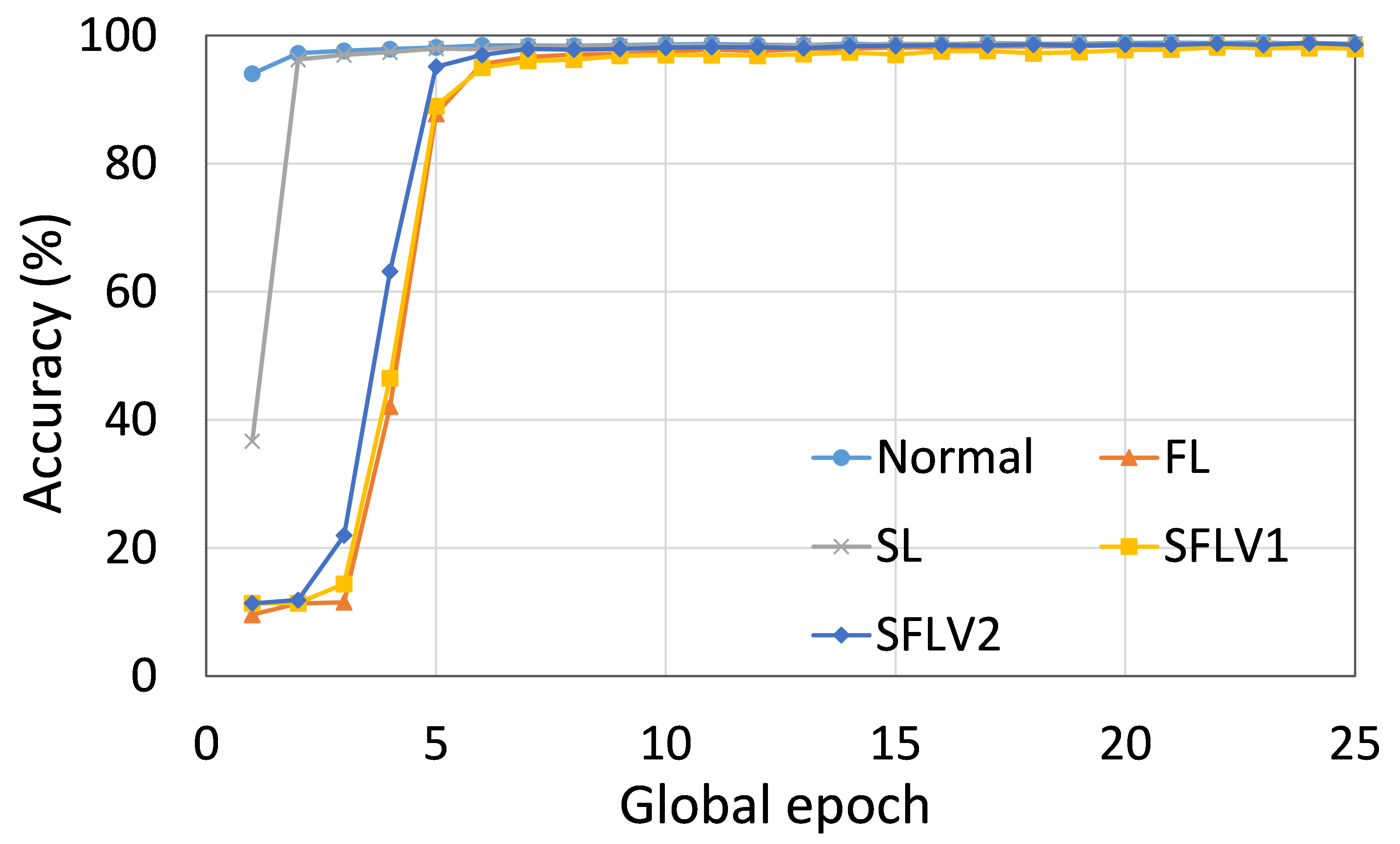}
		}	
			\hskip-2pt
%================================================================
	%-------------AlexNet on mnist-------------------------------->
	    \subfigure[\scriptsize Train: AlexNet on MNIST]{
			\includegraphics[width=0.23\linewidth]{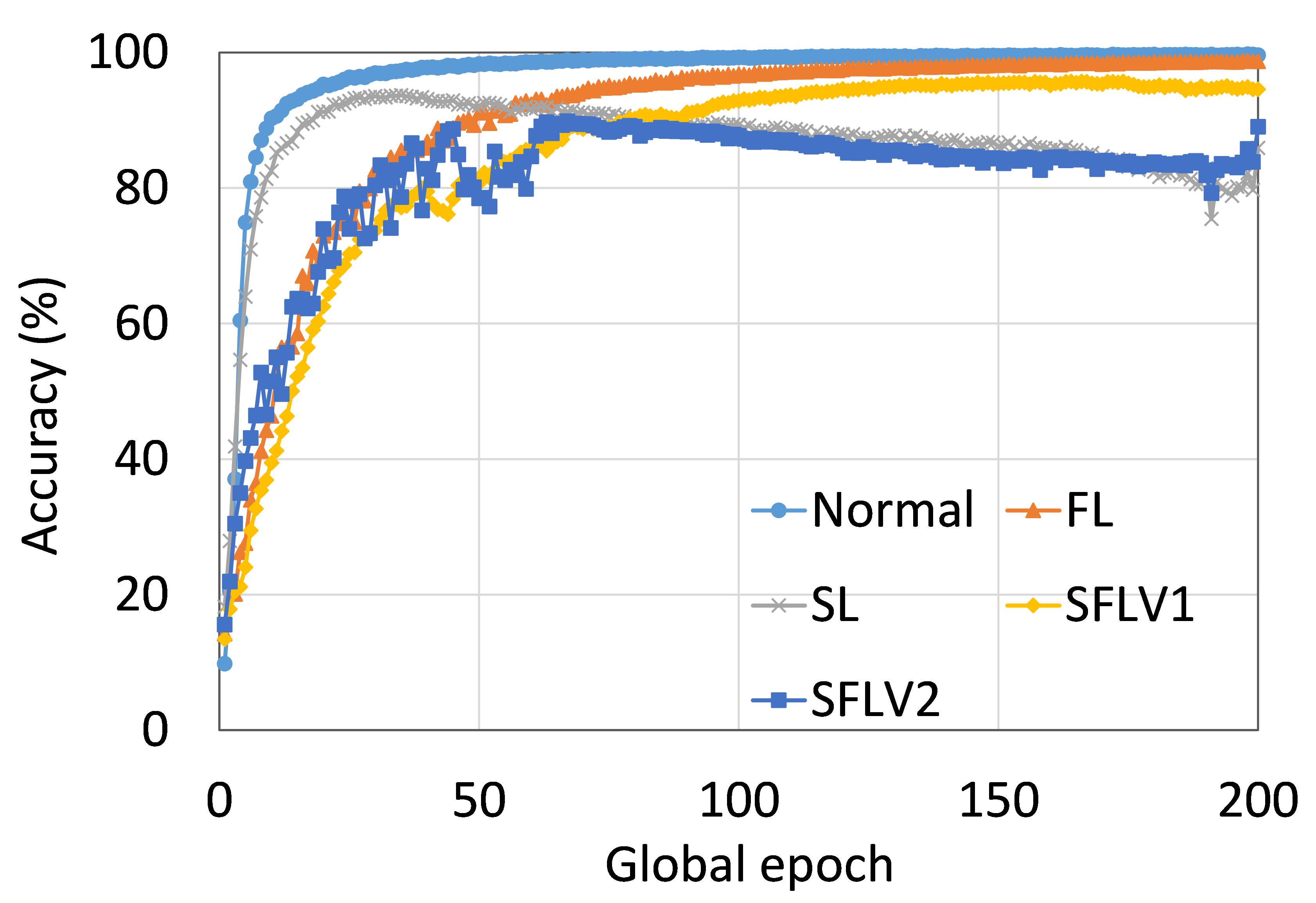}
		}
			\hskip-2pt
		\subfigure[\scriptsize Test: AlexNet on MNIST]{
			\includegraphics[width=0.23\linewidth]{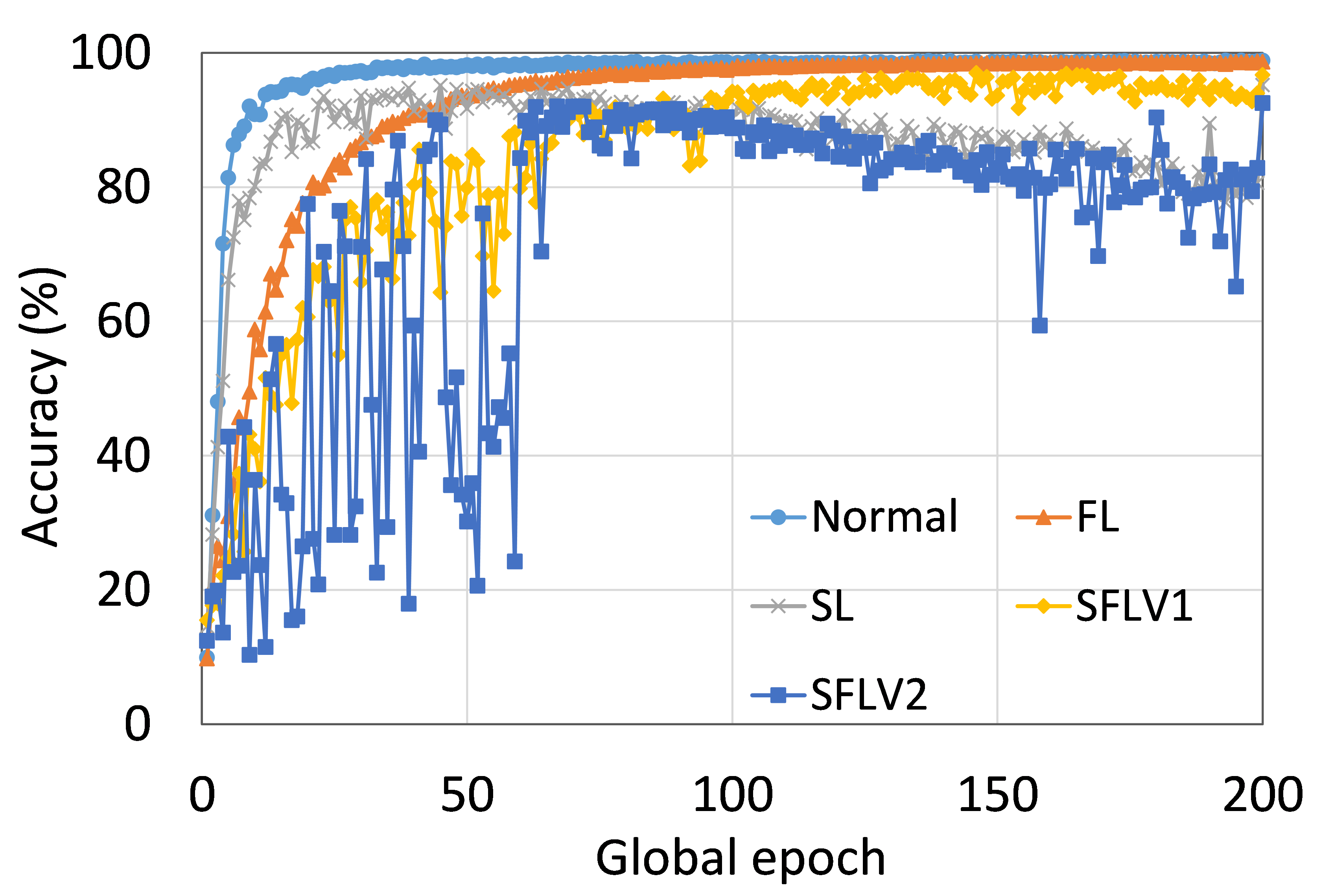}
		}	
			\hskip-2pt
	%----------LeNet on fmnist ----------------------------------->
	    \subfigure[\scriptsize Train: LeNet5 on FMNIST]{
			\includegraphics[width=0.23\linewidth]{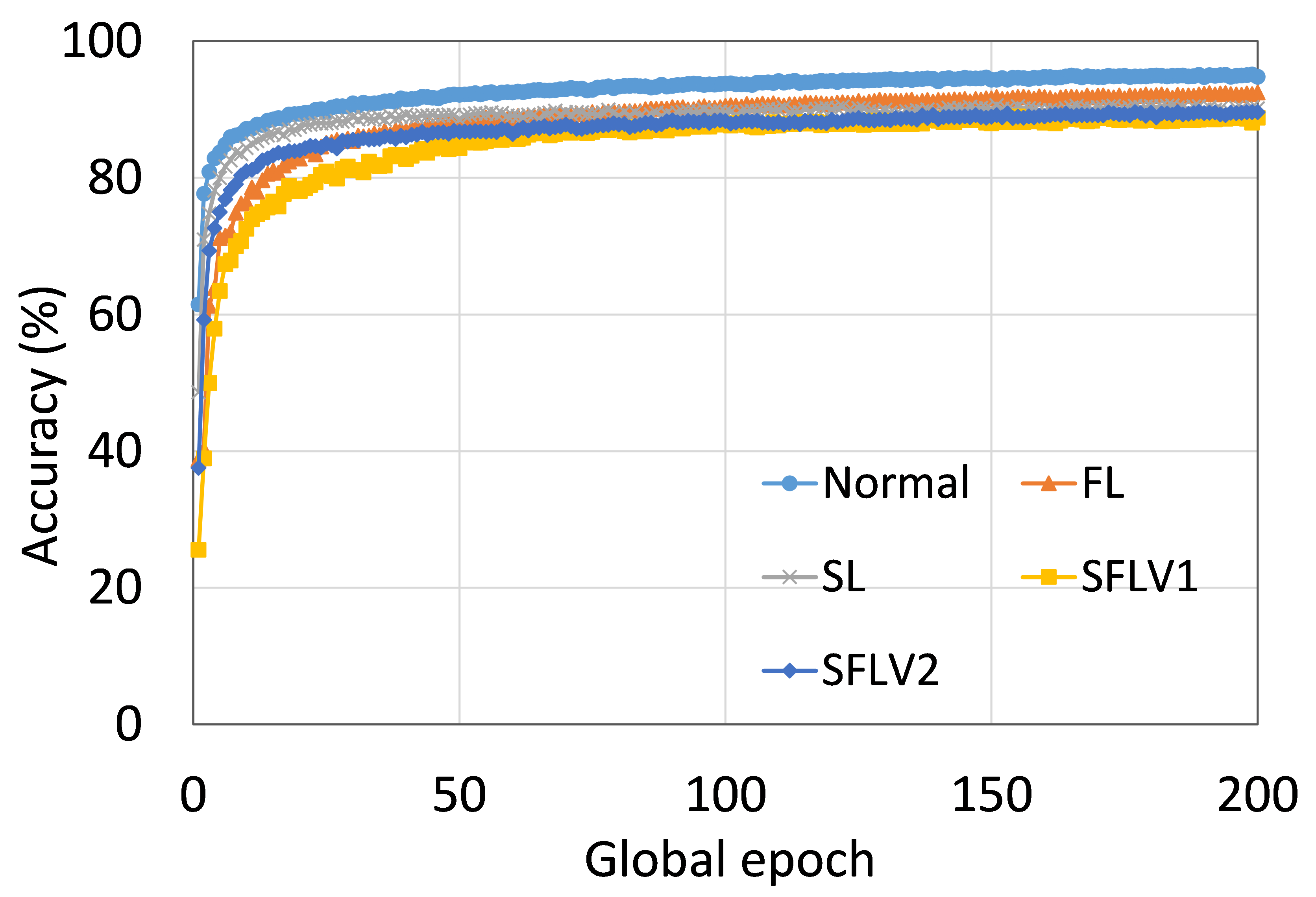}
		}
		\hskip-2pt
		\subfigure[\scriptsize Test: LeNet5 on FMNIST]{
			\includegraphics[width=0.23\linewidth]{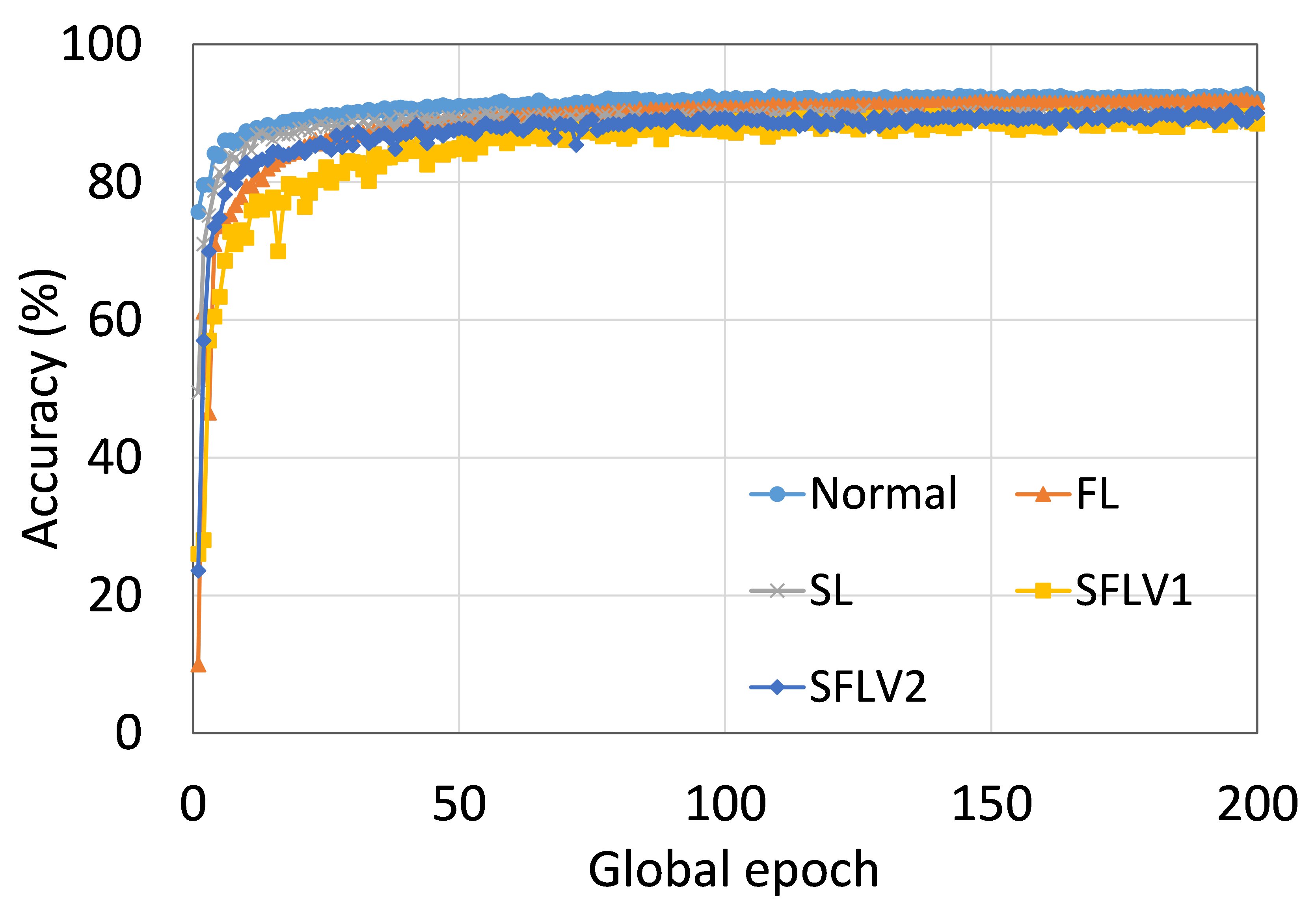}
		}	
			\hskip-2pt
%================================================================
	%----------AlexNet on fmnist ----------------------------------->
		\subfigure[\scriptsize Train: AlexNet on FMNIST]{
			\includegraphics[width=0.23\linewidth]{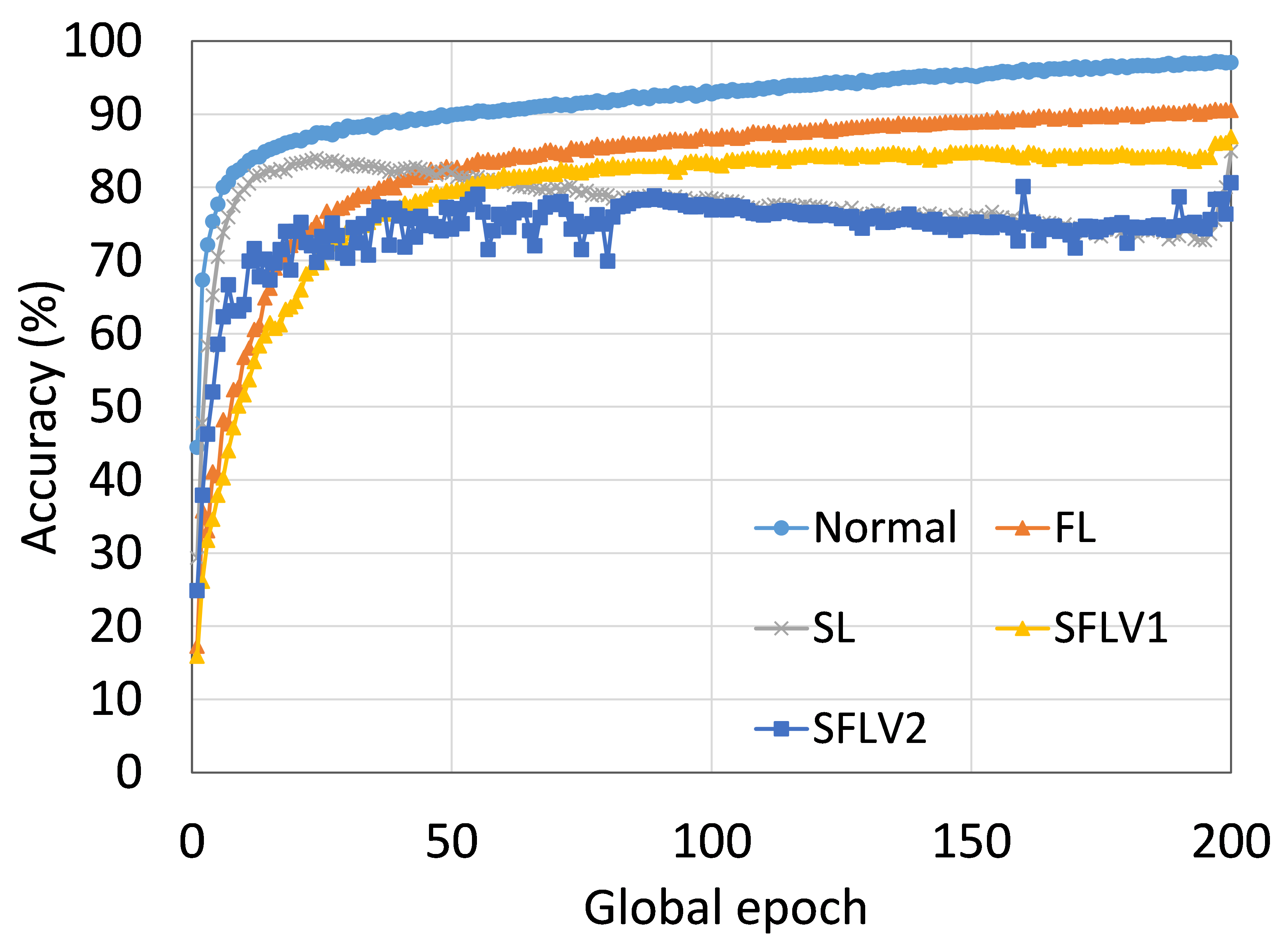}
		}
	\subfigure[\scriptsize Test: AlexNet on FMNIST]{
			\includegraphics[width=0.23\linewidth]{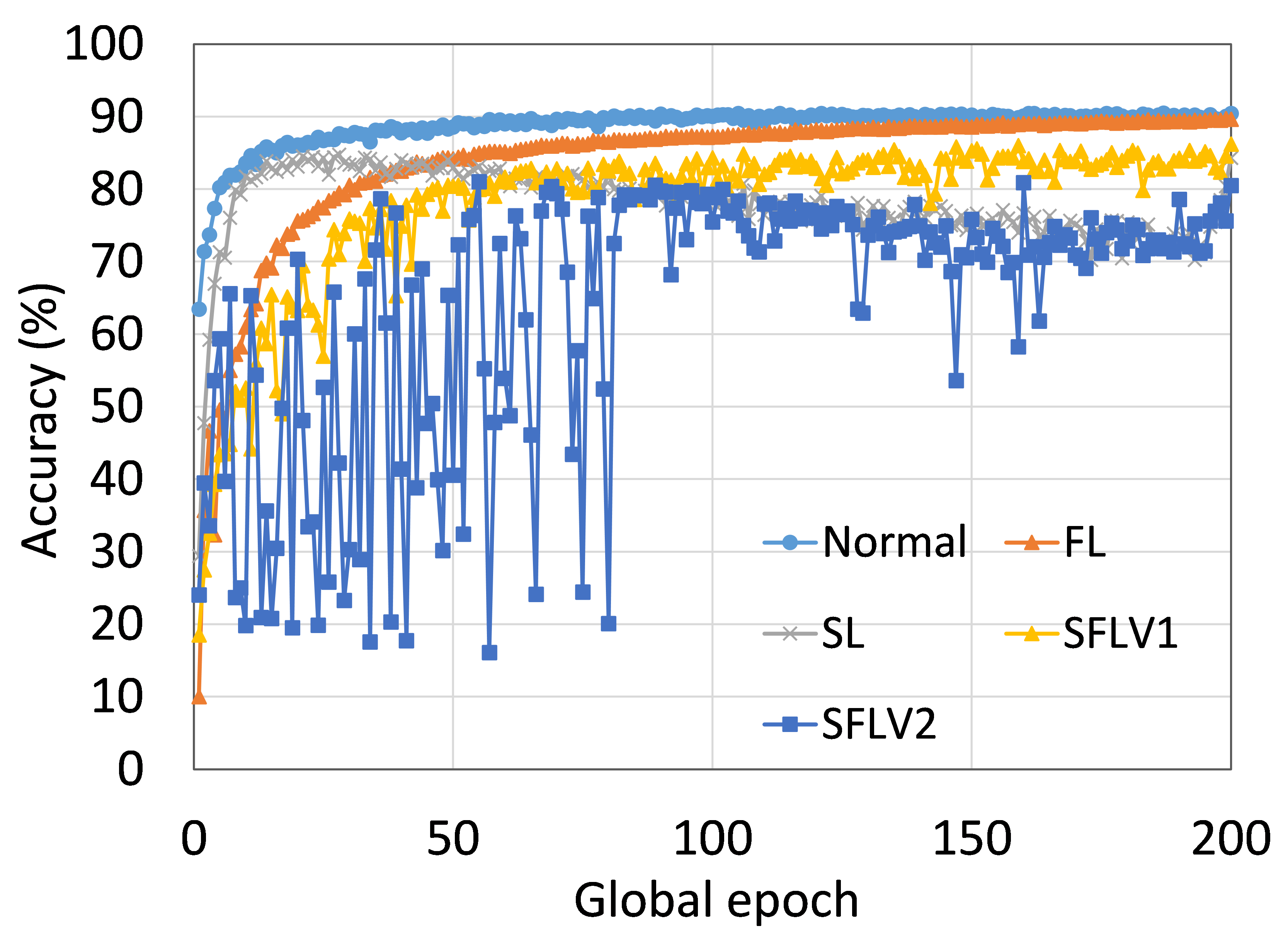}
		}	
			\hskip-2pt
    %------------LeNet on cifar10-------------------------------------------->	
	    \subfigure[\scriptsize Train: LeNet5 on CIFAR10]{
			\includegraphics[width=0.23\linewidth]{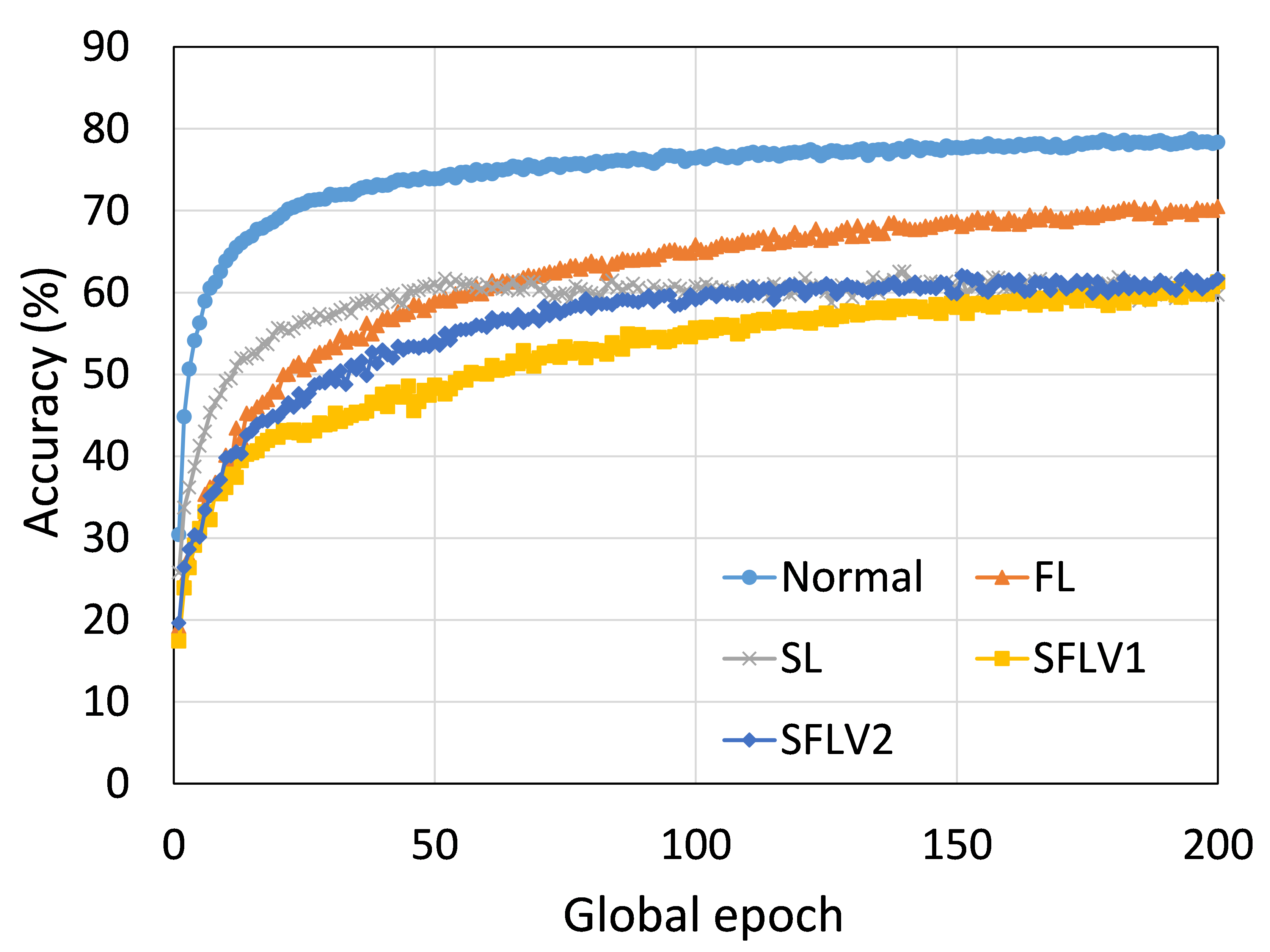}
		}
			\hskip-2pt
		\subfigure[\scriptsize Test: LeNet5 on CIFAR10]{
			\includegraphics[width=0.23\linewidth]{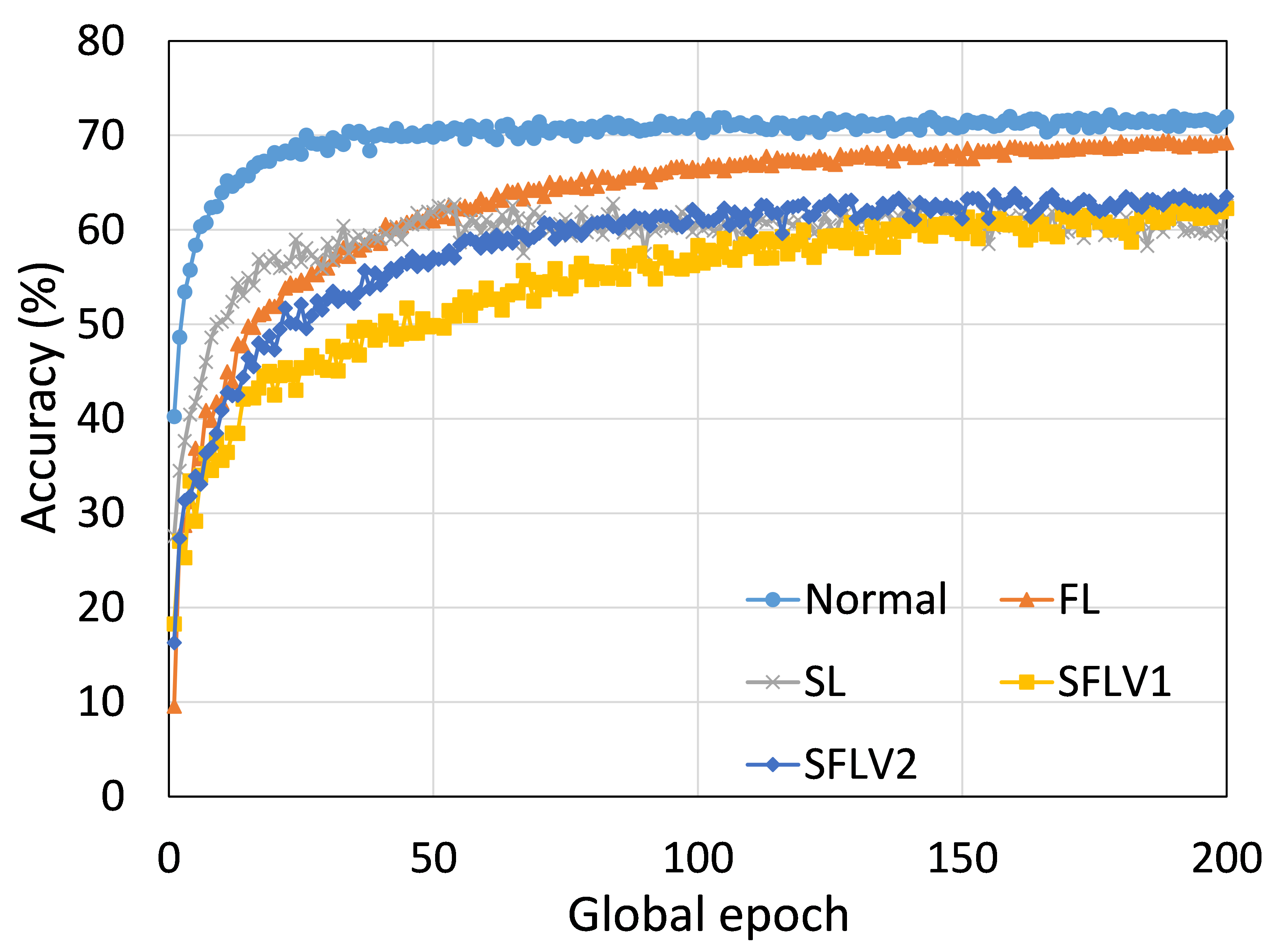}
		}	
			\hskip-2pt
%================================================================		
	%------ VGG16 on cifar10-------------------------------------------->	
	    \subfigure[\scriptsize Train: VGG16 on CIFAR10]{
			\includegraphics[width=0.24\linewidth]{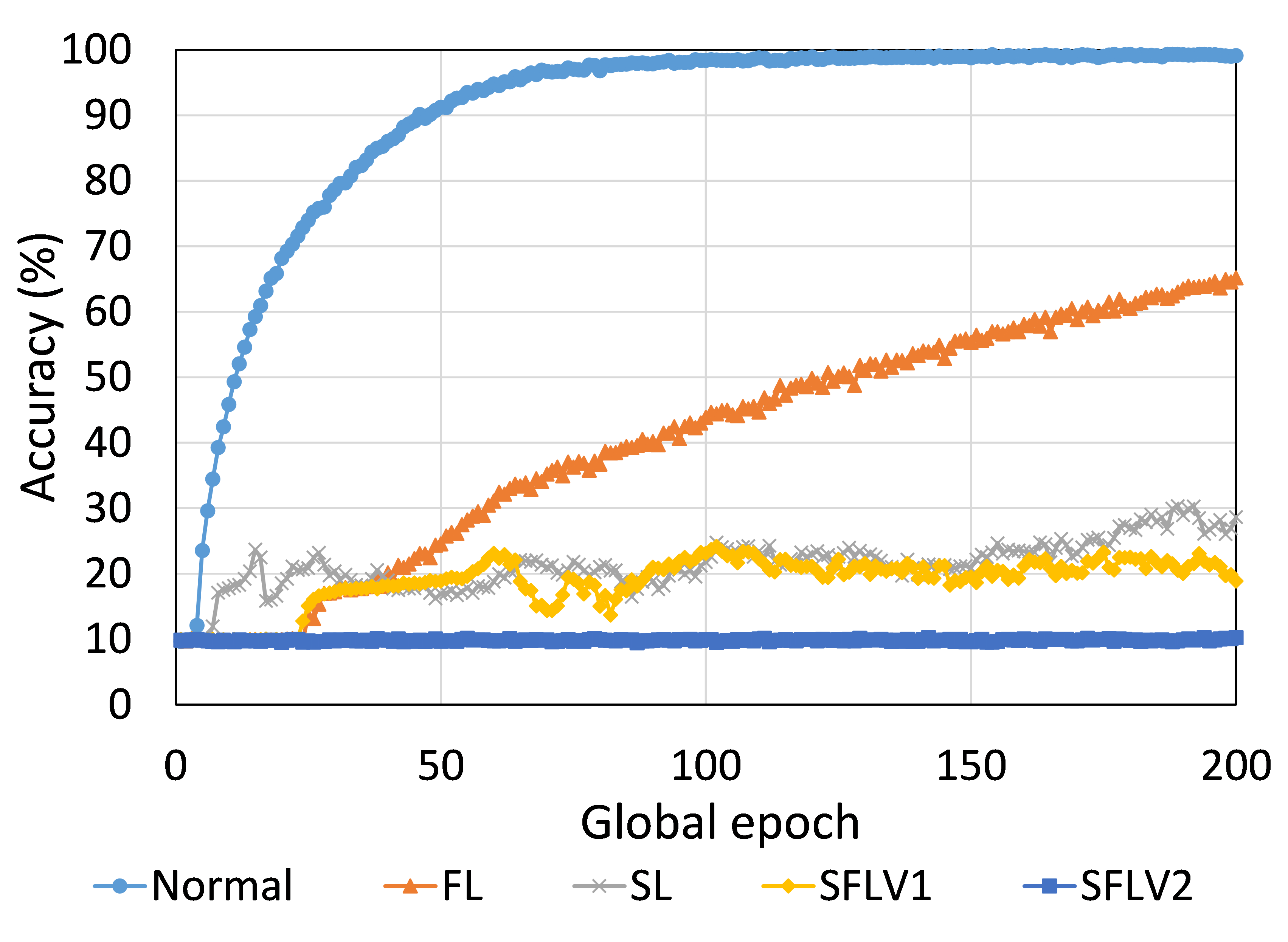}
			
		}
		\subfigure[\scriptsize Test: VGG16 on CIFAR10]{
			\includegraphics[width=0.24\linewidth]{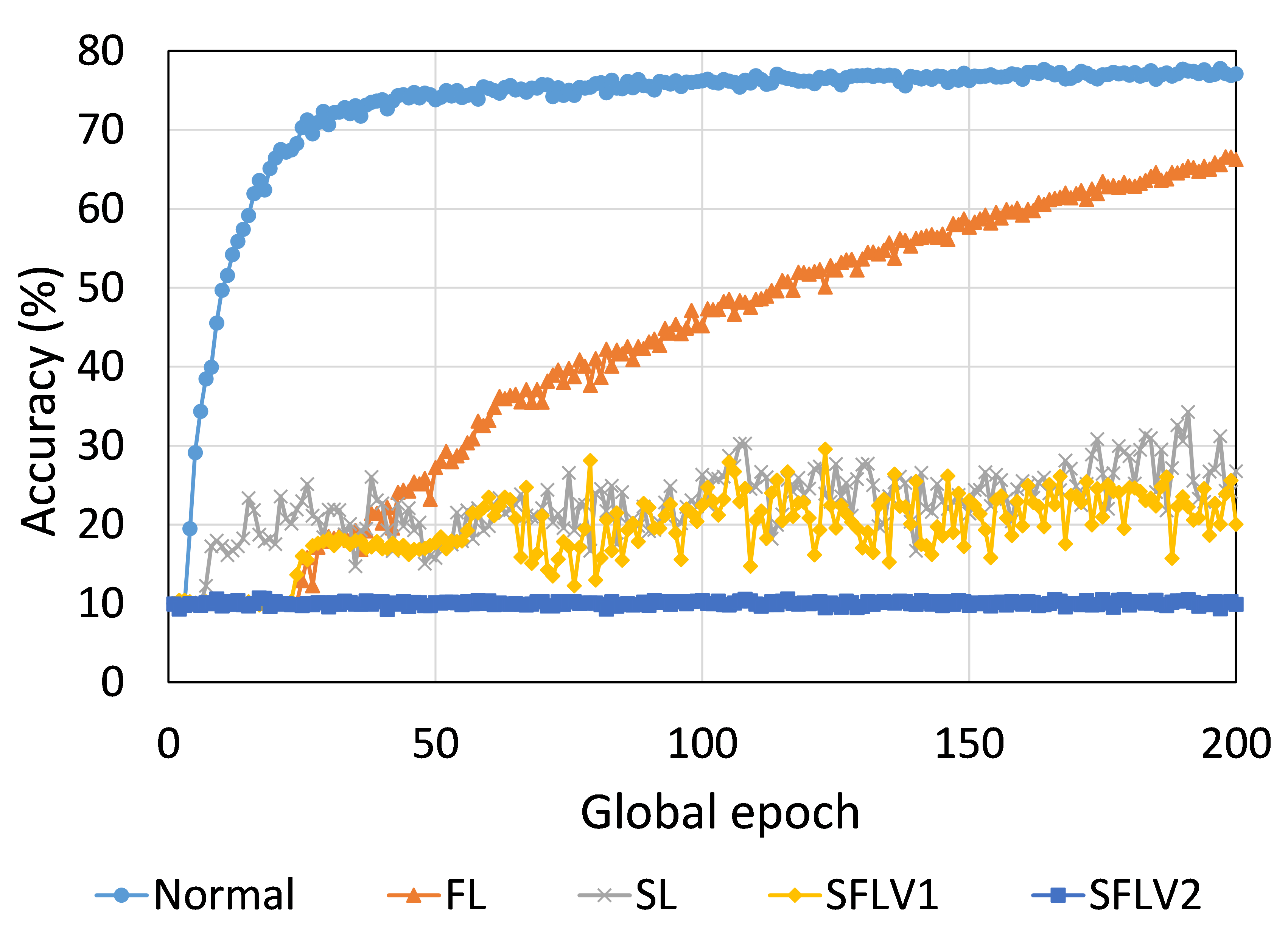}
			
		}	
		\vskip-5pt
		\caption{Performance of FL, SL, SFLV1, SFLV2: Training and testing convergence with five clients.}
		\label{fig:alllearningandperformance}
	
	\end{figure*}

	The main observation from these results is that FL, SL, SFLV1 and SFLV2 behave similar characteristics while training and testing in most of the cases. Besides, the best performance of an individual approach depends on the dataset. There can be the worst cases, where, for some cases, some approaches may not learn in a naive implementation. For example, SL, SFLV1, and SFLV2 can suffer from no learning in VGG16 on CIFAR10 (see Fig.~\ref{fig:alllearningandperformance}(m) and~\ref{fig:alllearningandperformance}(n)). This requires further investigation. In other cases, our DCML approaches learn in an expected way. For some datasets like CIFAR10, FL performs well compared to others in a similar setting; however, SL, SFLV1, and SFLV2 have advantages due to split network architecture, like low computation requirements at the client-side and model privacy, over FL. Furthermore, among SFLV1 and SFLV2, the FedAvg at the server-side network is not always efficient; for example, AlexNet on HAM10000 (see Fig.~\ref{fig:alllearningandperformance}(b); most of the cases, FedAvg (in FL and SFLV1) is contributing towards promising results.

	%================================== subsection starts----------------------------------
	\subsection{Effects of Number of Users on the Performance}
	\label{effectofusers}
	%\subsubsection{Effects of number of users for LeNet on FMNIST}
	%The following graphs depict the learning and performance of LeNet on FMNIST.
   
   This section presents the training and testing convergence plots of normal (centralized) learning, SL, FL, SFLV1, and SFLV2. The experiments are performed considering a various number of clients ranging from 5 to 100. Moreover, we consider LeNet5 on FMNIST and AlexNet on HAM10000.
   
\begin{figure*}[t]
	\centering
	%----------- LeNet on FMNIST Split learning--------------------------------------->
		\subfigure[\scriptsize FL Train: LeNet5 on FMNIST]{
			\includegraphics[width=0.23\linewidth]{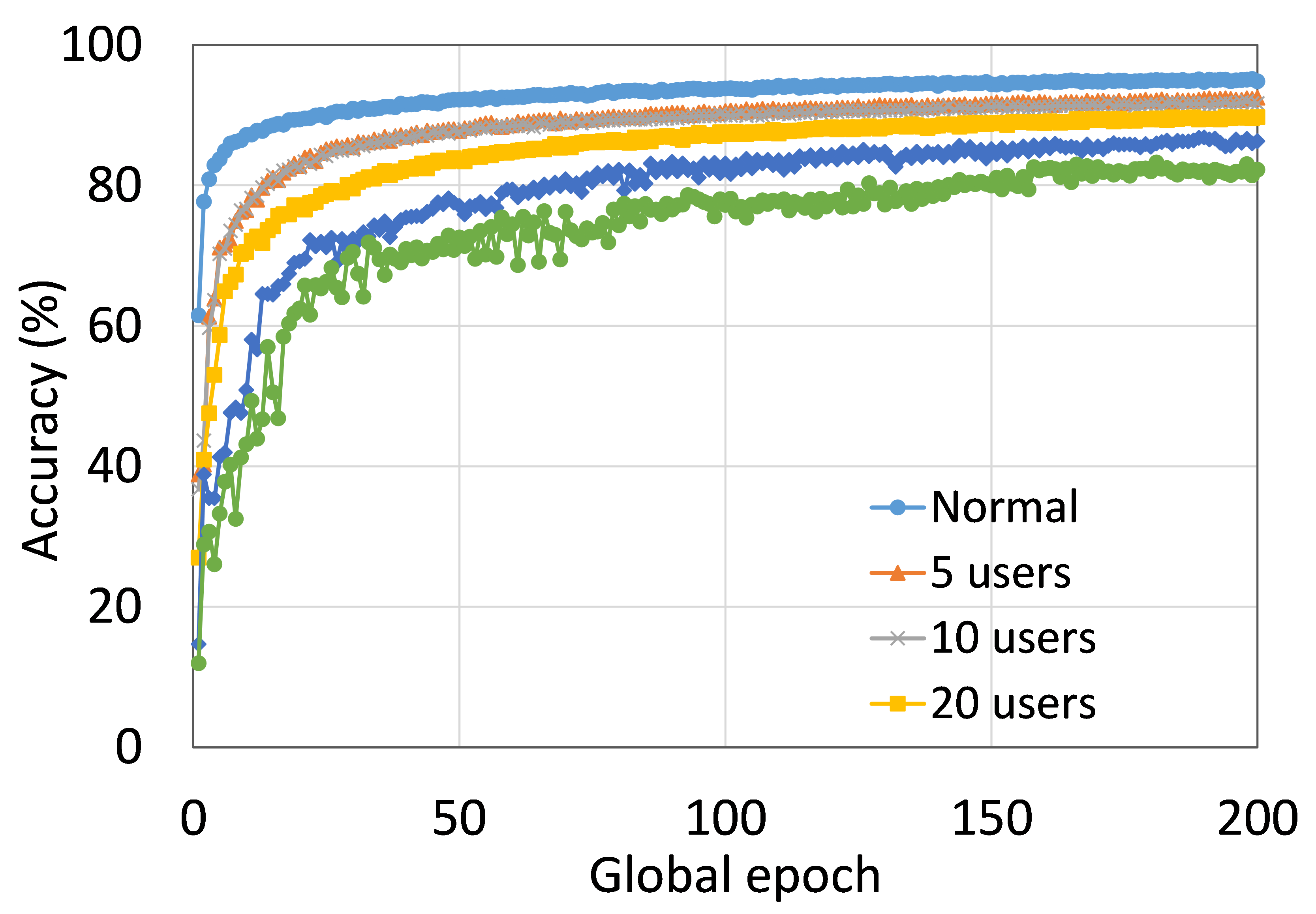}
		}
			\hskip-2pt
		\subfigure[\scriptsize FL Test: LeNet5 on FMNIST]{
			\includegraphics[width=0.24\linewidth]{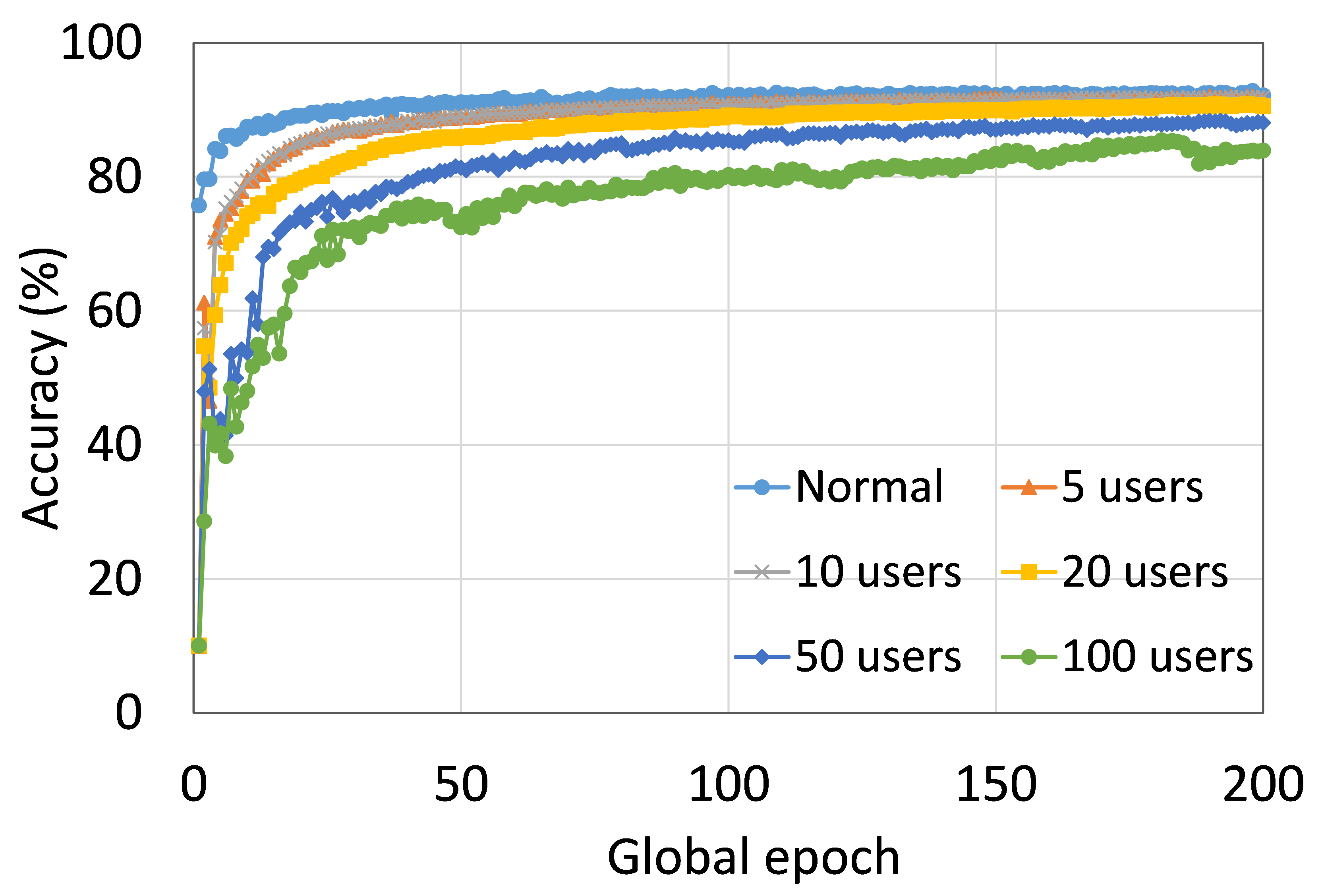}
		}
			\hskip-2pt
	%----------- LeNet on FMNIST Split learning--------------------------------------->	
		\subfigure[\scriptsize SL Train: LeNet5 on FMNIST]{
			\includegraphics[width=0.23\linewidth]{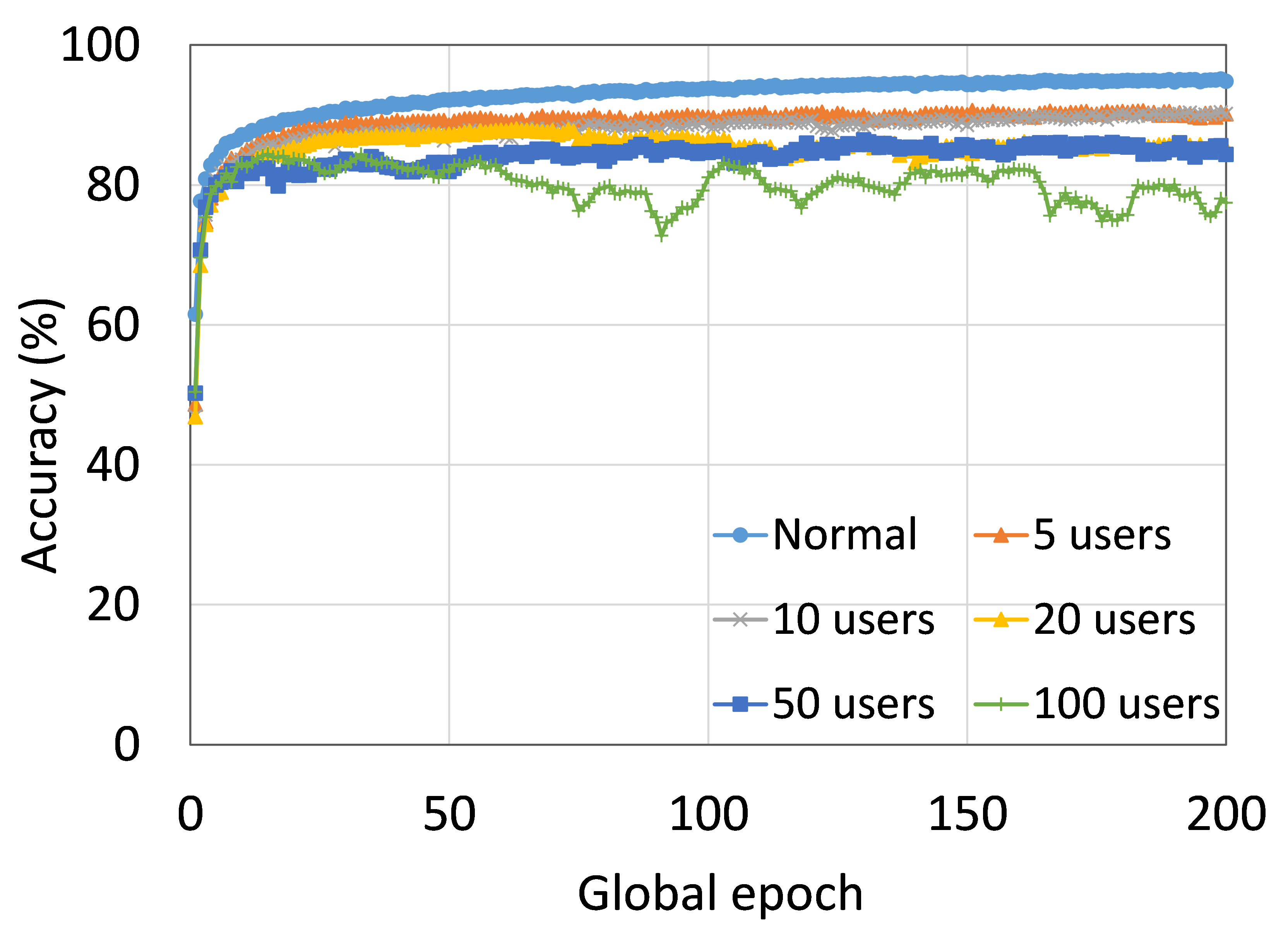}
		}
		\hskip-2pt
		\subfigure[\scriptsize SL Test: LeNet5 on FMNIST]{
			\includegraphics[width=0.23\linewidth]{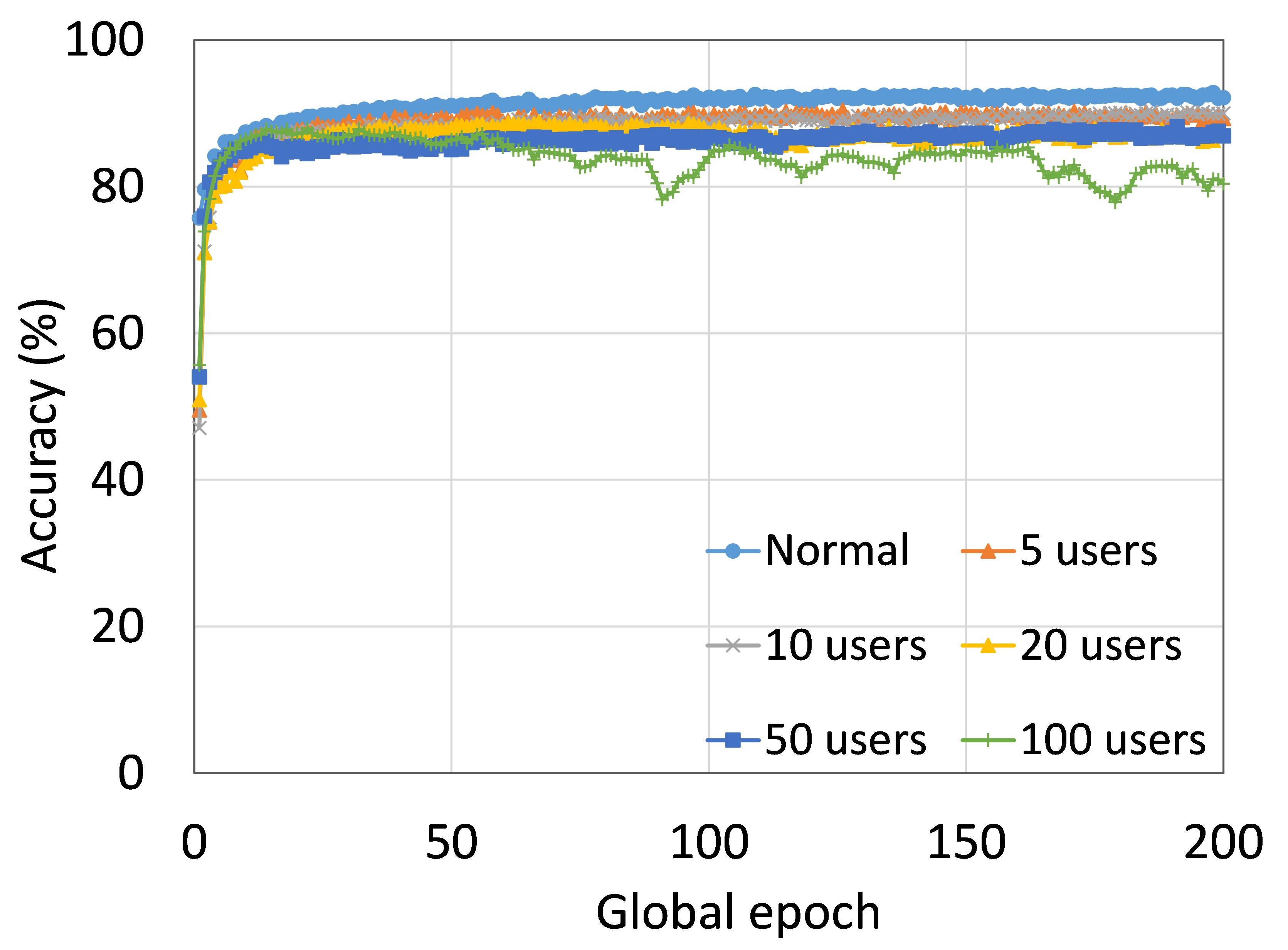}
		}	
			\hskip-2pt
%------------------------------------------------------
	%----------- LeNet on FMNIST splitfedv1 learning--------------------------------------->
	    \subfigure[\scriptsize SFLV1 Train: LeNet5 on FMNIST]{
			\includegraphics[width=0.24\linewidth]{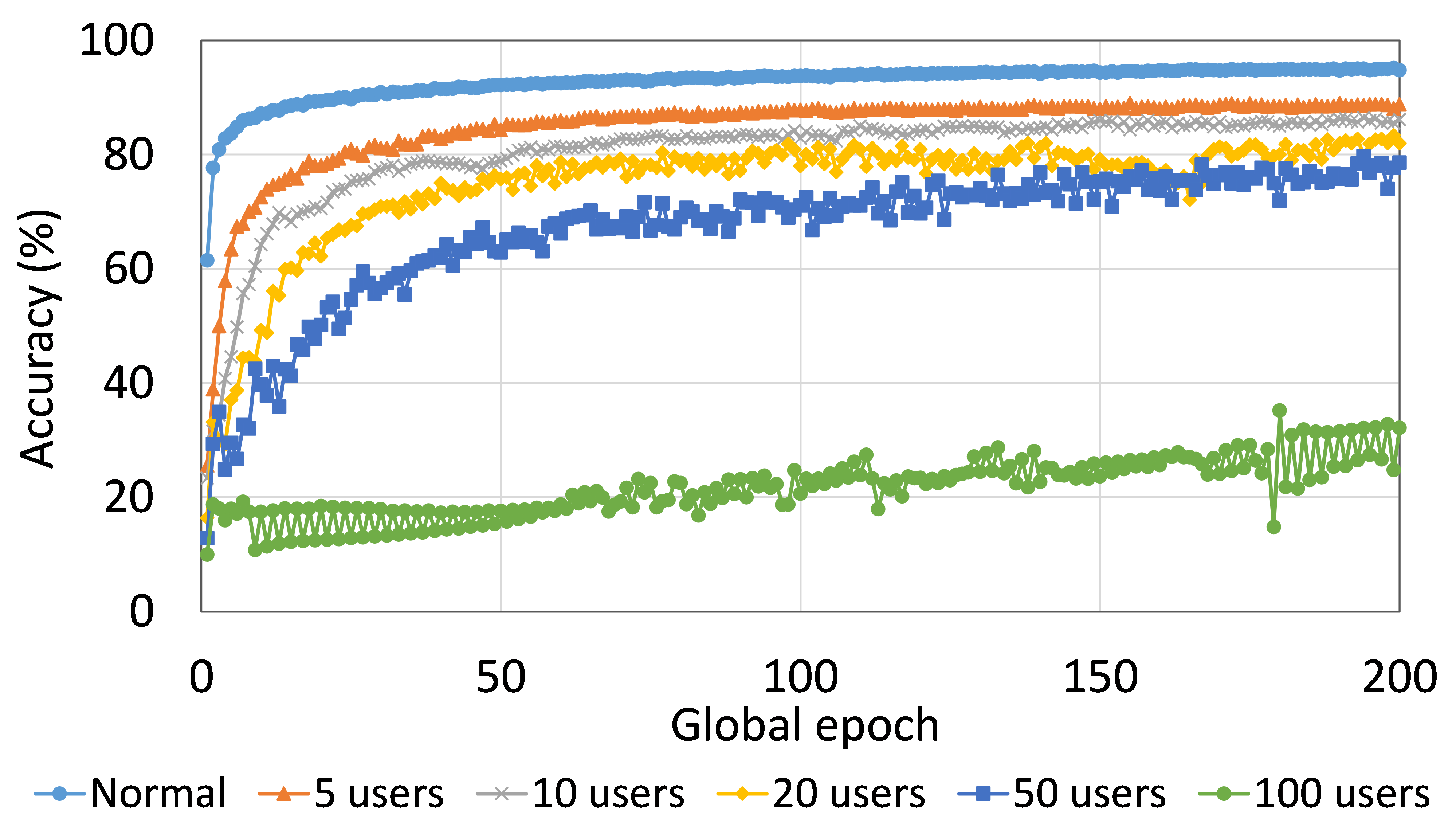}
		}
				\hskip-2pt
		\subfigure[\scriptsize SFLV1 Test: LeNet5 on FMNIST]{
			\includegraphics[width=0.24\linewidth]{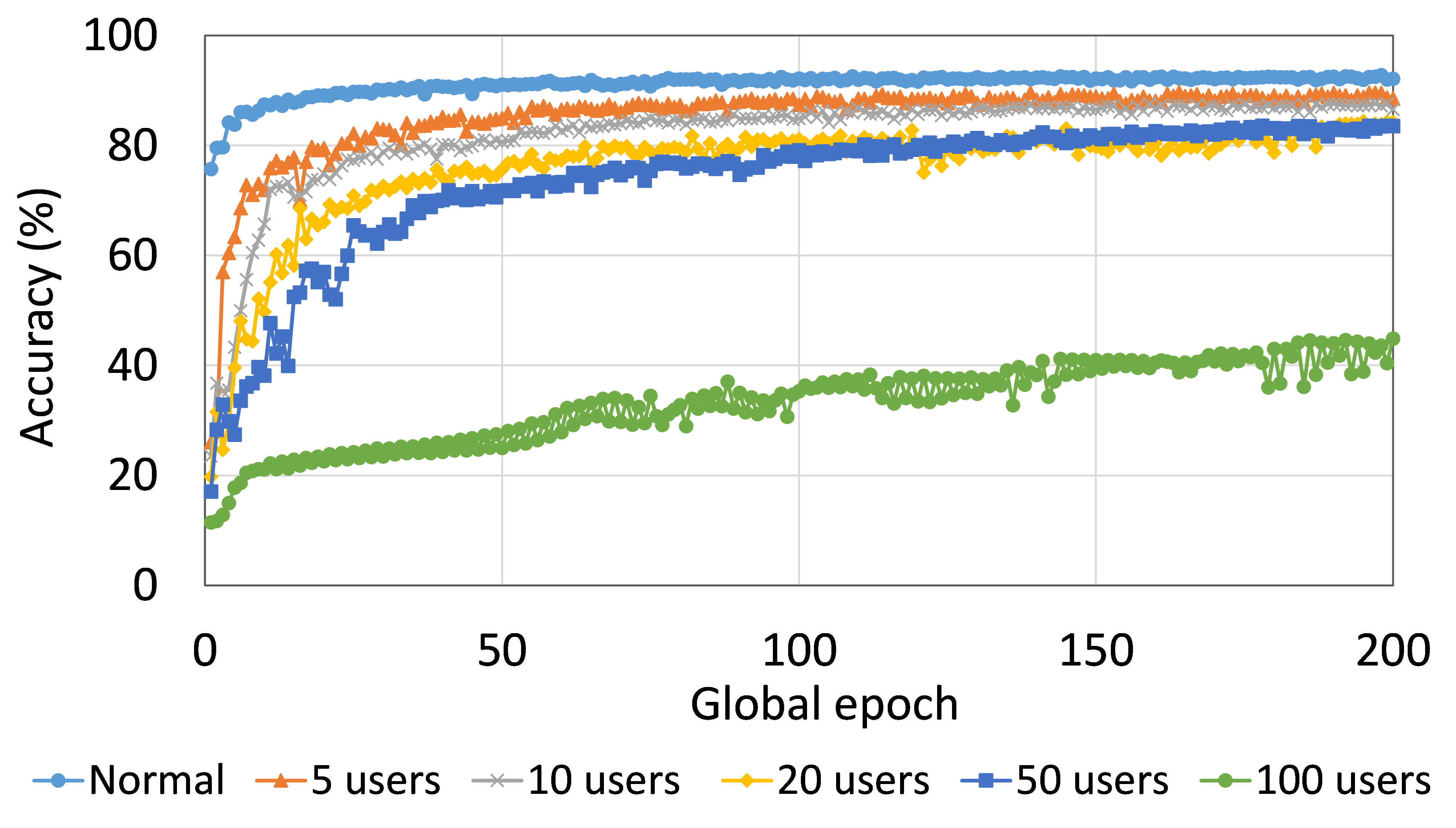}
		}
			\hskip-2pt
	%----------- LeNet on FMNIST splitfedv2 learning--------------------------------------->	
		\subfigure[\scriptsize SFLV2 Train: LeNet5 on FMNIST]{
			\includegraphics[width=0.22\linewidth]{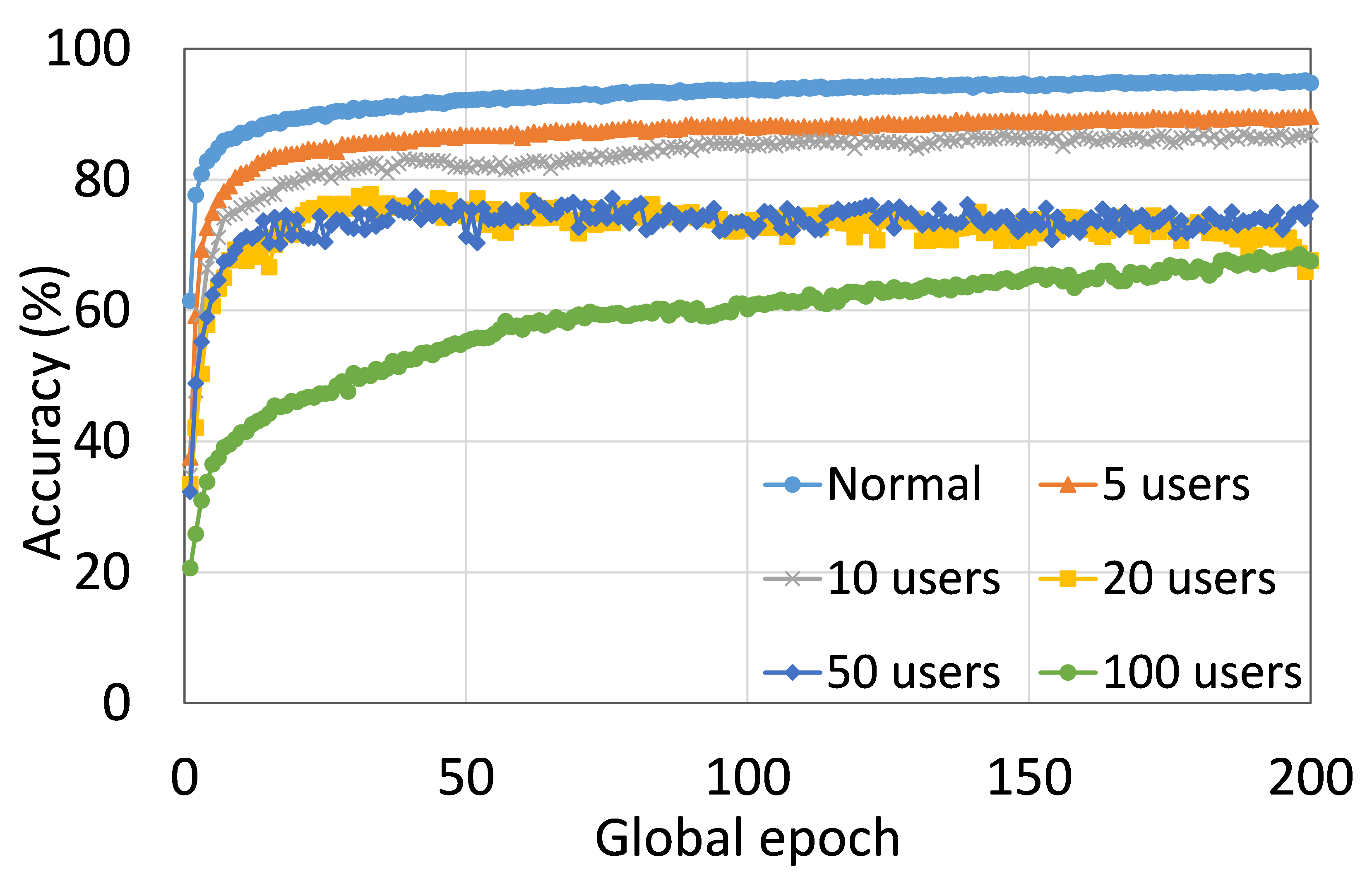}
		}
			\hskip-2pt
		\subfigure[\scriptsize SFLV2 Test: LeNet5 on FMNIST]{
			\includegraphics[width=0.22\linewidth]{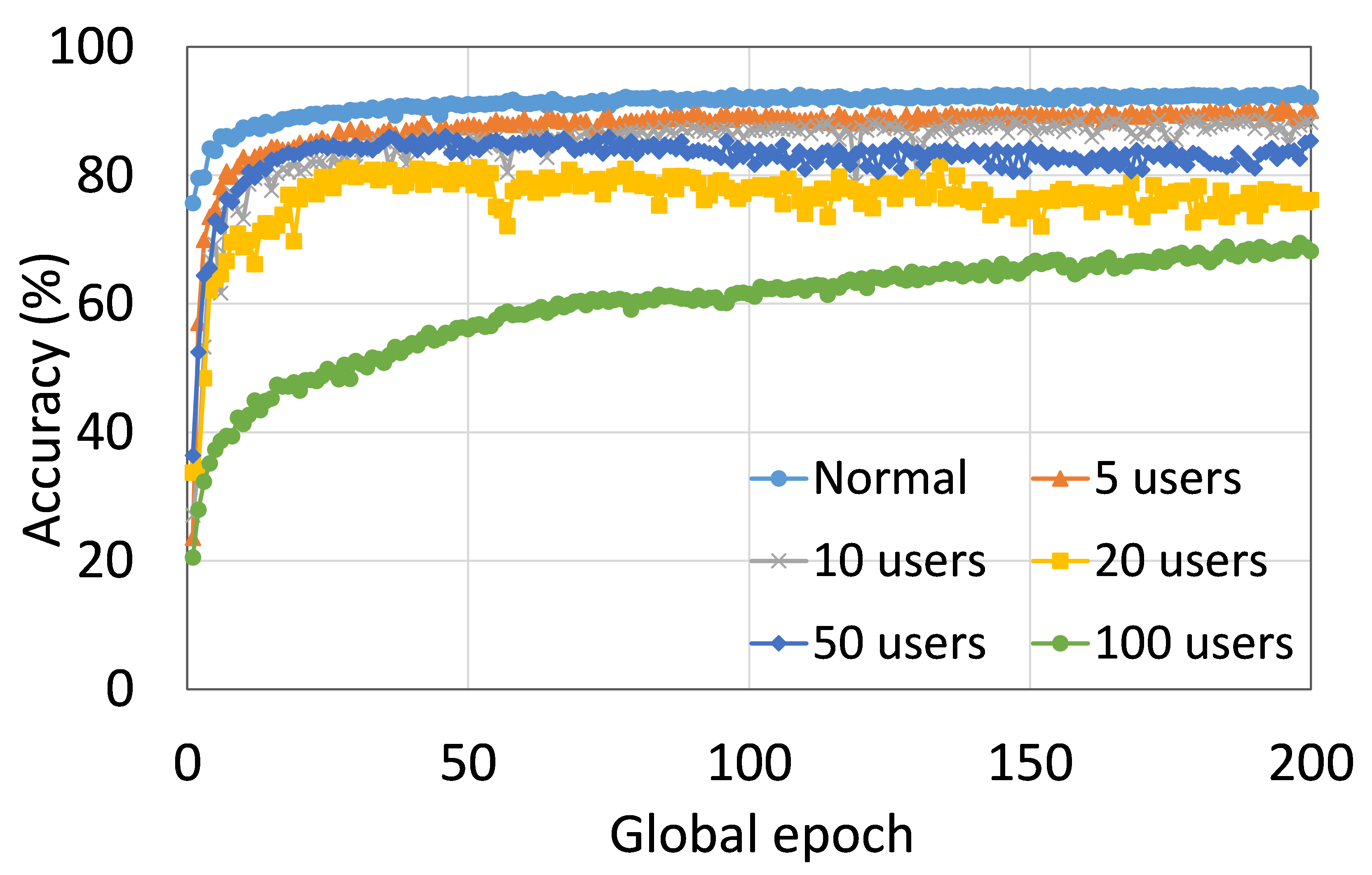}
		}
%=====================================================================
		%---------- next figures-------AlexNet--------------
		%----------fed learning -------
		\subfigure[\scriptsize FL Train: AlexNet on HAM10000]{
			\includegraphics[width=0.24\linewidth]{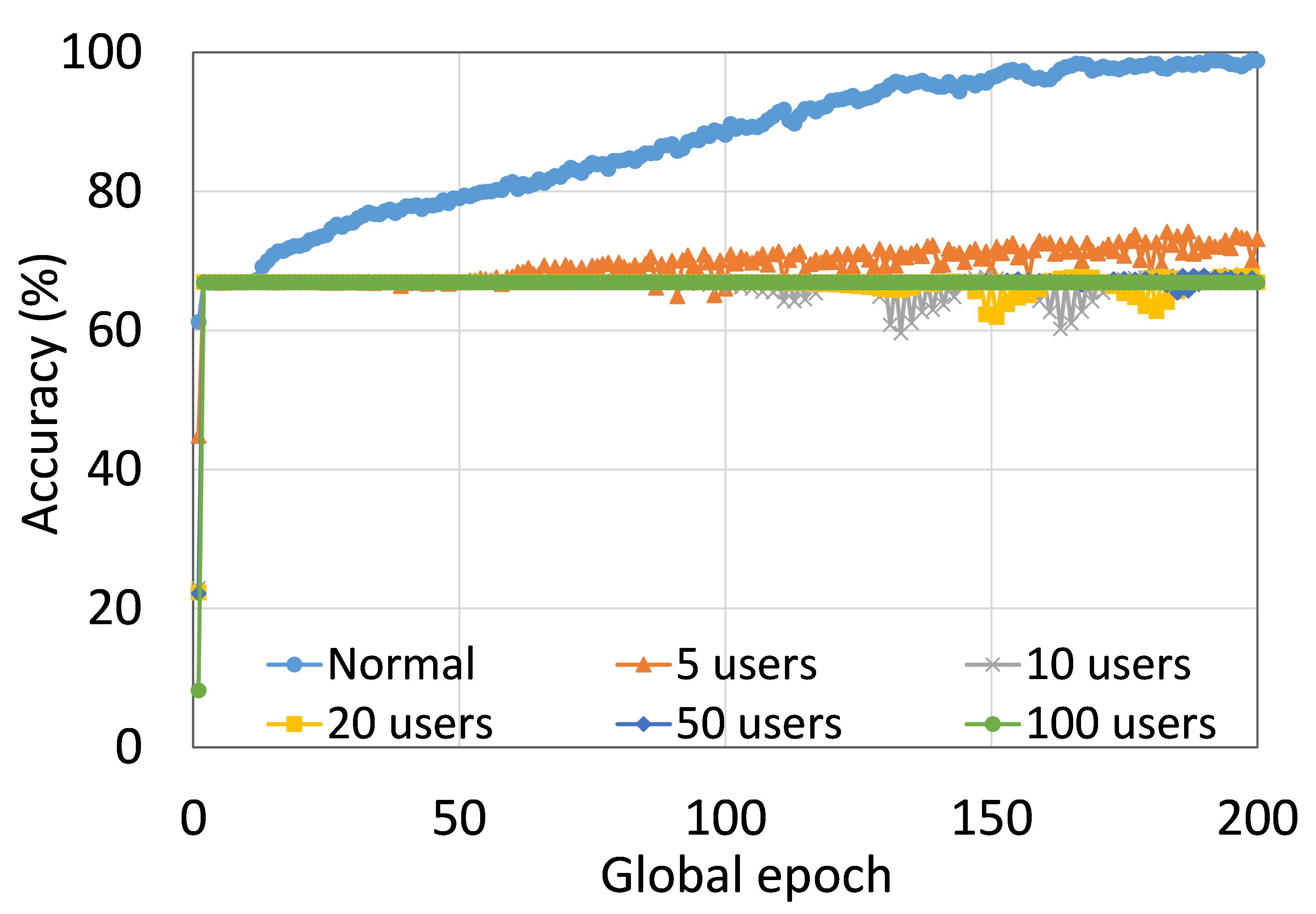}
		}
			\hskip-2pt
		\subfigure[\scriptsize FL Test: AlexNet on HAM10000]{
			\includegraphics[width=0.24\linewidth]{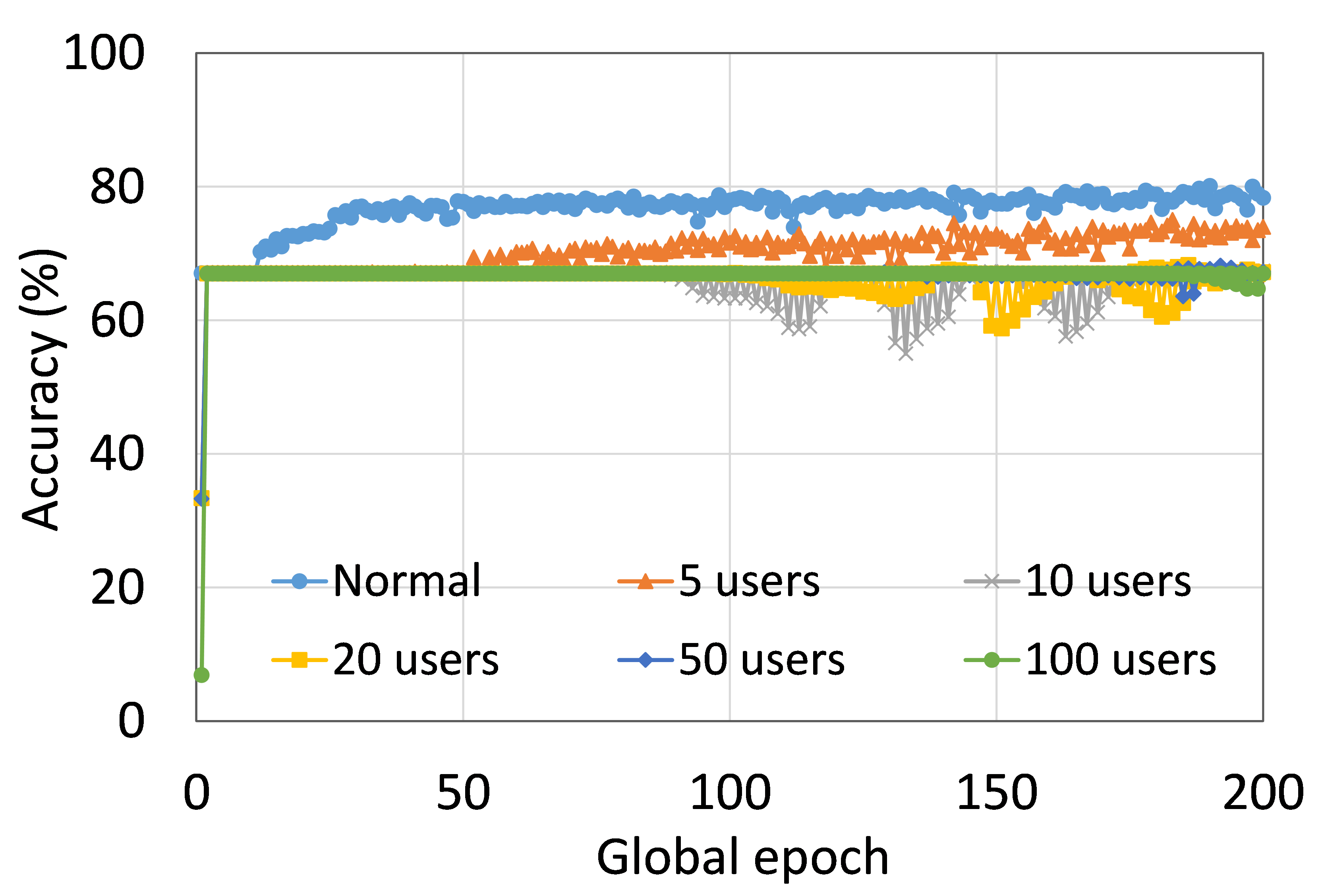}
		}
			\hskip-2pt
	%----------split learning -------	
	    \subfigure[\scriptsize SL Train: AlexNet on HAM10000]{
			\includegraphics[width=0.23\linewidth]{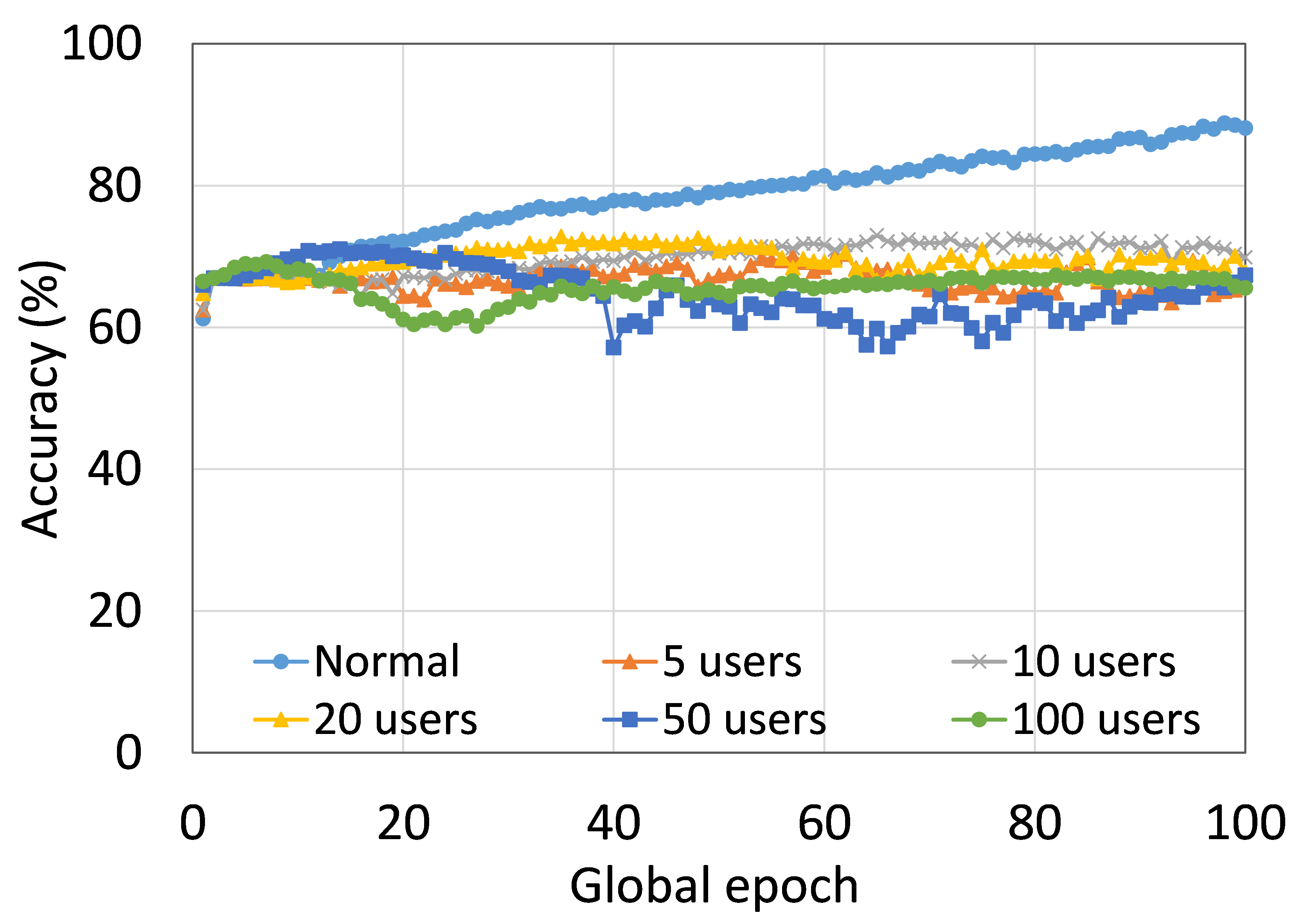}
		}
			\hskip-2pt
		\subfigure[\scriptsize SL Test: AlexNet on HAM10000]{
			\includegraphics[width=0.225\linewidth]{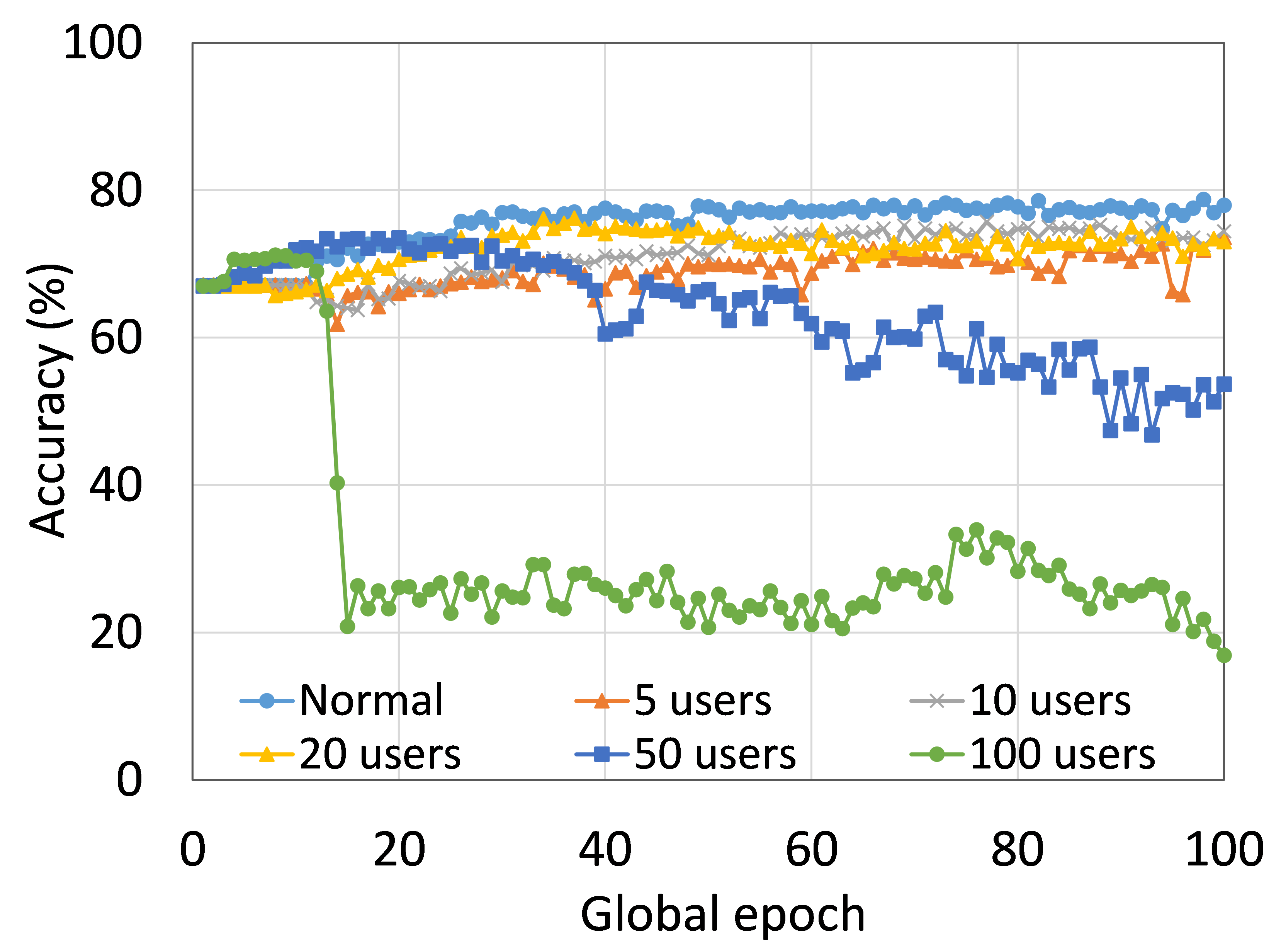}
		}	
			\hskip-2pt
%---------------------------------------------------------------------
	%----------splitfedv1 learning -------
		\subfigure[{\scriptsize SFLV1 Train: AlexNet on HAM10000}]{
			\includegraphics[width=0.23\linewidth]{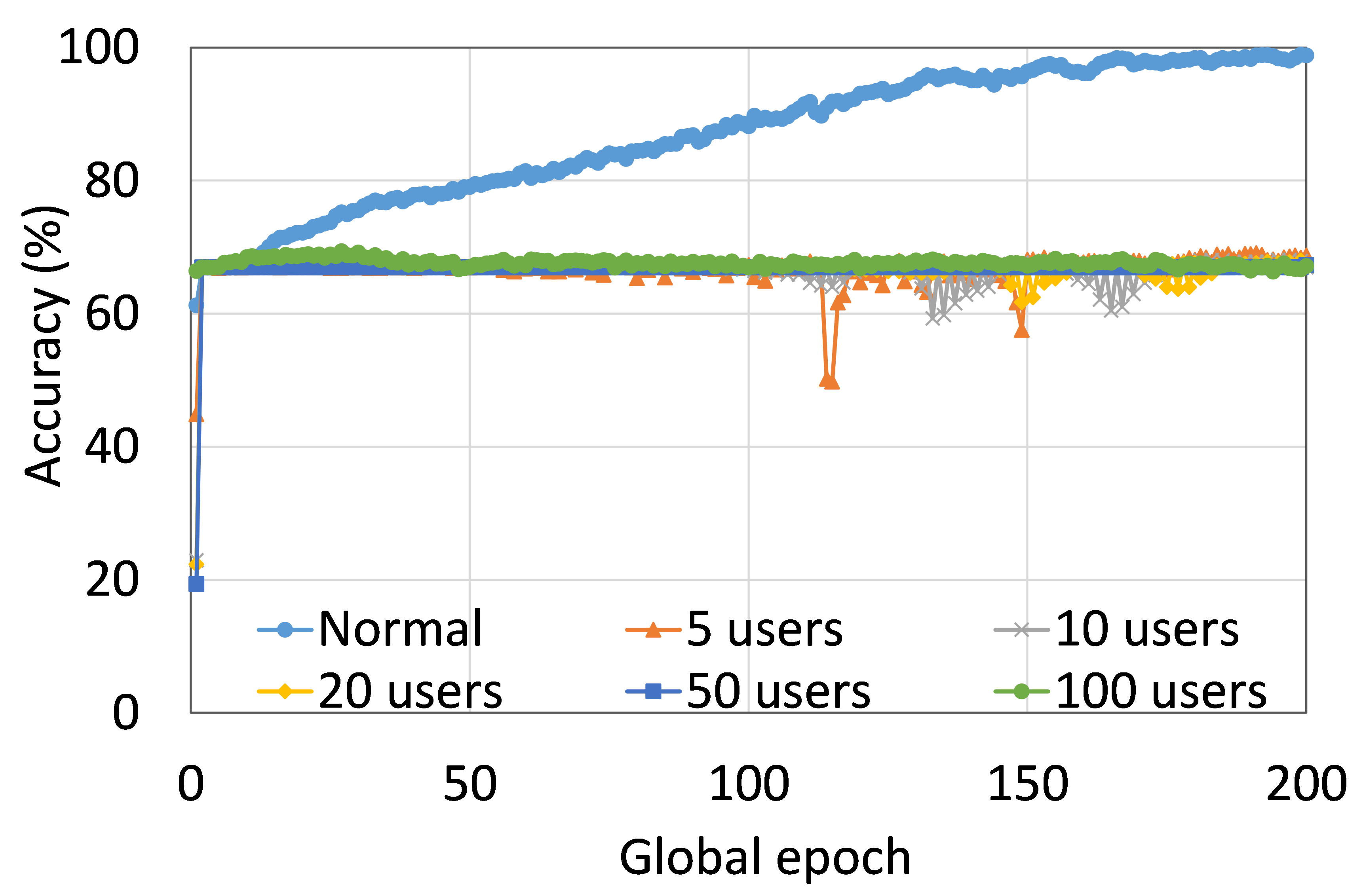}
		}
			\hskip-2pt
		\subfigure[{\scriptsize SFLV1 Test: AlexNet on HAM10000}]{
			\includegraphics[width=0.23\linewidth]{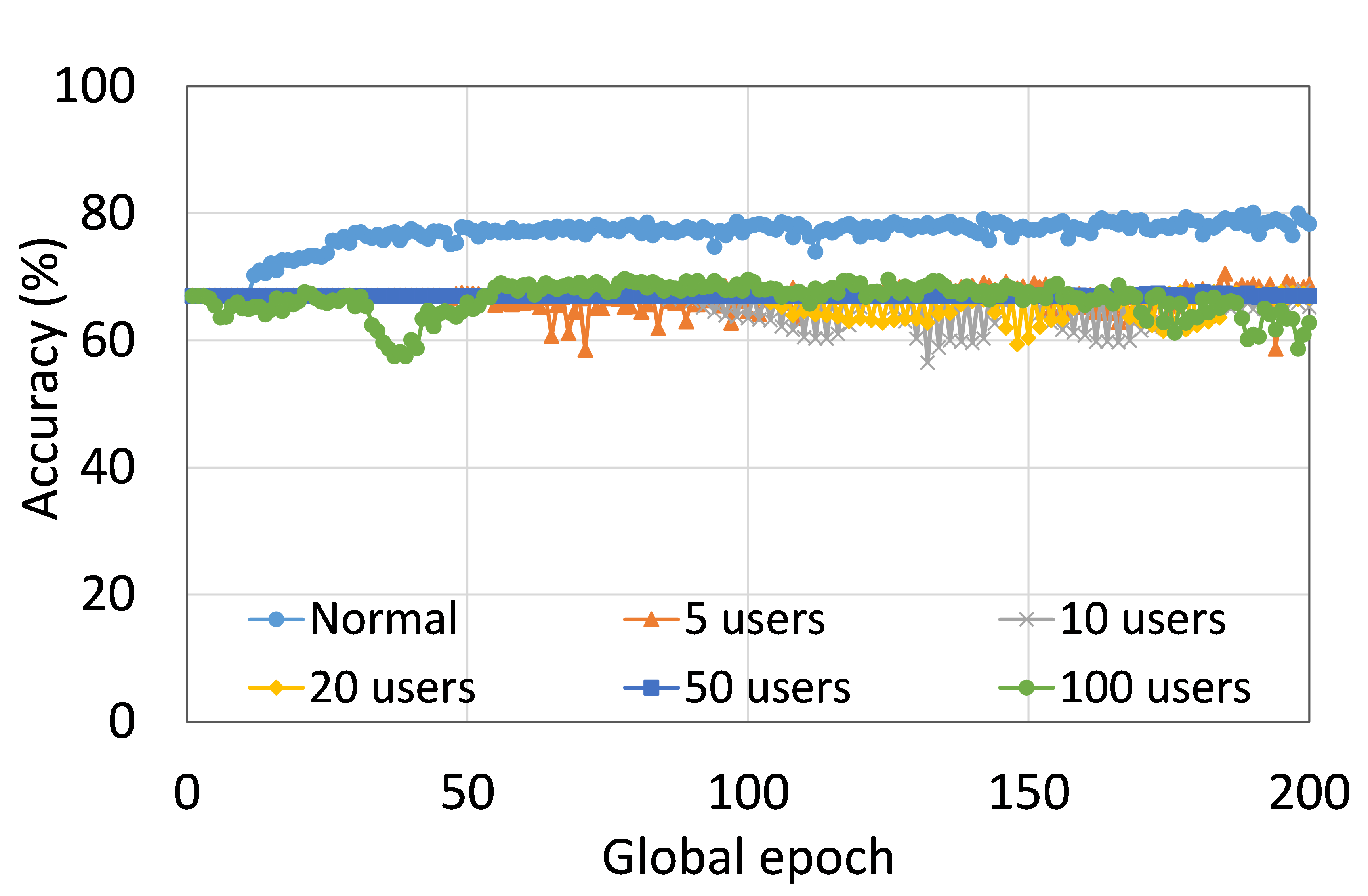}
		}	
			\hskip-2pt
	%----------splitfedv2 learning -------	
	    \subfigure[{\scriptsize SFLV2 Train: AlexNet on HAM10000}]{
			\includegraphics[width=0.245\linewidth]{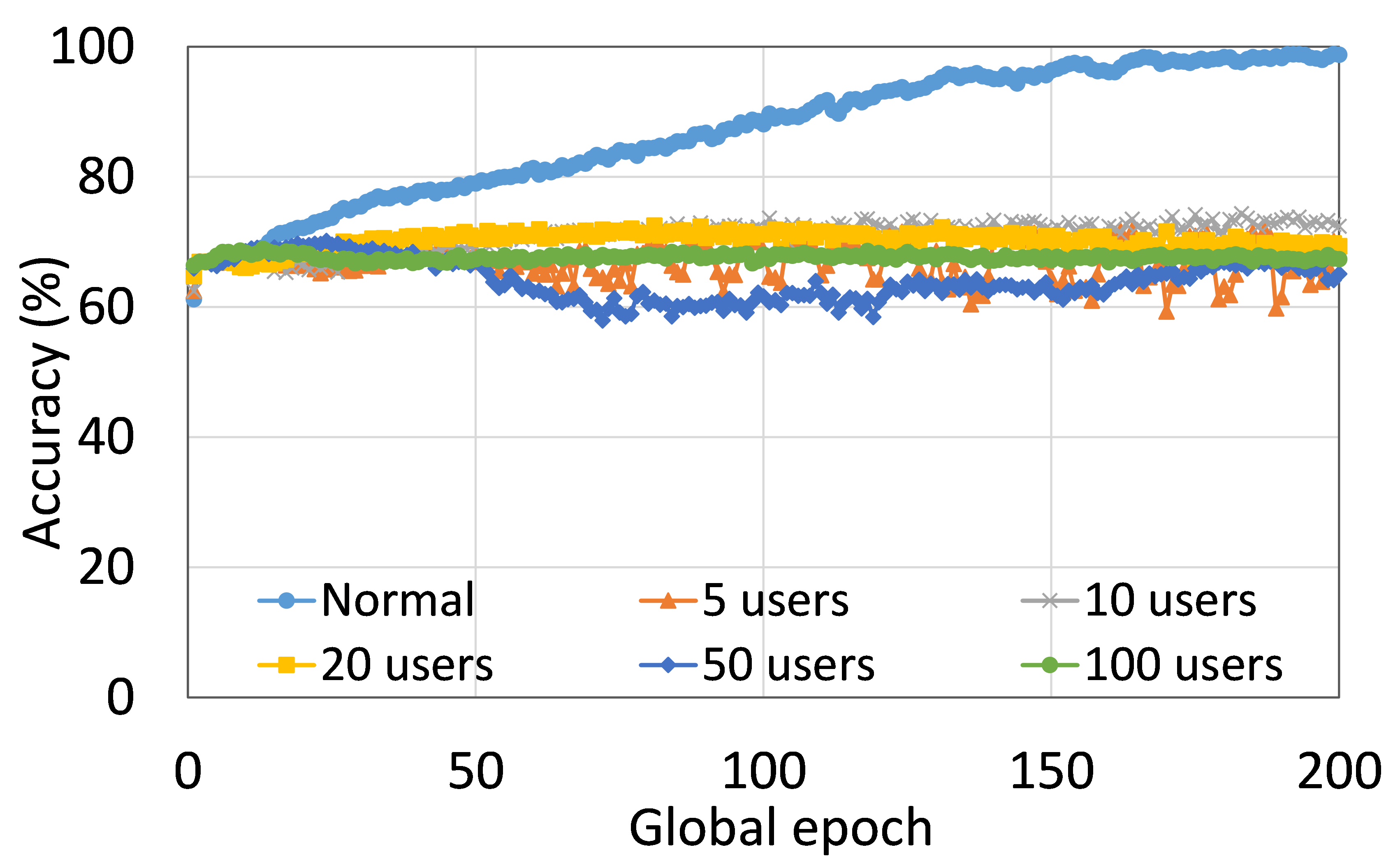}
		}
		\hskip-2pt
	\subfigure[{\scriptsize SFLV2 Test: AlexNet on HAM10000}]{
			\includegraphics[width=0.235\linewidth]{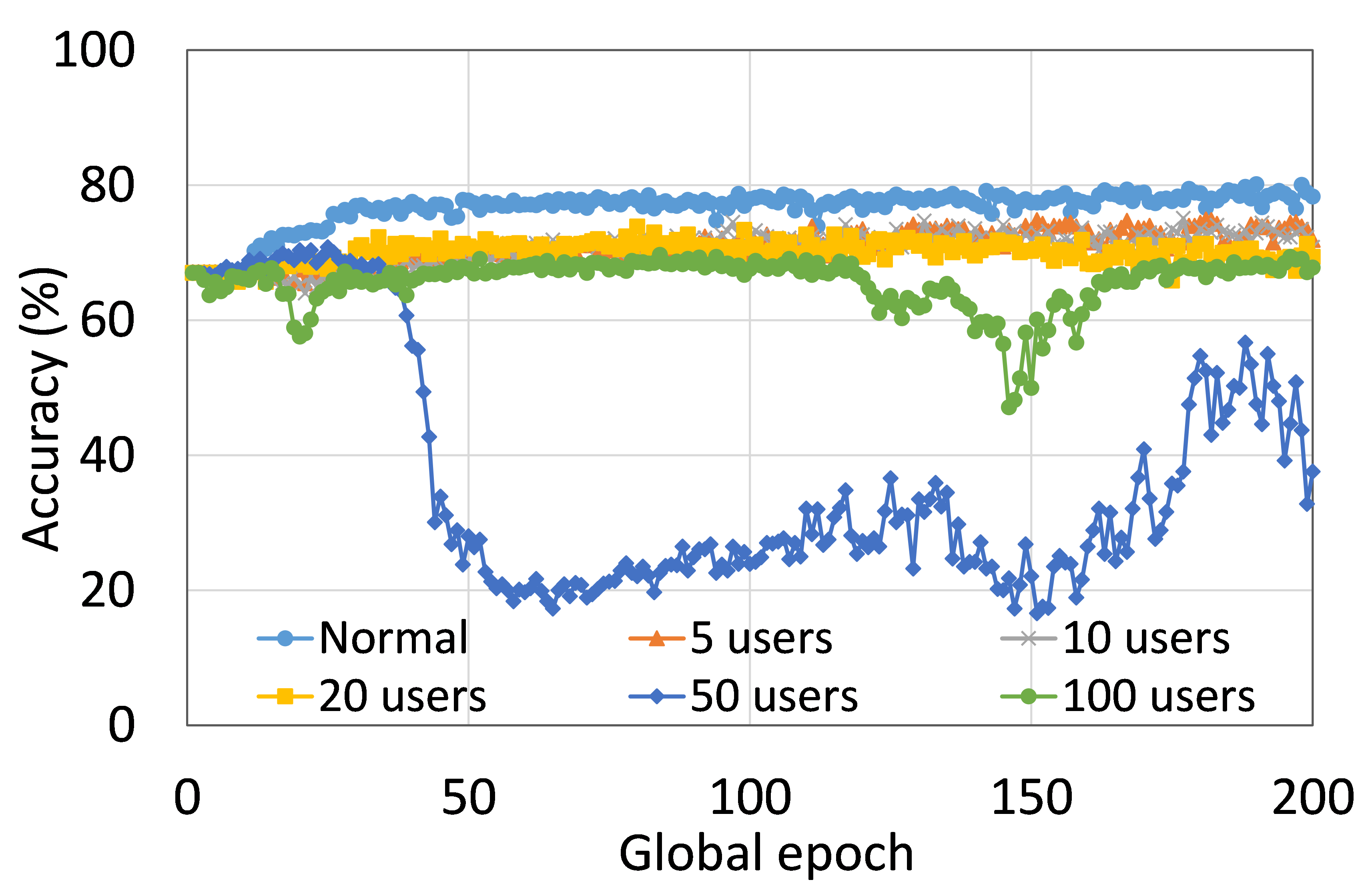}
		}	
		
		\caption{Effects of the number of users on the performance: Training and testing convergence with a various number of clients.}
		\label{fig:fmnistlenet4}
		
	\end{figure*}

   Our main observation is that usually, the convergence slows down, and performance degrades with the increase in the number of users (see Fig.~\ref{fig:fmnistlenet4} (b), (d), (f), and (h)) within the observation window of the global epochs. Furthermore, all our DCML approaches show similar behavior over the various number of users (clients). However, there are some cases with SL, where, despite model learning well during training, there exists a sharp fall in testing performance, for example, SL with AlexNet on HAM10000 with 100 users (see Fig.~\ref{fig:fmnistlenet4}(l)). Similarly, SFLV2, which acts similarly to the SL at the server-side operations, also shows a similar fall, but for 50 users. This particular case requires further investigation.

\end{document}